\documentclass[sigconf]{acmart}

\AtBeginDocument{%
  }

\setcopyright{cc}
\setcctype{by}
\copyrightyear{2026}
\acmYear{2026}
\acmDOI{10.1145/3767308.3835195}

\acmConference[MM '26]{Proceedings of the 34th ACM International Conference on Multimedia}{November 10--14, 2026}{Rio de Janeiro, Brazil}
\acmBooktitle{Proceedings of the 34th ACM International Conference on Multimedia (MM '26), November 10--14, 2026, Rio de Janeiro, Brazil}

\acmISBN{979-8-4007-2213-4/2026/11}

\usepackage{multirow}
\usepackage{changepage}
\usepackage[skip=2pt]{caption}
\usepackage{placeins}
\usepackage{utfsym}
\usepackage{colortbl}
\usepackage{graphicx}      
\usepackage[table]{xcolor} 
\usepackage{utfsym}

\begin{document}

\title[Compass: Degradation-Simulated Reciprocal Learning with Needle RWKV for Multimodal Crack Segmentation]%
{Compass: Degradation-Simulated Reciprocal Learning with Lightweight Needle RWKV for Multimodal Crack Segmentation under Missing Modalities}

\author{Hui Liu}
\affiliation{%
  \department{Engineering Research Center of Learning-Based Intelligent System (Ministry of Education)}
  \institution{Tianjin University of Technology}
  \city{Tianjin}
  \country{China}
}
\email{liuhui1109@stud.tjut.edu.cn}
\orcid{0009-0009-5742-0202}

\author{Chen Jia}
\correspondingauthor
\affiliation{%
  \department{Engineering Research Center of Learning-Based Intelligent System (Ministry of Education)}
  \institution{Tianjin University of Technology}
  \city{Tianjin}
  \country{China}
}
\email{jiachen@email.tjut.edu.cn}
\orcid{0000-0002-7535-479X}

\author{Fan Shi}
\affiliation{%
  \department{Engineering Research Center of Learning-Based Intelligent System (Ministry of Education)}
  \institution{Tianjin University of Technology}
  \city{Tianjin}
  \country{China}
}
\email{shifan@email.tjut.edu.cn}
\orcid{0000-0003-2074-0228}

\author{Xu Cheng}
\affiliation{%
  \department{Engineering Research Center of Learning-Based Intelligent System (Ministry of Education)}
  \institution{Tianjin University of Technology}
  \city{Tianjin}
  \country{China}
}
\email{xu.cheng@ieee.org}
\orcid{0000-0002-4724-5748}

\author{Mianzhao Wang}
\affiliation{%
  \department{Engineering Research Center of Learning-Based Intelligent System (Ministry of Education)}
  \institution{Tianjin University of Technology}
  \city{Tianjin}
  \country{China}
}
\email{wmz@email.tjut.edu.cn}
\orcid{0000-0001-7380-0955}

\author{Shengyong Chen}
\affiliation{%
  \department{Engineering Research Center of Learning-Based Intelligent System (Ministry of Education)}
  \institution{Tianjin University of Technology}
  \city{Tianjin}
  \country{China}
}
\email{sy@ieee.org}
\orcid{0000-0002-6705-3831}

\renewcommand{\shortauthors}{Hui Liu et al.}

\begin{abstract}
In multimodal crack segmentation for industrial facilities, the key challenge is preventing missing modalities from degrading pixel-level performance while maintaining low computational cost. Existing methods struggle to address semantic degradation caused by missing modalities. We propose Compass, a lightweight network for robust crack segmentation under arbitrary missing modalities. Compass comprises Degradation Simulation Distillation (DSD), Needle Block, and  Evidential Topology-Preserving Fusion (ETPF). DSD constructs a degradation simulation stream that mimics more severe missing conditions and performs reciprocal distillation with the original stream, decoupling complete perception from degradation adaptation. Within DSD, Feature-Aware Prototype Transmitter (FAPT) performs modality agnostic prototype-guided feature completion to maintain semantic integrity under incomplete modality conditions. As a lightweight backbone, Needle injects crack-direction cues into WKV modulation and combines connectivity-aware gating with anisotropic context probing for structure-aware modeling. ETPF fuses multimodal features via Dempster-Shafer evidential combination with uncertainty-gated decoding, preserving crack topology while suppressing unreliable features. Experiments on three datasets demonstrate state-of-the-art (SOTA) performance under diverse missing modality scenarios. Even with 90\% depth modality missing on CrackDepth, Compass achieves F1 of 0.8216 and mIoU of 0.8434 with only 2.58M parameters. The code is available at \url{https://github.com/Karl1109/Compass}.
\end{abstract}



\maketitle

\section{Introduction}

\begin{figure}[!t]
  \centering
\includegraphics[width=0.48\textwidth]{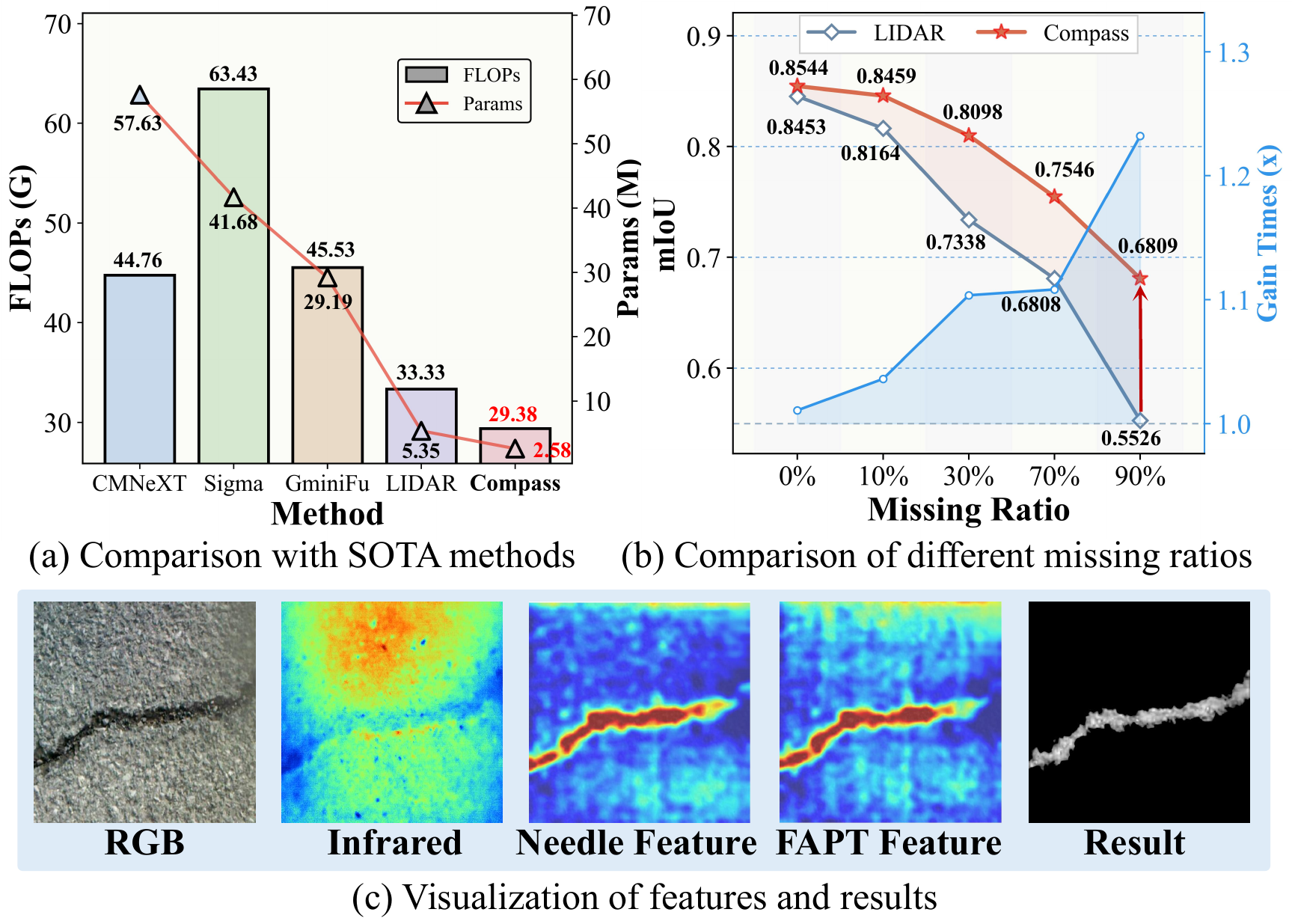}
  \caption{Performance of Compass. (a) compared the FLOPs and Params of different methods, (b) presented the performance under different RGB missing ratios  on CrackDepth \cite{Liu2025LIDAR}, (c) visualized the features and results on IRTCrack \cite{liu2022asphalt}.}
  \label{fig:intro}
  \vspace{-0.3cm}
\end{figure}

Multimodal visual crack segmentation has become a key technology in structural health monitoring for industrial and civil facilities \cite{Cheng2024Selective, Chen2024Mind, Benz2024Omni, Chu2024CrackGauGAN}. By fusing thermophysical distributions from infrared thermography, surface textures from polarization, and geometric priors from light-field depth, multimodal methods achieve significantly improved robustness over RGB-only approaches \cite{Liu2021CrackFormer, Liu2025SCSegamba, Yang2020Feature, Jaziri2024Designing} under complex scenarios such as abrupt illumination changes and background interference \cite{Liu2025LIDAR}. In general segmentation tasks, CMX \cite{Zhang2023CMX} and CMNeXT \cite{Zhang2023Delivering} achieve deep semantic alignment through cross-modal attention, MCubeS \cite{Liang2022Multimodal} integrates multimodal radiation characteristics via region-guided filters, and Sigma \cite{Wan2025Sigma} and LIDAR \cite{Liu2025LIDAR} leverage enhanced visual state space modules for efficient cue-aware fusion. However, all these methods assume full modality availability. In real-world industrial deployment, sensor failures and transmission interruptions are prevalent \cite{Lee2023Multimodal, Hong2022Cross, Maheshwari2024Missing}, causing not only complete modality loss but more commonly random partial missing within a modality. As shown in Figure \ref{fig:intro} \hyperref[fig:intro]{(b)}, existing architectures exhibit severe feature degradation and sharp performance drops as the intra-modal missing ratio increases.

\begin{figure*}[!t]
  \centering
  \includegraphics[width=\textwidth]{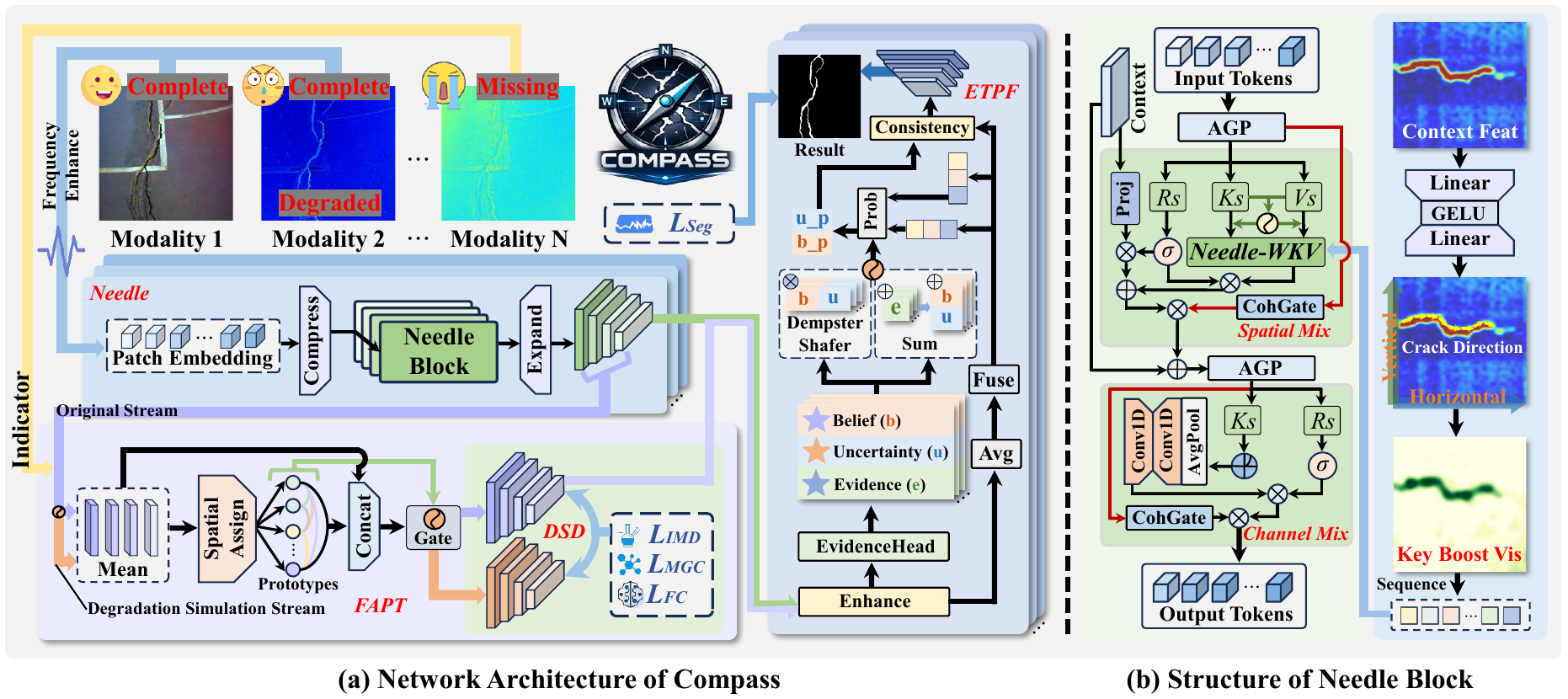}
  \caption{Overview of our Compass. Compass enhances and aggregates available modalities, completes missing modalities, and produces robust segmentation. (a) illustrates the architecture of Compass and the processing flow for multimodal data. (b) illustrates the structure of Needle Block.}
  \label{fig:Compass}
  \vspace{-0.1cm}
\end{figure*}

Methods specifically designed for crack segmentation under missing modalities remain scarce, though strategies have been explored in other domains. MD2N \cite{Dai2025Unbiased} proposes bidirectional diffusion for cross-modal refinement, CCSD \cite{Xie2025CCSD} employs hierarchical self-distillation to push incomplete features toward complete distributions, KnowledgeBridger \cite{Ke2025Knowledge} leverages large multimodal models for knowledge graph-based completion, MCMoE \cite{Xu2026MCMoE} generates missing inputs through adaptive gating, and TouchFormer \cite{Lyu2026TouchFormer} suppresses missing-modality noise via coupled cross-attention. However, generation-based and large model methods incur prohibitive computational overhead for edge deployment, while non-generative strategies suffer representation collapse under extreme missing scenarios. Furthermore, random modality dropout \cite{Neverova2015ModDrop, Wang2023Multi} forces a single training stream to learn semantics from complete data and adapt to degradation, two objectives whose gradients often conflict \cite{Nezakati2025MMP}, yielding limited gains under extreme missing ratios. How to decouple complete perception from degradation adaptation into a collaborative relationship remains an open problem.

Moreover, in efficient backbone design, accurately modeling crack anisotropic textures under strict computational budgets is a core challenge. CNNs offer efficient inference but lack long-range crack continuity due to local receptive fields \cite{Nam2024Modality, Seichter2022Efficient}. As shown in Figure \ref{fig:intro} \hyperref[fig:intro]{(a)}, Transformers possess global modeling capability but their quadratic complexity hinders edge deployment \cite{Liu2021Swin, Chen2025HSPFormer, Wang2024PSSD}. Mamba \cite{Gu2024Mamba, Gu2022Efficiently} achieves linear complexity, yet its causal nature restricts bidirectional context and leads to discontinuous boundaries \cite{Hatamizadeh2025MambaVision, Guo2025MambaIRv2, Yu2025MambaOut}. RWKV \cite{Peng2023RWKV} achieves sequence length independent linear complexity through recurrent state transfer, and VRWKV \cite{Duan2025Vision} introduces bidirectional modeling to eliminate the causal constraint. Subsequent works further extend RWKV in vision, PointRWKV \cite{He2025PointRWKV} for hierarchical point cloud learning, URWKV \cite{Xu2025URWKV} for noise suppression via multi-state Q-Shift, ZigRiR RWKV \cite{Chen2025Zig} for continuous feature modeling. However, existing RWKV vision methods universally rely on fixed-pattern spatial perception such as Q-Shift, applying preset directional shifts ill-suited for cracks with random orientations and irregular bifurcations. Meanwhile, the exponential decay in WKV applies a uniform rate across all tokens regardless of directional properties, rendering information propagation along crack extensions no more efficient than in background regions. A more structure-aware mechanism is needed to incorporate crack directional cues into sequence modeling, enabling the model to autonomously enhance propagation along crack orientations.

Furthermore, in multimodal fusion, missing modalities pose severe challenges, not only must heterogeneous features be effectively integrated, but each modal signal's reliability must be assessed to regulate fusion weights while maintaining topological connectivity on elongated, anisotropic crack targets. CEN \cite{Wang2023Channel} exchanges channels but lacks explicit reliability quantification. DFormerv2 \cite{Yin2025DFormerv2} injects depth geometry for cross-modal alignment but is limited to specific modality pairs without uncertainty modeling. FreqFusion \cite{Chen2024Frequency} improves boundary clarity but focuses on intra-modal fusion without addressing cross-modal conflict. In summary, existing fusion methods lack a unified treatment of uncertainty modeling and crack topological connectivity preservation.

To address these challenges, we propose Compass, a lightweight network for multimodal crack segmentation under arbitrary missing modalities. Compass comprises Degradation Simulation Distillation (DSD), Needle Block, and Evidential Topology-Preserving Fusion (ETPF). DSD decouples complete perception from degradation adaptation into collaborative dual-stream reciprocal learning, the original stream maintains current perception while the degradation simulation stream further masks modalities to simulate more severe degradation, and the two streams achieve reciprocity through mutual distillation. As shown in Figure \ref{fig:intro} \hyperref[fig:intro]{(c)}, within DSD, Feature-Aware Prototype Transmitter (FAPT) performs prototype-guided feature completion to enhance semantic integrity under incomplete modalities. As the lightweight backbone, Needle injects crack orientation into WKV key modulation through Geometry Conditioned Modulation (GCM) for direction-aware information propagation, integrated with connectivity-aware gating and anisotropic context probing. ETPF maps modal features to Dirichlet evidence parameters combined via the Dempster-Shafer rule, enhances crack connectivity through anisotropic evidence propagation, and suppresses unreliable features with an uncertainty-gated decoder. Experiments demonstrate that Compass achieves strong robustness and superior performance under diverse missing modality scenarios with extremely low computational cost.

Our main contributions are as follows:

\begin{adjustwidth}{2em}{0pt}
$\bullet$ We propose Compass, a lightweight Needle RWKV network for multimodal crack segmentation under missing modalities, achieving high precision segmentation with extremely low computational overhead.
\end{adjustwidth}

\begin{adjustwidth}{2em}{0pt}
$\bullet$ DSD enables proactive degradation adaptation through reciprocal learning with FAPT; Needle injects crack directional cues via GCM with connectivity-aware gating and anisotropic context probing; ETPF fuses multimodal features and generates the final segmentation map.
\end{adjustwidth}

\begin{adjustwidth}{2em}{0pt}
$\bullet$ Experiments demonstrate that Compass consistently outperforms SOTA methods under diverse missing modality conditions while maintaining lower computational overhead.
\end{adjustwidth}

\section{Related Works}
\subsection{Crack Segmentation Methods}

Deep learning dominates single-modal RGB crack segmentation \cite{Cheng2024Selective, Chen2024Mind, Benz2024Omni, Chu2024CrackGauGAN, Yang2020Feature, Jaziri2024Designing}, with recent works such as SCSegamba \cite{Liu2025SCSegamba} and Crackmer \cite{Wang2024Dual} advancing boundary sharpening and structure-aware detection. However, single-modal methods lack complementary cues and remain sensitive to challenging conditions.

In multimodal semantic segmentation, fusion methods span multiple paradigms. Channel level interaction methods such as CEN \cite{Wang2023Channel} and TokenFusion \cite{Wang2022Multimodal} exchange or replace tokens across modalities. Attention-based alignment methods such as CMX \cite{Zhang2023CMX}, CMNeXT \cite{Zhang2023Delivering}, and DFormerv2 \cite{Yin2025DFormerv2} achieve cross-modal semantic alignment. Encoder-sharing methods such as StitchFusion \cite{Li2025StitchFusion} and GminiFu \cite{Jia2024GeminiFusion} enable cross-modal information sharing during encoding. Frequency-domain methods such as FreqFusion \cite{Chen2024Frequency} address boundary displacement through adaptive filtering. However, all assume complete modality availability and lack explicit uncertainty quantification. Although DEF \cite{Huang2025Deep} introduces Dempster-Shafer theory for uncertainty modeling, it operates at the decision level without maintaining feature-level topological continuity, causing structural fragmentation for elongated targets such as cracks.

In missing modality handling, training augmentation methods such as random modality dropout \cite{Neverova2015ModDrop, Wang2023Multi} and MMP \cite{Nezakati2025MMP} enhance robustness but suffer gradient conflicts between semantic learning and degradation adaptation within a single stream \cite{Dai2024Study}. Feature completion methods such as KnowledgeBridger \cite{Ke2025Knowledge}, MCMoE \cite{Xu2026MCMoE}, and TouchFormer \cite{Lyu2026TouchFormer} passively compensate at inference with high overhead or collapse under extreme missing. Distribution alignment methods such as CCSD \cite{Xie2025CCSD} and SKD \cite{Sikdar2024SKD} force incomplete features toward complete distributions without explicitly modeling degradation conditions. No existing method decouples complete perception from degradation adaptation into dual-stream reciprocal learning, and missing modality research in crack segmentation remains nearly blank.

\vspace{-0.6cm}
\subsection{Receptance Weighted Key Value Model}

RWKV \cite{Peng2023RWKV} achieves sequence length independent linear complexity through recurrent state transfer, establishing itself as a promising foundation model following Transformers and Mamba \cite{Zhang2025L3TC, Sun2025RWKV3D, He2025RWKV}. VRWKV \cite{Duan2025Vision} introduces bidirectional modeling to eliminate the causal constraint and incorporates Q-Shift for local spatial perception. Subsequent works extend RWKV in vision along multiple dimensions. In scanning strategy, Dual-LoopLap RWKV \cite{Xie2025Learning}, RS-RWKV \cite{Zhou2025WKV}, and ZigRiR RWKV \cite{Chen2025Zig} improve sequence ordering through Laplace decomposition, random shuffling, and nested zigzag architectures. In state and channel modeling, URWKV \cite{Xu2025URWKV}, Restore-RWKV \cite{Yang2026Restore}, and CFDCNet \cite{Li2025Pan} enhance noise suppression, global dependency, and high-frequency recovery.

Despite these advances, existing variants universally rely on Q-Shift or similar fixed-pattern shifts for spatial perception, a rigid grid aligned bias ill-suited for cracks with random orientations and variable curvature. More critically, no existing work touches the WKV decay mechanism itself. The exponential decay applies a uniform rate across all tokens, unable to differentiate propagation needs along crack extensions from background directions. For crack segmentation, information should persist longer along crack orientations and decay faster perpendicularly, requiring directional morphological cues to be injected into the core WKV computation rather than merely altering peripheral scanning strategies.

\vspace{-0.5cm}
\section{Method}
\vspace{-0.1cm}
\subsection{Preliminary}

As shown in Figure~\ref{fig:Compass}, Compass consists of DSD, Needle, and ETPF. Given multimodal inputs ${X_m}_{m=1}^{N} \in \mathbb{R}^{H \times W \times C}$, an indicator $b_m \in {0,1}$ specifies whether the $m$-th modality is available, and missing modalities are replaced with random noise. Available images are first enhanced in the frequency domain and aggregated by cross-modal confidence weighting into a fused context, which is then fed into modality-specific Needle Blocks to extract four-scale feature maps. During training, features are split into an Original Stream and a Degradation Simulation Stream. In the original stream, all available features are completed by FAPT when needed and used to produce $P_o$. In the degradation simulation stream, one additional available modality is masked, then completed by FAPT to produce $P_d$, enabling reciprocal learning through multi-level distillation. During inference, only original stream is used, and the final prediction result $\hat{Y} \in \mathbb{R}^{H \times W \times 1}$ is obtained through ETPF.

\begin{figure}[!t]
  \centering
  \includegraphics[width=0.47\textwidth]{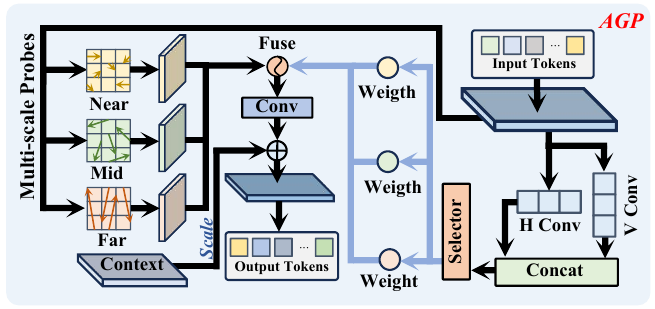}
  \caption{Architecture of AGP. AGP adaptively assigns each token suitable multi-scale spatial context based on anisotropy.}
  \label{fig:AGP}
  \vspace{-0.5cm}
\end{figure}

\vspace{-0.3cm}
\subsection{Degradation Simulation Distillation}
\label{sec:dsd}

Existing missing-modality methods couple semantic learning and degradation adaptation in a single stream, causing gradient conflict \cite{Nezakati2025MMP}. DSD decouples them through dual-stream reciprocal learning with explicit severe degradation. Each modality image $\mathbf{X}_m$ is enhanced by learnable frequency domain decomposition to obtain $\tilde{\mathbf{X}}_m$, and cross-modal confidence-weighted aggregation yields the fused context $\mathbf{c} = \sum_{m} \alpha_m \tilde{\mathbf{X}}_m / \sum_{m} \alpha_m$, where $\alpha_m$ denotes the confidence weight of modality $m$. The context $\mathbf{c}$ conditions the Needle and multimodal fusion. The enhanced images are then fed into modality-specific Needle Blocks to extract features $\{\mathbf{F}_m^{s}\}_{s=1}^{4}$.

In the dual-stream design, the original stream uses all available features, completes missing modalities via FAPT, and fuses all modal features to produce $\mathbf{P}_o$. The degradation simulation stream further masks one available modality from $\mathcal{I}_{avail}$ to create a harsher degradation setting. The degradation mask is defined as:

\vspace{-0.4cm}
\begin{equation}
\mathcal{M}_{deg} = \mathcal{I}_{avail} \setminus \{j\}, \quad j \sim \mathcal{U}\left(\mathcal{I}_{avail}\right), \quad \text{s.t.} \; |\mathcal{I}_{avail}| \geq 2,
\end{equation}
where $\mathcal{I}_{avail} = \{m \mid b_m = 1\}$ is the set of available modalities and $j$ is uniformly sampled. Requiring at least two available modalities ensures a reference signal. The masked features are completed by FAPT and fused to produce $\mathbf{P}_d$. 

To reconstruct semantically complete features without relying on modality-specific distributions, FAPT maintains $K$ learnable semantic prototypes $\mathcal{P} = \{\mathbf{p}_k\}_{k=1}^{K} \in \mathbb{R}^{K \times C}$ as modality-agnostic anchors. At each scale, available modal features are averaged to $\bar{\mathbf{F}}_{avail}$, and a spatial assignment module predicts per-pixel prototype weights for reconstruction:

\vspace{-0.5cm}
\begin{equation}
\mathbf{W} = \text{Softmax}\left(f_{assign}\left(\bar{\mathbf{F}}_{avail}\right)\right), \quad \hat{\mathbf{F}} = \sum_{k=1}^{K} \mathbf{W}_k \odot \mathbf{p}_k,
\end{equation}
where $f_{assign}$ is implemented with depthwise separable and pointwise convolutions, $\mathbf{W}_k \in \mathbb{R}^{B \times H \times W}$ is the spatial weight of prototype $k$, $\mathbf{p}_k \in \mathbb{R}^{C}$ is the prototype vector, and $\odot$ denotes element-wise multiplication. To recover fine spatial details, a gated residual refinement injects information from available modalities:

\vspace{-0.4cm}
\begin{equation}
\mathbf{F}_{out} = \hat{\mathbf{F}} + \sigma\left(f_{gate}\left(\mathbf{R}\right)\right) \odot \mathbf{R}, \quad \mathbf{R} = f_{refine}\left(\left[\hat{\mathbf{F}}; \bar{\mathbf{F}}_{avail}\right]\right),
\end{equation}
where $[\cdot;\cdot]$ denotes channel concatenation, $\mathbf{R}$ is the refined feature, $f_{gate}$ uses global average pooling and fully connected layers, and $\sigma$ is Sigmoid. This prototype-based design enables reasonable reconstruction even under extreme missing, as long as the available modalities provide sufficient semantic cues.

With predictions from both streams, reciprocal learning between $\mathbf{P}_o$ and $\mathbf{P}_d$ is enforced by three complementary losses. At the output level, $\mathcal{L}_{IMD}$ applies temperature-scaled bidirectional distillation, where forward distillation transfers soft labels from the original stream to guide the degradation stream, while backward distillation with a smaller weight $\alpha$ regularizes the original stream against overfitting to specific modality combinations:

\vspace{-0.4cm}
\begin{equation}
\mathcal{L}_{IMD} = \tau^2 \mathcal{H}\left(\sigma\left(\tfrac{\mathbf{P}_d}{\tau}\right), \sigma\left(\tfrac{\mathbf{P}_o^{\perp}}{\tau}\right)\right) + \alpha \tau^2 \mathcal{H}\left(\sigma\left(\tfrac{\mathbf{P}_o}{\tau}\right), \sigma\left(\tfrac{\mathbf{P}_d^{\perp}}{\tau}\right)\right),
\end{equation}
where $\tau$ is the temperature, $\alpha < 1$ is the backward weight, $\mathcal{H}$ is binary cross-entropy, and $\perp$ denotes gradient detach. Setting $\alpha < 1$ makes forward distillation dominant while preserving auxiliary backward feedback.

Beyond distributional consistency, crack segmentation also requires geometric consistency. At the structural level, $\mathcal{L}_{MGC}$ enforces both skeleton and edge alignment. For skeleton extraction, a soft erosion operator is defined as $\mathcal{E}(\cdot) \triangleq \text{NegMaxPool}(\cdot)$, which approximates morphological erosion by max pooling on the negated input followed by negation. Repeating it for $T$ steps with layer-wise differencing gives skeleton residuals:

\vspace{-0.4cm}
\begin{align}
\mathbf{S}_o^{(0)} &= \sigma\left(\mathbf{P}_o\right), \quad \mathbf{S}_d^{(0)} = \sigma\left(\mathbf{P}_d\right), \\
\mathbf{S}_o^{(t)} &= \mathbf{S}_o^{(t-1)} - \mathcal{E}\left(\mathbf{S}_o^{(t-1)}\right), \\
\mathbf{S}_d^{(t)} &= \mathbf{S}_d^{(t-1)} - \mathcal{E}\left(\mathbf{S}_d^{(t-1)}\right), \quad t = 1, \ldots, T
\end{align}
At the edge level, the Sobel operator computes gradient magnitude $\mathbf{M} = \sqrt{\mathbf{G}_x^2 + \mathbf{G}_y^2 + \epsilon}$ and normalized direction $\mathbf{n} = \left(\mathbf{G}_x / \mathbf{M}, \; \mathbf{G}_y / \mathbf{M}\right)$. $\mathcal{L}_{MGC}$ is defined as:

\vspace{-0.3cm}
\begin{equation}
\mathcal{L}_{MGC} = \left\|\mathbf{S}_d^{\left(T\right)} - \mathbf{S}_o^{\left(T\right)}\right\|_1 + \mathbb{E}\left[\left(1 - \langle \mathbf{n}_o, \mathbf{n}_d \rangle\right)^{+} \cdot \mathbf{M}_o\right],
\end{equation}
where $\langle \cdot, \cdot \rangle$ is inner product, $\left(\cdot\right)^{+} \triangleq \max\left(0, \cdot\right)$, and $\mathbb{E}$ denotes spatial expectation. The first term preserves crack topology, and the second enforces boundary orientation consistency.

To provide direct supervision in feature space, $\mathcal{L}_{FC}$ computes the MSE between the FAPT-completed features $\mathbf{F}^{comp}$ and the real Needle features $\mathbf{F}^{real}$ of the newly masked modality:

\vspace{-0.3cm}
\begin{equation}
\mathcal{L}_{FC} = \frac{1}{|\mathcal{S}|} \sum_{\left(m,s\right) \in \mathcal{S}} \left\|\mathbf{F}_{m,s}^{comp} - \mathbf{F}_{m,s}^{real\perp}\right\|_2^2,
\end{equation}
where $\mathcal{S}$ indexes the masked modality across scales and $\perp$ ensures that this loss updates only FAPT. 

Overall, DSD improves missing-modality robustness through proactive degradation simulation and reciprocal learning rather than passive augmentation.

\newcolumntype{G}{c}
\begin{table*}[!t]
    \centering
    \scriptsize
    \setlength{\tabcolsep}{2pt}
    \renewcommand{\arraystretch}{0.9}
    \caption{Performance comparison under dual-modal and multi missing settings. {\setlength{\fboxsep}{1pt}\colorbox{cyan!10}{\textbf{Blue}}} and {\setlength{\fboxsep}{1pt}\colorbox{orange!12}{\underline{Orange}}} denote best and second-best.}
    \resizebox{\textwidth}{!}{
    \begin{tabular}{lcc|*{9}{cc}}
    \toprule
    \multirow{2}{*}{\rotatebox{90}{}} & \multicolumn{2}{c|}{Missing Ratio}
    & \multicolumn{2}{c}{CMX \cite{Zhang2023CMX}} & \multicolumn{2}{c}{PrimKD\cite{Hao2024PrimKD}} & \multicolumn{2}{c}{Sigma \cite{Wan2025Sigma}}
    & \multicolumn{2}{c}{CMNeXT \cite{Zhang2023Delivering}} & \multicolumn{2}{c}{GminiFu\cite{Jia2024GeminiFusion}} & \multicolumn{2}{c}{PWRF \cite{Liu2025Part}}
    & \multicolumn{2}{c}{LIDAR \cite{Liu2025LIDAR}} & \multicolumn{2}{c}{VRWKV \cite{Duan2025Vision}} & \multicolumn{2}{c}{Compass} \\
    \midrule

    \multirow{12}{*}{\rotatebox{90}{IRTCrack \cite{liu2022asphalt}}}
        & RGB & Infrared & F1 & mIoU & F1 & mIoU & F1 & mIoU & F1 & mIoU & F1 & mIoU & F1 & mIoU & F1 & mIoU & F1 & mIoU & F1 & mIoU \\ \midrule
        & 10\% & Full & \cellcolor{orange!12}\underline{0.8330} & \cellcolor{orange!12}\underline{0.8414} & 0.8161 & 0.8327 & 0.8327 & 0.8344 & 0.8224 & 0.8311 & 0.7718 & 0.7885 & 0.8003 & 0.8224 & 0.8036 & 0.8235 & 0.8068 & 0.8173 & \cellcolor{cyan!10}\textbf{0.8496} & \cellcolor{cyan!10}\textbf{0.8503} \\
        & 30\% & Full & 0.7628 & \cellcolor{orange!12}\underline{0.8143} & 0.7720 & 0.8054 & \cellcolor{orange!12}\underline{0.7892} & 0.8040 & 0.7809 & 0.8116 & 0.7002 & 0.7673 & 0.7573 & 0.7977 & 0.7722 & 0.7847 & 0.7776 & 0.7952 & \cellcolor{cyan!10}\textbf{0.8218} & \cellcolor{cyan!10}\textbf{0.8438} \\
        & 50\% & Full & 0.6960 & 0.7750 & 0.7193 & 0.7639 & 0.7238 & 0.7749 & \cellcolor{orange!12}\underline{0.7371} & \cellcolor{orange!12}\underline{0.7837} & 0.6093 & 0.7184 & 0.7048 & 0.7740 & 0.7083 & 0.7676 & 0.7022 & 0.7828 & \cellcolor{cyan!10}\textbf{0.7642} & \cellcolor{cyan!10}\textbf{0.8191} \\
        & 70\% & Full & 0.6766 & 0.7257 & 0.6638 & 0.7384 & \cellcolor{orange!12}\underline{0.6942} & 0.7376 & 0.6408 & 0.7289 & 0.5509 & 0.6976 & 0.6536 & 0.7421 & 0.6516 & 0.7200 & 0.6627 & \cellcolor{cyan!10}\textbf{0.7703} & \cellcolor{cyan!10}\textbf{0.7044} & \cellcolor{orange!12}\underline{0.7618} \\
        & 90\% & Full & 0.5929 & 0.6940 & 0.5795 & 0.6932 & 0.6088 & 0.7023 & 0.6186 & 0.7081 & 0.5082 & 0.6488 & 0.6151 & 0.6958 & 0.6209 & 0.6956 & \cellcolor{orange!12}\underline{0.6361} & \cellcolor{cyan!10}\textbf{0.7598} & \cellcolor{cyan!10}\textbf{0.6667} & \cellcolor{orange!12}\underline{0.7260} \\
        & Full & 10\% & 0.8387 & 0.8458 & 0.8146 & 0.8389 & \cellcolor{orange!12}\underline{0.8426} & 0.8477 & 0.8302 & \cellcolor{orange!12}\underline{0.8481} & 0.7911 & 0.8237 & 0.8120 & 0.8381 & 0.8234 & 0.8280 & 0.8088 & 0.8222 & \cellcolor{cyan!10}\textbf{0.8685} & \cellcolor{cyan!10}\textbf{0.8656} \\
        & Full & 30\% & 0.8315 & \cellcolor{orange!12}\underline{0.8477} & 0.7931 & 0.8306 & \cellcolor{orange!12}\underline{0.8328} & 0.8341 & 0.8286 & 0.8452 & 0.7877 & 0.8214 & 0.8060 & 0.8351 & 0.8197 & 0.8249 & 0.8005 & 0.8119 & \cellcolor{cyan!10}\textbf{0.8580} & \cellcolor{cyan!10}\textbf{0.8591} \\
        & Full & 50\% & \cellcolor{orange!12}\underline{0.8289} & \cellcolor{orange!12}\underline{0.8441} & 0.7946 & 0.8298 & 0.8131 & 0.8284 & 0.8265 & 0.8388 & 0.7664 & 0.8103 & 0.7949 & 0.8292 & 0.8115 & 0.8165 & 0.7987 & 0.8148 & \cellcolor{cyan!10}\textbf{0.8551} & \cellcolor{cyan!10}\textbf{0.8578} \\
        & Full & 70\% & \cellcolor{orange!12}\underline{0.8327} & \cellcolor{orange!12}\underline{0.8456} & 0.7905 & 0.8278 & 0.8021 & 0.8130 & 0.8156 & 0.8399 & 0.7654 & 0.7996 & 0.7972 & 0.8274 & 0.8065 & 0.8137 & 0.7825 & 0.8160 & \cellcolor{cyan!10}\textbf{0.8509} & \cellcolor{cyan!10}\textbf{0.8571} \\
        & Full & 90\% & \cellcolor{orange!12}\underline{0.8245} & \cellcolor{orange!12}\underline{0.8437} & 0.7879 & 0.8306 & 0.7995 & 0.8240 & 0.8022 & 0.8210 & 0.7609 & 0.7980 & 0.7715 & 0.8056 & 0.7877 & 0.7969 & 0.7763 & 0.8075 & \cellcolor{cyan!10}\textbf{0.8489} & \cellcolor{cyan!10}\textbf{0.8524} \\
        & Full & Full & 0.8463 & 0.8518 & 0.8331 & 0.8447 & 0.8576 & 0.8509 & 0.8330 & 0.8466 & 0.8108 & 0.8214 & 0.8203 & 0.8383 & \cellcolor{orange!12}\underline{0.8580} & \cellcolor{orange!12}\underline{0.8538} & 0.8418 & 0.8453 & \cellcolor{cyan!10}\textbf{0.8696} & \cellcolor{cyan!10}\textbf{0.8628} \\
    \midrule

    \multirow{12}{*}{\rotatebox{90}{CrackDepth \cite{Liu2025LIDAR}}}
        & RGB & Depth & F1 & mIoU & F1 & mIoU & F1 & mIoU & F1 & mIoU & F1 & mIoU & F1 & mIoU & F1 & mIoU & F1 & mIoU & F1 & mIoU \\ \midrule
        & 10\% & Full & 0.7127 & 0.7890 & 0.7770 & \cellcolor{orange!12}\underline{0.8235} & 0.7274 & 0.7838 & 0.7705 & 0.8178 & 0.6988 & 0.7787 & 0.7448 & 0.8050 & 0.7762 & 0.8164 & \cellcolor{orange!12}\underline{0.7782} & 0.8162 & \cellcolor{cyan!10}\textbf{0.8189} & \cellcolor{cyan!10}\textbf{0.8459} \\
        & 30\% & Full & 0.7039 & 0.7658 & 0.7077 & 0.7738 & 0.7164 & 0.7812 & \cellcolor{orange!12}\underline{0.7166} & \cellcolor{orange!12}\underline{0.7856} & 0.6471 & 0.7553 & 0.6876 & 0.7668 & 0.6646 & 0.7338 & 0.7137 & 0.7773 & \cellcolor{cyan!10}\textbf{0.7673} & \cellcolor{cyan!10}\textbf{0.8098} \\
        & 50\% & Full & 0.6475 & 0.7485 & \cellcolor{orange!12}\underline{0.6494} & \cellcolor{orange!12}\underline{0.7498} & 0.5475 & 0.7364 & 0.6248 & 0.7437 & 0.5303 & 0.7024 & 0.6198 & 0.7312 & 0.5424 & 0.7188 & 0.6281 & 0.7345 & \cellcolor{cyan!10}\textbf{0.6797} & \cellcolor{cyan!10}\textbf{0.7621} \\
        & 70\% & Full & \cellcolor{orange!12}\underline{0.5439} & 0.6914 & 0.5348 & 0.6886 & 0.4995 & 0.6744 & 0.5139 & 0.6712 & 0.1668 & 0.5364 & 0.5348 & 0.6808 & 0.4557 & 0.6170 & 0.5331 & \cellcolor{orange!12}\underline{0.6938} & \cellcolor{cyan!10}\textbf{0.6407} & \cellcolor{cyan!10}\textbf{0.7546} \\
        & 90\% & Full & 0.4304 & 0.6355 & 0.0641 & 0.3472 & 0.0570 & 0.0584 & 0.4046 & 0.6227 & 0.1095 & 0.5136 & 0.4046 & 0.6233 & 0.2445 & 0.5526 & \cellcolor{orange!12}\underline{0.4852} & \cellcolor{orange!12}\underline{0.6460} & \cellcolor{cyan!10}\textbf{0.5418} & \cellcolor{cyan!10}\textbf{0.6809} \\
        & Full & 10\% & 0.8003 & 0.8408 & 0.7944 & 0.8363 & \cellcolor{orange!12}\underline{0.8082} & \cellcolor{orange!12}\underline{0.8426} & 0.7953 & 0.8369 & 0.7519 & 0.8218 & 0.7682 & 0.8198 & 0.7915 & 0.8292 & 0.8079 & 0.8343 & \cellcolor{cyan!10}\textbf{0.8303} & \cellcolor{cyan!10}\textbf{0.8500} \\
        & Full & 30\% & 0.7939 & \cellcolor{orange!12}\underline{0.8386} & 0.7905 & 0.8299 & \cellcolor{orange!12}\underline{0.7971} & 0.8380 & 0.7854 & 0.8286 & 0.7570 & 0.8175 & 0.7603 & 0.8166 & 0.7802 & 0.7924 & 0.7897 & 0.8206 & \cellcolor{cyan!10}\textbf{0.8288} & \cellcolor{cyan!10}\textbf{0.8518} \\
        & Full & 50\% & 0.7881 & \cellcolor{orange!12}\underline{0.8367} & 0.7862 & 0.8294 & \cellcolor{orange!12}\underline{0.7929} & 0.8291 & 0.7848 & 0.8257 & 0.6889 & 0.7984 & 0.7497 & 0.8126 & 0.7607 & 0.8114 & 0.7782 & 0.8162 & \cellcolor{cyan!10}\textbf{0.8260} & \cellcolor{cyan!10}\textbf{0.8480} \\
        & Full & 70\% & \cellcolor{orange!12}\underline{0.7896} & \cellcolor{orange!12}\underline{0.8336} & 0.7894 & 0.8300 & 0.7855 & 0.8287 & 0.7641 & 0.8165 & 0.6224 & 0.7306 & 0.7481 & 0.8098 & 0.7570 & 0.8029 & 0.7567 & 0.8053 & \cellcolor{cyan!10}\textbf{0.8232} & \cellcolor{cyan!10}\textbf{0.8447} \\
        & Full & 90\% & 0.7712 & \cellcolor{orange!12}\underline{0.8283} & \cellcolor{orange!12}\underline{0.7733} & 0.8278 & 0.7517 & 0.8184 & 0.7482 & 0.8157 & 0.5482 & 0.7025 & 0.7203 & 0.8023 & 0.7276 & 0.7861 & 0.7329 & 0.7963 & \cellcolor{cyan!10}\textbf{0.8216} & \cellcolor{cyan!10}\textbf{0.8434} \\
        & Full & Full & 0.8109 & 0.8410 & 0.7989 & 0.8338 & 0.8172 & 0.8445 & 0.8043 & 0.8407 & 0.7887 & 0.8261 & 0.7795 & 0.8210 & \cellcolor{orange!12}\underline{0.8180} & \cellcolor{orange!12}\underline{0.8453} & 0.8174 & 0.8422 & \cellcolor{cyan!10}\textbf{0.8343} & \cellcolor{cyan!10}\textbf{0.8544} \\
    \bottomrule
    \end{tabular}
    }
    \label{tab:double_modals}
    \vspace{-0.1cm}
\end{table*}

\begin{figure*}[!t]
  \centering
  \includegraphics[width=\textwidth]{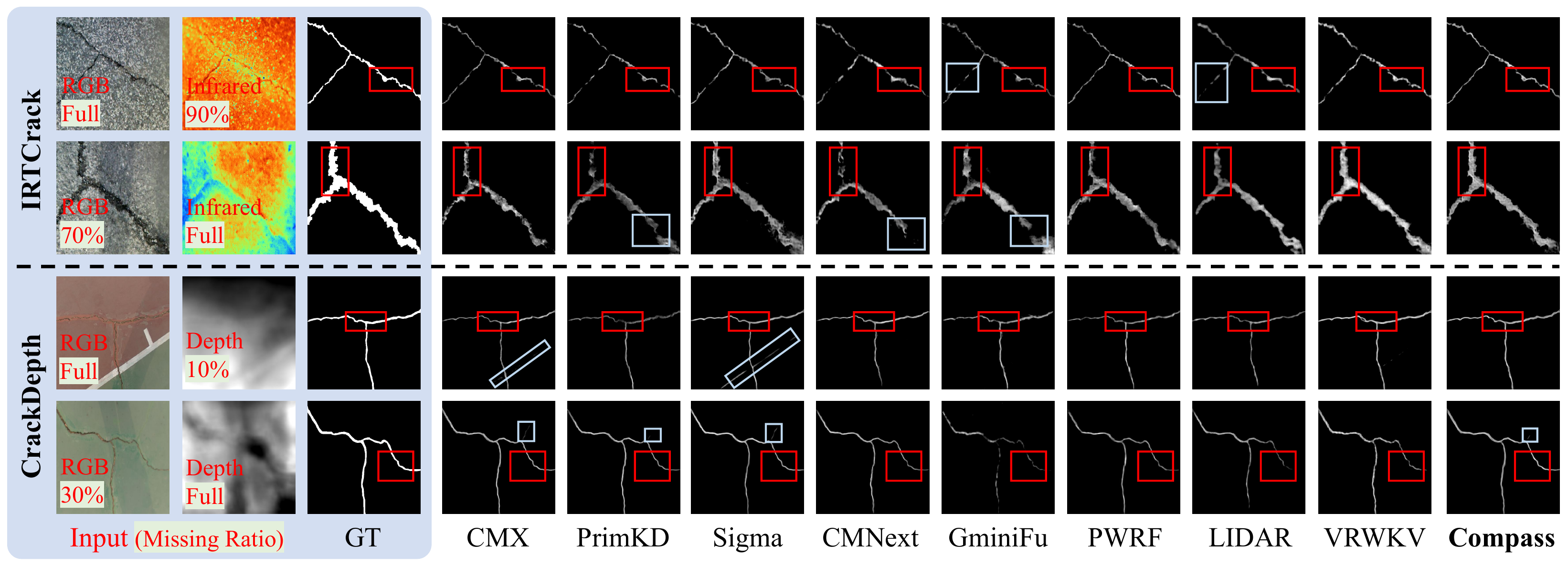}
  \caption{Visual comparison under dual-modal missing settings. Red boxes denote key regions; blue denote misidentifications.}
  \label{fig:dual_results}
  \vspace{-0.3cm}
\end{figure*}

\vspace{-0.2cm}
\subsection{Geometry-Conditioned Needle Block}

The WKV mechanism achieves linear complexity through recurrent state transfer. At step $t$, WKV maintains the accumulated state $\mathbf{a}_t \in \mathbb{R}^{C}$ and normalization factor $\mathbf{b}_t \in \mathbb{R}^{C}$:

\vspace{-0.3cm}
\begin{equation}
\mathbf{wkv}_t = \frac{e^{-\mathbf{w}} \odot \mathbf{a}_{t-1} + e^{\mathbf{k}_t} \odot \mathbf{v}_t + e^{\mathbf{u} + \mathbf{k}_t} \odot \mathbf{v}_t}{e^{-\mathbf{w}} \odot \mathbf{b}_{t-1} + e^{\mathbf{k}_t}},
\end{equation}
where $\mathbf{w} \in \mathbb{R}^{C}$ is the channel-wise decay vector, $\mathbf{u} \in \mathbb{R}^{C}$ is the learnable gain for the current token, $\mathbf{k}_t, \mathbf{v}_t \in \mathbb{R}^{C}$ are Key and Value vectors, and $\odot$ denotes Hadamard product.

Since $\mathbf{w}$ is shared across tokens, standard WKV cannot distinguish propagation along crack and background directions. GCM is introduced to address this limitation. 

GCM injects crack orientation into the Key vector $\mathbf{k}_t$ to enable direction-aware propagation. From the cross-modal context $\mathbf{c}_{aligned}$, a lightweight predictor $f_{dir}$ estimates the crack direction unit vector $\mathbf{d} = (d_x, d_y) = \mathrm{Norm}(f_{dir}(\mathbf{c}_{aligned}))$. Under sequential token traversal, Needle performs directional scanning, and modulates the original Key by the squared projection between the crack direction and each scan direction before the WKV update:

\vspace{-0.3cm}
\begin{equation}
\mathbf{k}'_h = \mathbf{k}_t \cdot \left(1 + \sigma\left(\gamma\right) \cdot d_x^2\right), \quad \mathbf{k}'_v = \mathbf{k}_t \cdot \left(1 + \sigma\left(\gamma\right) \cdot d_y^2\right),
\end{equation}
where $\gamma$ is a learnable modulation strength initialized to zero, ensuring equivalence to standard WKV at the start of training. The modulated Keys replace $\mathbf{k}_t$ in updating $\mathbf{a}_t$ and $\mathbf{b}_t$, while $\mathbf{v}_t$ and $\mathbf{w}$ remain unchanged. The two directional outputs are then averaged:

\vspace{-0.2cm}
\begin{equation}
\mathbf{rwkv} = \frac{1}{2}\left(\text{WKV}\left(\mathbf{k}'_h, \mathbf{v}_t\right) + \text{WKV}\left(\mathbf{k}'_v, \mathbf{v}_t\right)\right),
\end{equation}

This mechanism can be further interpreted by the contribution of token $\tau$ to the accumulated state at step $t$. In standard WKV, its unnormalized weight is $e^{-(t-1-\tau)\mathbf{w} + \mathbf{k}_\tau}$. GCM modulates the Key as $\mathbf{k}'_\tau = \mathbf{k}_\tau(1 + \delta a_\tau)$, where $\delta = \sigma(\gamma)$ and $a_\tau \in [0,1]$ denotes directional alignment, giving:

\vspace{-0.2cm}
\begin{equation}
\hat{\mathbf{w}}_\tau^{gcm} \propto e^{-(t-1-\tau)\mathbf{w} + \mathbf{k}_\tau + \delta a_\tau \mathbf{k}_\tau} = \hat{\mathbf{w}}_\tau^{std} \cdot e^{\delta a_\tau \mathbf{k}_\tau},
\end{equation}
where $\hat{\mathbf{w}}_\tau^{std}$ is the standard WKV contribution weight. This introduces a salience gain $\mathbf{\Gamma}_\tau = e^{\delta a_\tau \mathbf{k}_\tau}$, which amplifies crack-aligned tokens with $a_\tau \approx 1$ and active $\mathbf{k}_\tau > 0$, while preserving standard decay for background tokens with $a_\tau \approx 0$. Thus, GCM achieves direction-adaptive propagation without modifying the global decay $\mathbf{w}$ or Value $\mathbf{v}_t$, and bidirectional fusion ensures coverage of cracks with arbitrary orientations.

To inject cross-modal information, the fused context is projected as a residual signal onto the WKV output, while the receptance gate $\sigma(\mathbf{r})$ restricts the injection to semantically active tokens:

\vspace{-0.4cm}
\begin{equation}
\mathbf{rwkv} \leftarrow \mathbf{rwkv} + \sigma\left(\alpha_{cm}\right) \cdot \tanh\left(\mathbf{W}_{cm} \mathbf{c}_{aligned}\right) \odot \sigma\left(\mathbf{r}\right),
\end{equation}
where $\mathbf{W}_{cm}$ is the projection matrix and $\alpha_{cm}$ is a learnable injection strength initialized to zero to avoid interference in early training.

\begin{table}[!t]
    \centering

    \setlength{\tabcolsep}{7pt}
    \renewcommand{\arraystretch}{0.95}
    \caption{Complexity comparison of dual-modal input.}
    \resizebox{\linewidth}{!}{
    \begin{tabular}{ccccc}
    \hline
    Method & Year & FLOPs$\downarrow$ & Params$\downarrow$ & Size$\downarrow$ \\ \hline
    CMX \cite{Zhang2023CMX} & TITS 2023 & 56.99G & 66.56M & 762MB \\
    CMNeXT \cite{Zhang2023Delivering} & CVPR 2023 & 44.76G & 57.63M & 660MB \\
    PrimKD \cite{Hao2024PrimKD} & TGRS 2024 & 679.09G & 615.64M & 7322MB \\
    GminiFu \cite{Jia2024GeminiFusion} & ICML 2024 & 45.53G & 29.19M & 472MB \\
    PWRF \cite{Liu2025Part} & IJCV 2025 & 185.14G & 252.36M & 1024MB \\
    Sigma \cite{Wan2025Sigma} & WACV 2025 & 63.43G & 41.68M & 553MB \\
    LIDAR \cite{Liu2025LIDAR} & MM 2025 & \cellcolor{orange!12}\underline{33.33G} & \cellcolor{orange!12}\underline{5.35M} & \cellcolor{orange!12}\underline{78MB} \\
    VRWKV \cite{Duan2025Vision} & ICLR 2025 & 151.99G & 10.22M & 129MB \\
    \textbf{Compass (Ours)} & \textbf{MM 2026} & \cellcolor{cyan!10}\textbf{29.38G} & \cellcolor{cyan!10}\textbf{2.58M} & \cellcolor{cyan!10}\textbf{58MB} \\ \hline
    \end{tabular}
    }
    \label{tab:Complex_dual_modals}
    \vspace{-0.3cm}
\end{table}

Beyond token-level modulation, preserving structural coherence across neighboring tokens is also critical for crack segmentation. The Coherence Gate introduces structure-aware gating over a $3 \times 3$ neighborhood, encouraging co-activation in connected crack regions while suppressing isolated noise:

\vspace{-0.3cm}
\begin{equation}
\mathbf{g} = \mathbf{g}_{self} \cdot \left(\left(1 - \lambda\right) + \lambda \cdot \mathbf{g}_{nb}\right), \quad \lambda = \sigma\left(\lambda_{raw}\right),
\end{equation}
where $\mathbf{g}_{self}$ is self-response gating, $\mathbf{g}_{nb} = \sigma(\text{DWConv}_{3 \times 3}(\sigma(\mathbf{W}_n \mathbf{x})))$ is the neighborhood consistency signal, $\mathbf{x}$ is the input feature map of the current Needle block, and $\lambda_{raw}$ is initialized to zero so that training starts from pure self-response gating.

To equip each token with suitable spatial context, as shown in \ref{fig:AGP}, the Anisotropic Guided Probe employs three groups of dilated depthwise convolutions for near, mid, and far range receptive fields. Local anisotropy, captured by horizontal $1 \times 3$ and vertical $3 \times 1$ convolutions, guides scale selection:

\vspace{-0.3cm}
\begin{equation}
\mathbf{F}_{out} = \mathbf{x} + \sigma\left(\beta\right) \cdot f_{out}\left(\textstyle\sum_{i} \pi_i \mathbf{q}_i\right),
\end{equation}
where $\boldsymbol{\pi} = \text{Softmax}(f_{sel}([\mathbf{r}_h; \mathbf{r}_v]))$ are the normalized scale-selection weights, $\mathbf{q}_i$ is the $i$-th probe output, $\mathbf{r}_h$ and $\mathbf{r}_v$ are directional response magnitudes, and $\beta$ is a learnable residual scale.

\vspace{-0.2cm}
\subsection{Evidential Topology-Preserving Fusion}

Under missing modalities, multimodal fusion must both assess modal reliability for adaptive weighting and preserve crack topology. ETPF addresses this with Dempster-Shafer evidence theory, anisotropic evidence propagation, and uncertainty-gated decoding.

At each scale, enhanced features are projected to non-negative Dirichlet evidence, from which belief and uncertainty are derived:

\vspace{-0.3cm}
\begin{equation}
\begin{aligned}
&\mathbf{e} = \text{Softplus}\left(f_{proj}\left(\mathbf{F}\right)\right), \quad \boldsymbol{\alpha} = \mathbf{e} + \mathbf{1}, \\
&\mathbf{Z} = \sum_{k=1}^{K} \boldsymbol{\alpha}_k, \quad \mathbf{b}_k = \mathbf{e}_k / \mathbf{Z}, \quad \mathbf{u} = K / \mathbf{Z},
\end{aligned}
\end{equation}
where $\mathbf{e} \in \mathbb{R}^{B \times K \times H \times W}$ is non-negative evidence, $\mathbf{Z}$ is Dirichlet strength, $K$ is the class number, $\mathbf{b}_k$ is the belief of class $k$, and $\mathbf{u}$ denotes uncertainty.

Given per-modality beliefs and uncertainties, multimodal evidence is fused by the Dempster-Shafer rule (DS):

\vspace{-0.2cm}
\begin{equation}
\mathbf{b}_c = \frac{\mathbf{b}_1 \odot \mathbf{b}_2 + \mathbf{b}_1 \odot \mathbf{u}_2 + \mathbf{u}_1 \odot \mathbf{b}_2}{1 - \mathbf{C}}, \quad \mathbf{u}_c = \frac{\mathbf{u}_1 \odot \mathbf{u}_2}{1 - \mathbf{C}},
\end{equation}
where $\mathbf{C} = \langle \mathbf{b}_1, \mathbf{1} \rangle \langle \mathbf{b}_2, \mathbf{1} \rangle - \langle \mathbf{b}_1, \mathbf{b}_2 \rangle$ is the conflict term. To avoid instability under large conflict in early training, a learnable blend gate $\eta = \sigma(\eta_{raw})$ combines the DS result $\mathbf{b}_{DS}$ with simple evidence summation $\mathbf{b}_{sum}$:

\vspace{-0.3cm}
\begin{equation}
\mathbf{b}_{fused} = \eta \cdot \mathbf{b}_{DS} + (1 - \eta) \cdot \mathbf{b}_{sum}, \quad \mathbf{u}_{fused} = \eta \cdot \mathbf{u}_{DS} + (1 - \eta) \cdot \mathbf{u}_{sum},
\end{equation}
where $\eta_{raw}$ is initialized to zero to favor the stable summation path before gradually shifting to fusion.

Although fused beliefs improve reliability, they may still be discontinuous along crack structures. Anisotropic evidence propagation alleviates this by estimating crack orientation with directional convolutions and propagating belief accordingly, using direction weights $\boldsymbol{\rho}_h = |\mathbf{r}_h| / (|\mathbf{r}_h| + |\mathbf{r}_v| + \epsilon)$ and $\boldsymbol{\rho}_v = |\mathbf{r}_v| / (|\mathbf{r}_h| + |\mathbf{r}_v| + \epsilon)$:

\vspace{-0.3cm}
\begin{equation}
\mathbf{b}' = \mathbf{b} + \sigma\left(\gamma_p\right) \cdot \left(\boldsymbol{\rho}_h \odot \mathcal{K}_h\left(\mathbf{b}\right) + \boldsymbol{\rho}_v \odot \mathcal{K}_v\left(\mathbf{b}\right) + \mathcal{K}_{loc}\left(\mathbf{b}\right) - \mathbf{b}\right),
\end{equation}
where $\mathcal{K}_h$, $\mathcal{K}_v$, and $\mathcal{K}_{loc}$ are $1 \times 7$, $7 \times 1$, and $3 \times 3$ convolutions, and $\gamma_p$ is a learnable propagation strength. To enforce cross-scale consistency, upsampled coarse-scale beliefs $\hat{\mathbf{b}}_{coarse} = \text{Up}(\mathbf{b}_{coarse})$ are used as priors for fine-scale beliefs:

\vspace{-0.4cm}
\begin{equation}
\mathbf{b}_{fine}' = \left(1 - \omega\right) \mathbf{b}_{fine} + \omega \, \mathbf{b}_{fine} \odot \hat{\mathbf{b}}_{coarse} + \xi \omega \, \hat{\mathbf{b}}_{coarse},
\end{equation}
where $\omega = \sigma(\omega_{raw})$ is a learnable consistency weight and $\xi$ is the prior injection strength.

During decoding, skip connections are modulated by uncertainty gating to suppress unreliable features:

\vspace{-0.3cm}
\begin{equation}
\mathbf{F}'_{skip} = \mathbf{F}_{skip} \odot \sigma\left(\mathbf{W}_{proj}\left(\mathbf{1} - \mathbf{u}\right)\right),
\end{equation}
where $\mathbf{W}_{proj}$ is a $1 \times 1$ convolution that projects single-channel uncertainty to the channel dimension of $\mathbf{F}_{skip}$. Overall, ETPF forms a closed loop from evidential modeling to anisotropic topological propagation and uncertainty-gated decoding, balancing fusion reliability and crack topology under missing modalities.

\begin{table*}[!t]
    \centering

    \setlength{\heavyrulewidth}{0.5pt}
    \setlength{\lightrulewidth}{0.3pt}
    \setlength{\cmidrulewidth}{0.2pt}

    \setlength{\tabcolsep}{7pt}
    \renewcommand{\arraystretch}{0.88}
    \caption{Performance comparison under tri-modal and multi missing settings. {\setlength{\fboxsep}{1pt}\colorbox{cyan!10}{\textbf{Blue}}} and {\setlength{\fboxsep}{1pt}\colorbox{orange!12}{\underline{Orange}}} denote best and second-best.}
    \resizebox{\linewidth}{!}{
    \begin{tabular}{ccc|*{7}{cc}}
    \toprule
    \multicolumn{3}{c|}{Missing Ratio}
      & \multicolumn{2}{c}{GminiFusion \cite{Jia2024GeminiFusion}} & \multicolumn{2}{c}{mmsFormer \cite{Reza2024MMSFormer}} & \multicolumn{2}{c}{PWRF \cite{Liu2025Part}} & \multicolumn{2}{c}{CMNeXT \cite{Zhang2023Delivering}} & \multicolumn{2}{c}{LIDAR \cite{Liu2025LIDAR}} & \multicolumn{2}{c}{VRWKV \cite{Duan2025Vision}} & \multicolumn{2}{c}{\textbf{Compass}} \\
    \midrule
    RGB & AoP & DoP & F1 & mIoU & F1 & mIoU & F1 & mIoU & F1 & mIoU & F1 & mIoU & F1 & mIoU & F1 & mIoU \\
    \midrule
    Full & 10\% & 30\% & 0.6766 & 0.7706 & 0.7191 & 0.7912 & 0.7056 & 0.7823 & 0.7288 & \cellcolor{orange!12}\underline{0.7961} & 0.7179 & 0.7891 & \cellcolor{orange!12}\underline{0.7349} & 0.7958 & \cellcolor{cyan!10}\textbf{0.7441} & \cellcolor{cyan!10}\textbf{0.8054} \\
    30\% & Full & 10\% & 0.5836 & 0.7012 & 0.7056 & \cellcolor{orange!12}\underline{0.7896} & 0.6904 & 0.7782 & 0.7091 & 0.7889 & 0.6948 & 0.7708 & \cellcolor{orange!12}\underline{0.7195} & 0.7882 & \cellcolor{cyan!10}\textbf{0.7344} & \cellcolor{cyan!10}\textbf{0.8005} \\
    10\% & 30\% & Full & 0.6997 & 0.7816 & 0.7104 & 0.7904 & 0.7035 & 0.7794 & \cellcolor{orange!12}\underline{0.7154} & \cellcolor{orange!12}\underline{0.7913} & 0.6903 & 0.7549 & 0.6992 & 0.7794 & \cellcolor{cyan!10}\textbf{0.7499} & \cellcolor{cyan!10}\textbf{0.8042} \\
    Full & 70\% & 90\% & 0.6699 & 0.7653 & 0.7043 & 0.7897 & 0.6927 & 0.7804 & \cellcolor{orange!12}\underline{0.7111} & \cellcolor{orange!12}\underline{0.7927} & 0.6984 & 0.7784 & 0.6760 & 0.7640 & \cellcolor{cyan!10}\textbf{0.7404} & \cellcolor{cyan!10}\textbf{0.8019} \\
    90\% & Full & 70\% & 0.3003 & 0.5927 & 0.6410 & 0.7490 & 0.6078 & 0.7282 & \cellcolor{orange!12}\underline{0.6795} & \cellcolor{orange!12}\underline{0.7691} & 0.5964 & 0.7165 & 0.5695 & 0.7333 & \cellcolor{cyan!10}\textbf{0.7280} & \cellcolor{cyan!10}\textbf{0.7958} \\
    70\% & 90\% & Full & 0.5412 & 0.7051 & \cellcolor{orange!12}\underline{0.6410} & \cellcolor{orange!12}\underline{0.7490} & 0.6404 & 0.7373 & 0.5372 & 0.6897 & 0.5846 & 0.7135 & 0.5788 & 0.7395 & \cellcolor{cyan!10}\textbf{0.6956} & \cellcolor{cyan!10}\textbf{0.7800} \\
    10\% & 30\% & 70\% & 0.6187 & 0.7439 & 0.6834 & 0.7628 & 0.6670 & 0.7704 & \cellcolor{orange!12}\underline{0.6988} & \cellcolor{orange!12}\underline{0.7805} & 0.6868 & 0.7622 & 0.6829 & 0.7670 & \cellcolor{cyan!10}\textbf{0.7426} & \cellcolor{cyan!10}\textbf{0.7977} \\
    30\% & 70\% & 10\% & 0.5728 & 0.7163 & 0.6696 & 0.7692 & 0.6290 & 0.7493 & 0.6935 & 0.7795 & 0.5810 & 0.6978 & \cellcolor{orange!12}\underline{0.6957} & \cellcolor{orange!12}\underline{0.7842} & \cellcolor{cyan!10}\textbf{0.7284} & \cellcolor{cyan!10}\textbf{0.7923} \\
    70\% & 10\% & 30\% & 0.4041 & 0.6722 & 0.6899 & 0.7721 & 0.6612 & 0.7677 & \cellcolor{orange!12}\underline{0.6900} & \cellcolor{orange!12}\underline{0.7788} & 0.6261 & 0.7077 & 0.6810 & 0.7761 & \cellcolor{cyan!10}\textbf{0.7221} & \cellcolor{cyan!10}\textbf{0.7903} \\
    30\% & 70\% & 90\% & 0.1938 & 0.6302 & 0.5408 & \cellcolor{orange!12}\underline{0.7234} & 0.2025 & 0.6679 & \cellcolor{orange!12}\underline{0.5731} & 0.7153 & 0.4555 & 0.7034 & 0.5392 & 0.7233 & \cellcolor{cyan!10}\textbf{0.6791} & \cellcolor{cyan!10}\textbf{0.7427} \\
    70\% & 90\% & 30\% & 0.2249 & 0.5682 & 0.1958 & 0.6503 & 0.2226 & 0.6585 & 0.4863 & 0.6965 & 0.4164 & 0.6243 & \cellcolor{orange!12}\underline{0.5693} & \cellcolor{orange!12}\underline{0.7360} & \cellcolor{cyan!10}\textbf{0.6364} & \cellcolor{cyan!10}\textbf{0.7368} \\
    90\% & 30\% & 70\% & 0.5001 & 0.6754 & 0.5066 & 0.6690 & 0.2207 & 0.6649 & \cellcolor{orange!12}\underline{0.5922} & 0.7130 & 0.3716 & 0.5885 & 0.5392 & \cellcolor{orange!12}\underline{0.7233} & \cellcolor{cyan!10}\textbf{0.6240} & \cellcolor{cyan!10}\textbf{0.7234} \\
    Full & Full & Full & 0.6920 & 0.7794 & 0.7265 & 0.7952 & 0.7063 & 0.7837 & 0.7124 & 0.7875 & \cellcolor{orange!12}\underline{0.7353} & \cellcolor{orange!12}\underline{0.7976} & 0.7291 & 0.7919 & \cellcolor{cyan!10}\textbf{0.7503} & \cellcolor{cyan!10}\textbf{0.8055} \\
    \bottomrule
    \end{tabular}
    }
    \label{tab:tri_modals}
    \vspace{-0.4cm}
\end{table*}

\begin{figure*}[!t]
  \centering
  \includegraphics[width=\textwidth]{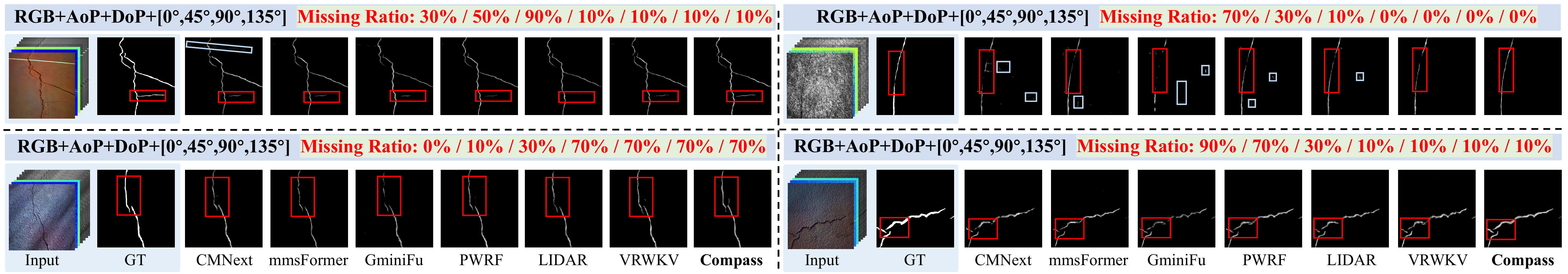}
  \caption{Visual comparison under tri-modal missing settings. Red boxes denote key regions; blue denote misidentifications.}
  \label{fig:tri_results}
  \vspace{-0.2cm}
\end{figure*}

\vspace{-0.3cm}
\subsection{Objective Function}

The segmentation loss $\mathcal{L}_{seg}$ is the sum of binary cross-entropy and Dice loss \cite{Liu2025SCSegamba}. The total objective function is:

\vspace{-0.3cm}
\begin{equation}
\mathcal{L} = \mathcal{L}_{seg}^{o} + \mathcal{L}_{seg}^{d} + \mathcal{L}_{IMD} + \mathcal{L}_{MGC} + \mathcal{L}_{FC},
\end{equation}
where $\mathcal{L}_{seg}^{o}$ and $\mathcal{L}_{seg}^{d}$ are the segmentation losses of the original and degradation simulation streams.

\vspace{-0.3cm}
\section{Experiments}
\vspace{-0.1cm}
\subsection{Datasets}

\noindent \textbf{Multimodal Crack Datasets.} CrackDepth \cite{Liu2025LIDAR}, CrackPolar \cite{Liu2025LIDAR}, and IRTCrack \cite{liu2022asphalt} contain 656 RGB-Depth pairs, 986 RGB-polarization samples, and 448 RGB-Thermal pairs, respectively. Their complementary depth, polarization, and thermal cues enhance structural perception and improve crack boundary detection under complex illumination, varying pavement surfaces, and low-visibility conditions.




\begin{table}[!t]
\vspace{-0.3cm}
    \centering
    \setlength{\tabcolsep}{7pt}
    \renewcommand{\arraystretch}{0.95}
    \caption{Complexity comparison of tri-modal input.}
    \resizebox{\linewidth}{!}{
    \begin{tabular}{ccccc}
    \hline
    Method & Year & FLOPs$\downarrow$ & Params$\downarrow$ & Size$\downarrow$ \\ \hline
    CMNeXT \cite{Zhang2023Delivering} & CVPR 2023 & \cellcolor{orange!12}\underline{45.99G} & 57.64M & 693MB \\
    mmsFormer \cite{Reza2024MMSFormer} & OJSP 2024 & 59.28G & 33.31M & 435MB \\
    GminiFu \cite{Jia2024GeminiFusion} & ICML 2024 & 46.79G & 29.22M & 494MB \\
    PWRF \cite{Liu2025Part} & IJCV 2025 & 238.42G & 344.02M & 3789MB \\
    LIDAR \cite{Liu2025LIDAR} & MM 2025 & 50.00G & \cellcolor{orange!12}\underline{8.02M} & \cellcolor{orange!12}\underline{150MB} \\
    VRWKV \cite{Duan2025Vision} & ICLR 2025 & 188.86G & 14.74M & 187MB \\
    \textbf{Compass (Ours)} & \textbf{MM 2026} & \cellcolor{cyan!10}\textbf{42.82G} & \cellcolor{cyan!10}\textbf{3.80M} & \cellcolor{cyan!10}\textbf{58MB} \\ \hline
    \end{tabular}
    }
    \label{tab:Complex_tri_modals}
    \vspace{-0.3cm}
\end{table}

\begin{figure}[!t]
  \centering
  \includegraphics[width=0.47\textwidth]{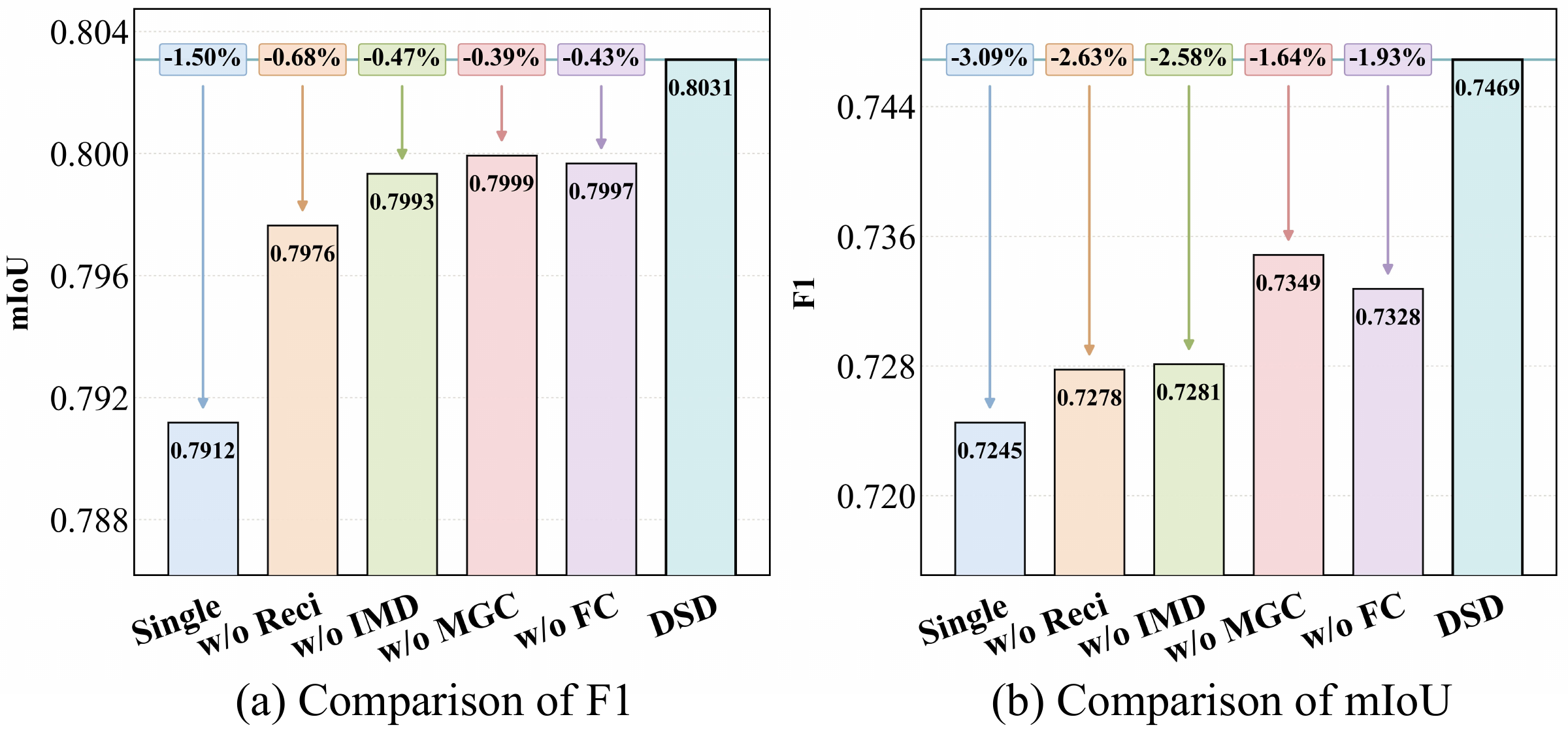}
    \caption{Comparison of DSD variants. Single represents single stream, Reci represents reciprocal learning, IMD represents $\mathcal{L}_{IMD}$, MGC represents $\mathcal{L}_{MGC}$, and FC represents $\mathcal{L}_{FC}$.}
  \label{fig:abl_distill}
  \vspace{-0.4cm}
\end{figure}

\vspace{-0.3cm}
\subsection{Implementation Details}
\vspace{-0.1cm}

\noindent \textbf{Experimental Settings:} Compass is implemented in PyTorch 2.9.1 and trained on a server with an Intel Xeon Platinum 8558 CPU, an NVIDIA L20 GPU, and Ubuntu 22.04.5. We use AdamW with an initial learning rate of $2\times10^{-3}$, weight decay of 0.01. Training runs for 80 epochs, including a 5 epoch warmup stage. 

\noindent \textbf{Evaluation Metrics:} Performance is evaluated using F1 score and mIoU, while computational complexity is measured by Params, FLOPs, and Model Size \cite{Liu2025LIDAR}.

\vspace{-0.3cm}
\subsection{Comparison with SOTA Methods}

\noindent \textbf{Comparative Analysis on Dual-Modal Datasets.} As shown in   Table~\ref{tab:double_modals} and Figure~\ref{fig:dual_results}, on IRTCrack, with 50\% RGB missing, Compass improves F1 and mIoU by 3.68\% and 4.52\% over CMNeXT~\cite{Zhang2023Delivering}; with 90\% RGB missing, it surpasses VRWKV~\cite{Duan2025Vision} by 4.82\% in F1; with 90\% infrared missing, it exceeds CMX~\cite{Zhang2023CMX} by 2.95\% in F1. Under 90\% infrared missing, CMX~\cite{Zhang2023CMX} and PrimKD~\cite{Hao2024PrimKD} suffer from crack fragmentation, while Sigma~\cite{Wan2025Sigma} and CMNeXT~\cite{Zhang2023Delivering} produce false positives, whereas Compass restores complete crack structures, validating the effectiveness of DSD. Under full modalities, Compass still outperforms LIDAR~\cite{Liu2025LIDAR} by 1.35\% in F1 and 1.05\% in mIoU. On CrackDepth \cite{Liu2025LIDAR}, with 70\% and 90\% RGB missing, Compass improves F1 by 17.80\% and 11.65\% over CMX~\cite{Zhang2023CMX} and VRWKV~\cite{Duan2025Vision}, showing that FAPT reconstructs complete features while ETPF suppresses degraded-modality noise. Under 30\% RGB missing, Compass preserves branching connectivity, unlike most methods. With 90\% depth missing, Compass exceeds PrimKD~\cite{Hao2024PrimKD} by 6.24\% in F1. 

As shown in Table~\ref{tab:Complex_dual_modals}, Compass has the lowest computational cost, reducing Model Size, Params, and FLOPs by 25.64\%, 51.78\%, and 11.85\% over LIDAR~\cite{Liu2025LIDAR}, and Params and Model Size by 239 times and 126 times over PrimKD~\cite{Hao2024PrimKD}. This efficiency comes from Needle’s effective feature extraction with only four layers.

\noindent \textbf{Comparative Analysis on Tri-Modal Dataset.}  As shown in Table \ref{tab:tri_modals} and Figure \ref{fig:tri_results}, Compass achieves the best performance. With RGB complete and AoP and DoP missing at 10\% and 30\%, Compass improves F1 by 2.09\% over CMNeXT~\cite{Zhang2023Delivering}. Under severe polarization missing, CMNeXT~\cite{Zhang2023Delivering} and mmsFormer~\cite{Reza2024MMSFormer} suffer from crack fragmentation, while Compass maintains topological connectivity, showing that DSD enables FAPT to recover polarization cues. When RGB is 90\% missing with only AoP complete, Compass surpasses CMNeXT~\cite{Zhang2023Delivering} by 7.14\% and LIDAR~\cite{Liu2025LIDAR} by 22.07\% in F1, confirming the efficacy of ETPF in estimating modality reliability. When all three modalities degrade, Compass outperforms CMNeXT~\cite{Zhang2023Delivering} by 18.48\% in F1 and LIDAR~\cite{Liu2025LIDAR} by 67.93\% in F1 and 22.92\% in mIoU, demonstrating the synergy of DSD, Needle, and ETPF. 

As shown in Table \ref{tab:Complex_tri_modals}, Compass requires only 42.82G FLOPs, 3.80M Params, and 58MB Model Size under tri-modal input, reducing FLOPs, Params, and Model Size by 14.36\%, 52.62\%, and 61.33\% over LIDAR~\cite{Liu2025LIDAR}, and Params by 91 times compared to PWRF~\cite{Liu2025Part}.

\vspace{-0.3cm}
\subsection{Ablation Studies}

We conducted ablations on CrackPolar~\cite{Liu2025LIDAR} with RGB available and AoP and DoP each 70\% missing.

\noindent \textbf{Ablation Study on Different DSD Variants.} As shown in Figure~\ref{fig:abl_distill}, compared with the single-stream baseline, the full DSD improves F1 and mIoU by 3.09\% and 1.50\%. Removing reciprocal learning reduces F1 and mIoU by 2.63\% and 0.68\%, indicating the importance of degradation stream feedback for regularization. Removing $\mathcal{L}_{IMD}$, $\mathcal{L}_{MGC}$, and $\mathcal{L}_{FC}$ decreases F1 by 2.58\%, 1.64\%, and 1.93\%, highlighting the need for bidirectional distillation, topology preservation, and feature-level supervision.

\begin{figure}[!t]
  \centering
  \includegraphics[width=0.47\textwidth]{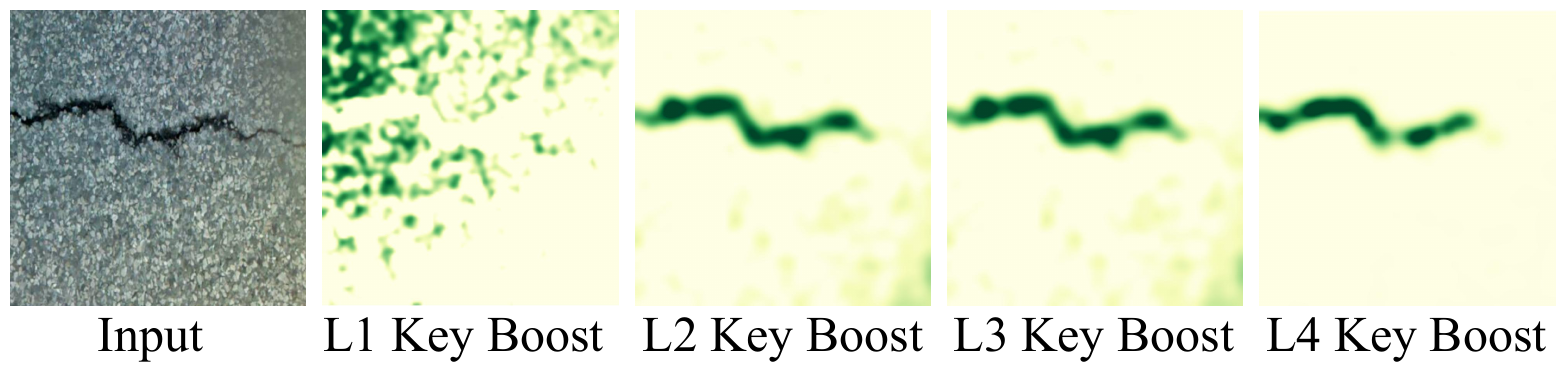}
    \caption{Visualization of Key boost by GCM. The greener the color, the greater the enhancement.}
  \label{fig:gcm_key_vis}
  \vspace{-0.5cm}
\end{figure}

\noindent \textbf{Ablation Study on Token Processing Strategies.} As shown in Table \ref{tab:abl_scan}, GCM outperforms EfficientDilated \cite{Pei2025EfficientVMamba}, ZigZag \cite{Chen2025Zig}, Q-Shift \cite{Duan2025Vision}, and SAVSS \cite{Liu2025SCSegamba}, improving F1 by 1.02\%, 1.04\%, 1.31\%, and 1.75\%, respectively, while increasing mIoU by 0.19\% over Q-Shift. Unlike methods that only modify token ordering, GCM injects crack orientation into WKV key modulation, assigning higher weights to directionally aligned tokens and improving propagation along anisotropic crack structures. With a token-processing delay of only 1.18E-04s, it is 4.40$\times$ and 3.83$\times$ faster than SAVSS and 4Dir \cite{Liu2024VMamba}, respectively, due to its lightweight directional prediction and scalar key modulation. As shown in Figure \ref{fig:gcm_key_vis}, GCM progressively shifts modulation from scattered crack and background regions to the crack skeleton, emphasizing direction-aligned tokens while suppressing background responses. This enhances their contribution to WKV aggregation and forms an anisotropic receptive field.



\begin{table}[!t]
    \centering
    \setlength{\tabcolsep}{5pt}
    \renewcommand{\arraystretch}{0.9}
    \caption{Comparison of different token process  strategies.}
    \resizebox{\linewidth}{!}{
    \begin{tabular}{cccc}
    \hline
    Method & F1 & mIoU & Delay Time$\downarrow$ \\ \hline
    Window (NeurIPS 2024) \cite{Huang2024LocalMamba} & 0.7350 & 0.7981 & 2.81E-04s \\
    4Dir (NeurIPS 2024) \cite{Liu2024VMamba} & 0.7360 & \cellcolor{orange!12}\underline{0.8022} & 4.53E-04s \\
    EfficientDilated (AAAI 2025) \cite{Pei2025EfficientVMamba}  & \cellcolor{orange!12}\underline{0.7394} & 0.8015 & \cellcolor{orange!12}\underline{2.65E-04s} \\
    SAVSS (CVPR 2025) \cite{Liu2025SCSegamba} & 0.7341 & 0.8007 & 5.20E-04s \\
    ZigZag (TMI 2025) \cite{Chen2025Zig} & 0.7392 & 0.8012 & 3.18E-04s \\
    Q-Shift (ICLR 2025) \cite{Duan2025Vision} & 0.7373 & 0.8015 & 2.88E-04s \\
    \textbf{GCM (Ours)} & \cellcolor{cyan!10}\textbf{0.7469} & \cellcolor{cyan!10}\textbf{0.8031} & \cellcolor{cyan!10}\textbf{1.18E-04s} \\
    \hline
    \end{tabular}
    }
    \label{tab:abl_scan}
    \vspace{-0.5cm}
\end{table}

\noindent \textbf{Ablation Study on Different Missing Handling Strategies.} As shown in Table \ref{tab:abl_miss}, Compass outperforms SKD~\cite{Sikdar2024SKD}, MCMoE~\cite{Xu2026MCMoE}, and MD2N~\cite{Dai2025Unbiased}, improving F1 by 0.91\%, 1.13\%, and 2.74\%, respectively, while requiring substantially fewer parameters and FLOPs. Unlike methods that passively compensate for missing modalities at inference, Compass models degradation during training through DSD-based reciprocal learning and dual-stream distillation. Meanwhile, FAPT preserves semantic integrity under severe modality absence, achieving superior accuracy and efficiency.


\begin{table}[!t]
    \centering
    \setlength{\tabcolsep}{4pt}
    \renewcommand{\arraystretch}{0.9}
    \caption{Comparison of different missing modality processing strategies.}
    \resizebox{\linewidth}{!}{
    \begin{tabular}{cccccc}
    \hline
        Method & F1 & mIoU & Params$\downarrow$ & FLOPs$\downarrow$ & Size$\downarrow$ \\ \hline
        SKD (ICRA 2024) \cite{Sikdar2024SKD} & \cellcolor{orange!12}\underline{0.7402} & 0.7996 & 14.46M & 611.18G & 185MB \\
        KB (CVPR 2025) \cite{Ke2025Knowledge} & 0.7310 & 0.7989 & 4.15M & 73.19G & 58MB \\
        MD2N (ICCV 2025) \cite{Dai2025Unbiased} & 0.7270 & 0.7922 & 11.85M & 159.4G & 148MB \\
        CCSD (arXiv 2025) \cite{Xie2025CCSD} & 0.7322 & 0.7990 & \cellcolor{orange!12}\underline{3.88M} & \cellcolor{orange!12}\underline{53.50G} & \cellcolor{cyan!10}\textbf{56MB} \\
        MCMoE (AAAI 2026) \cite{Xu2026MCMoE} & 0.7386 & \cellcolor{orange!12}\underline{0.8008} & 3.97M & 54.27G & \cellcolor{orange!12}\underline{57MB} \\
        \textbf{DSD (Ours)} & \cellcolor{cyan!10}\textbf{0.7469} & \cellcolor{cyan!10}\textbf{0.8031} & \cellcolor{cyan!10}\textbf{3.80M} & \cellcolor{cyan!10}\textbf{42.82G} & 58MB \\
        \hline
    \end{tabular}
    }
    \label{tab:abl_miss}
    \vspace{-0.4cm}
\end{table}
\begin{table}[!t]
    \centering
    \setlength{\tabcolsep}{3pt}
    \renewcommand{\arraystretch}{0.9}
    \caption{Comparison of different fusion strategies.}
    \resizebox{\linewidth}{!}{
    \begin{tabular}{cccccc}
    \hline
        Method & F1 & mIoU & Params$\downarrow$ & FLOPs$\downarrow$ & Size$\downarrow$ \\ \hline
        CEN (TPAMI 2023) \cite{Wang2023Channel} & 0.7377 & \cellcolor{orange!12}\underline{0.8023} & \cellcolor{orange!12}\underline{4.78M} & \cellcolor{orange!12}\underline{57.03G} & \cellcolor{orange!12}\underline{69MB} \\
        FreqFusion (TPAMI 2024) \cite{Chen2024Frequency} & 0.7400 & 0.8012 & 4.92M & 58.70G & 71MB \\
        DFormerv2 (CVPR 2025) \cite{Yin2025DFormerv2} & 0.7399 & 0.7974 & 5.12M & 61.61G & 73MB \\
        DEF (IF 2025) \cite{Huang2025Deep} & 0.7354 & 0.8004 & 4.87M & 60.35G & 70MB \\
        StitchFusion (MM 2025) \cite{Li2025StitchFusion} & \cellcolor{orange!12}\underline{0.7419} & 0.7987 & 5.27M & 77.50G & 76MB \\
        \textbf{ETPF (Ours)} & \cellcolor{cyan!10}\textbf{0.7469} & \cellcolor{cyan!10}\textbf{0.8031} & \cellcolor{cyan!10}\textbf{3.80M} & \cellcolor{cyan!10}\textbf{42.82G} & \cellcolor{cyan!10}\textbf{58MB} \\ \hline
    \end{tabular}
    }
    \label{tab:abl_fusion}
    \vspace{-0.7cm}
\end{table}

\noindent \textbf{Ablation Study on Different Fusion Strategies.} As shown in Table \ref{tab:abl_fusion}, compared to DEF~\cite{Huang2025Deep}, Compass improves F1 and mIoU by 1.56\% and 0.34\%, while reducing FLOPs by 29.05\%. ETPF has two key advantages over traditional evidence fusion, anisotropic evidence propagation preserves topology by diffusing confidence along crack directions, and multi-scale evidence consistency regularizes confidence from coarse to fine scales. Compared to StitchFusion~\cite{Li2025StitchFusion}, Compass improves F1 and mIoU by 0.68\% and 0.55\%, reducing FLOPs by 44.75\% and Params by 27.89\%. Compared to DFormerv2~\cite{Yin2025DFormerv2}, Compass improves F1 and mIoU by 0.95\% and 0.71\%, with FLOPs reduced by 30.50\%. Compared to CEN~\cite{Wang2023Channel} and FreqFusion~\cite{Chen2024Frequency}, Compass improves F1 by 1.25\% and 0.93\%, respectively. These results show that existing fusion methods lack explicit reliability assessment for missing modalities, while ETPF models modality confidence and uncertainty using Dempster-Shafer theory and an uncertainty-gated decoder, enabling more reliable fusion under missing conditions.

\FloatBarrier

\vspace{-0.3cm}
\section{Conclusion} 

In this paper, we present Compass, a lightweight Needle RWKV network for multimodal crack segmentation under arbitrary missing modalities. DSD constructs harsher missing conditions via degradation simulation and reciprocal learning, enhancing robustness from passive compensation to active adaptation. FAPT reconstructs complete features for missing modalities using modality-agnostic semantic prototypes. Needle models crack anisotropic structures through GCM, CohGate and AGP. ETPF models modality confidence and uncertainty with evidence theory, combining anisotropic evidence propagation and uncertainty-gated decoding for reliable fusion and topology preservation under missing conditions. Experiments on three datasets show that Compass achieves optimal performance, including gains of 67.93\% in F1 and 22.92\% in mIoU over LIDAR under a tri-modal setting with 90\%, 30\%, and 70\% missing rates. Compass demonstrates potential for deployment in resource-constrained industrial settings. Future work will integrate ETPF’s uncertainty quantification into active learning to reduce reliance on large labeled datasets.

\section{Acknowledgement}

This work was supported by the Tianjin Natural Science Foundation under Grant 23JCJQJC00070; the National Natural Science Foundation of China (NSFC) under Grants 62272342, T2422015, and 62306212; the Beijing-Tianjin-Hebei Natural Science Foundation Cooperation Project under Grant 25JJJJC0009; the China Postdoctoral Science Foundation under Grant 2026M791667; and the Postdoctoral Fellowship Program of the China Postdoctoral Science Foundation (CPSF) under Grant GZC20260889.


\section{Appendix}

\appendix

\section{Theoretical Analysis}

\subsection{Theorem 1: Robustness Guarantee of DSD}
\label{sec:theorem1}

\textbf{Proposition.} The dual-stream training framework of DSD provides theoretical guarantees for robustness to modality absence from both the distributional and parametric perspectives. At the distributional level, the degradation simulation during training extends the support of the training distribution from the current modality-available distribution $\mathcal{P}_{\text{avail}}$ to a mixture distribution covering more severe missing patterns, which is equivalent to performing empirical risk minimization over a broader space of missing patterns. At the parametric level, the backward distillation term in $\mathcal{L}_{IMD}$ imposes an output-space consistency regularization on the model, constraining it to produce consistent output distributions under both complete and degraded inputs, thereby preventing the original stream from overfitting to specific modality combinations.

\textbf{Derivation.} Let the modality availability mask for an $N$-modal input be $\mathbf{m} = (m_1, \ldots, m_N) \in \{0,1\}^N$, where $m_i = 1$ indicates that the $i$-th modality is available and $m_i = 0$ indicates that it is missing. Define the missing severity of a mask as $s(\mathbf{m}) = \sum_{i=1}^N (1 - m_i)$, i.e., the number of missing modalities. The modality availability during training follows the distribution $\mathcal{P}_{\text{avail}}$. The degradation simulation stream of DSD uniformly selects one available modality from $\mathbf{m}$ at random and sets it to zero, producing a degraded mask $\mathbf{m}'$ satisfying $s(\mathbf{m}') = s(\mathbf{m}) + 1$, whose distribution is denoted $\mathcal{P}_{\text{deg}}$.

The model shares decoder parameters $\theta$, and the segmentation loss under mask $\mathbf{m}$ is $\ell(\theta; \mathbf{m})$. To focus on the core mechanism, the following analysis considers the segmentation losses of the two streams and $\mathcal{L}_{\text{IMD}}$, omitting the auxiliary terms $\mathcal{L}_{\text{MGC}}$ and $\mathcal{L}_{\text{FC}}$ which do not affect the distributional argument. The training objective of DSD can be formalized as:
\begin{equation}
\mathcal{L}_{\text{DSD}}(\theta) =  \; \mathbb{E}_{\mathbf{m} \sim \mathcal{P}_{\text{avail}}}[\ell(\theta; \mathbf{m})] \\
 + \mathbb{E}_{\mathbf{m}' \sim \mathcal{P}_{\text{deg}}}[\ell(\theta; \mathbf{m}')] + \mathcal{L}_{\text{IMD}},
\end{equation}
where the first term is the original stream loss, the second term is the degradation simulation stream loss, and $\mathcal{L}_{\text{IMD}}$ is the interactive mutual distillation loss. Since $\mathcal{P}_{\text{deg}}$ always has one more missing modality than $\mathcal{P}_{\text{avail}}$, the joint optimization of the first two terms is equivalent to performing empirical risk minimization over the mixture distribution $\mathcal{P}_{\text{mix}} = \frac{1}{2}\mathcal{P}_{\text{avail}} + \frac{1}{2}\mathcal{P}_{\text{deg}}$. The support of $\mathcal{P}_{\text{mix}}$ strictly contains that of $\mathcal{P}_{\text{avail}}$, covering the mask space at missing severities $s$ and $s+1$.

We further assume that the expected loss $\mathbb{E}_{\mathcal{P}}[\ell(\theta;\mathbf{m})]$ depends on the distribution $\mathcal{P}$ only through its expected missing severity $\bar{s} = \mathbb{E}_{\mathcal{P}}[s(\mathbf{m})]$, and that this dependence is convex in $\bar{s}$. Under this assumption, for a test-time missing distribution $\mathcal{P}_{\text{test}}$ whose expected missing severity satisfies $\bar{s}_{\text{test}} \in [\bar{s}_{\text{avail}},\, \bar{s}_{\text{deg}}]$, where $\bar{s}_{\text{avail}}$ and $\bar{s}_{\text{deg}}$ denote $\mathbb{E}[s(\mathbf{m})]$ under $\mathcal{P}_{\text{avail}}$ and $\mathcal{P}_{\text{deg}}$ respectively, the test risk is upper bounded by the training risk:
\begin{equation}
\begin{split}
& \mathbb{E}_{\mathcal{P}_{\text{test}}}[\ell(\theta; \mathbf{m})] \\
& \leq \max\!\left\{\mathbb{E}_{\mathcal{P}_{\text{avail}}}[\ell(\theta; \mathbf{m})],\; \mathbb{E}_{\mathcal{P}_{\text{deg}}}[\ell(\theta; \mathbf{m}')]\right\} \\
& \leq \mathcal{L}_{\text{DSD}}(\theta),
\end{split}
\end{equation}
where the first inequality follows from the convexity of the expected loss with respect to $\bar{s}$ and the fact that $\bar{s}_{\text{test}}$ lies within the interval $[\bar{s}_{\text{avail}},\, \bar{s}_{\text{deg}}]$. Therefore, minimizing $\mathcal{L}_{\text{DSD}}$ provides a risk upper bound guarantee for test distributions within this range of missing severity.

At the parametric level, the backward distillation term in IMDLoss can be expressed as:
\begin{equation}
\mathcal{L}_{\text{bwd}} = \alpha \cdot D_{\text{KL}}\!\left(\text{sg}(P_d^\tau) \;\|\; P_o^\tau\right),
\end{equation}
where $P_o^\tau = \text{softmax}(z_o / \tau)$ is the softmax output of the original stream at temperature $\tau$, with $z_o$ being the corresponding logit vector; $P_d^\tau = \text{softmax}(z_d / \tau)$ is the softmax output of the degradation simulation stream, with $z_d$ being the corresponding logit vector; $\text{sg}$ denotes the stop-gradient operator that treats $P_d^\tau$ as a fixed target; $\alpha = 0.3$ is the backward distillation weight; and $D_{\text{KL}}$ is the KL divergence. Since the two streams share decoder parameters $\theta$ but receive different inputs, $\mathcal{L}_{\text{bwd}}$ constitutes an output-space consistency regularization term: it penalizes the divergence between the output distributions produced by the model under the complete input $\mathcal{X}_o$ and the degraded input $\mathcal{X}_d$. Taking the gradient with respect to $\theta$ yields:
\begin{equation}
\nabla_\theta \mathcal{L}_{\text{bwd}} = -\alpha \!\sum_{u \in \Omega}\sum_{k=1}^{K} \text{sg}(P_{d,k}^\tau(u)) \cdot \nabla_\theta \log P_{o,k}^\tau(u),
\end{equation}
where $\Omega$ is the set of pixels, $u$ indexes pixel locations, $k$ indexes the $K$ semantic categories, and $P_{o,k}^\tau(u)$ and $P_{d,k}^\tau(u)$ denote the $k$-th class probability at pixel $u$ for the original and degraded streams respectively. This gradient pulls the original stream output toward the degraded stream output across all pixels and categories. This is equivalent to constraining the Lipschitz continuity of the model's output manifold with respect to input degradation: let $f_\theta$ denote the mapping from input to output distribution parameterized by $\theta$. Since minimizing $\mathcal{L}_{\text{DSD}}$ implicitly minimizes $\mathcal{L}_{\text{bwd}}$, the training process imposes an upper bound on the KL divergence between the two output distributions:
\begin{equation}
D_{\text{KL}}\!\left(f_\theta(\mathcal{X}_d) \;\|\; f_\theta(\mathcal{X}_o)\right) = \frac{\mathcal{L}_{\text{bwd}}}{\alpha},
\end{equation}
and minimizing $\mathcal{L}_{\text{bwd}}$ drives this divergence toward zero. When $\alpha = 0.3 < 1$, the forward distillation dominates in semantic transfer, while the backward term with a smaller weight constrains the output of the original stream to remain close to that of the degraded stream, enabling the model to maintain output consistency under degraded conditions.

\textbf{Conclusion.} DSD synergistically guarantees missing-modality robustness from both the distributional and parametric perspectives. The degradation simulation incorporates $\mathcal{P}_{\text{deg}}$ into the training distribution, extending the coverage of the model's empirical risk from the current missing severity $s$ to $s+1$, thereby providing a risk upper bound for test missing patterns lying between the two. The backward term of the reciprocal distillation enforces output consistency constraints, ensuring that the model maintains Lipschitz stability against input degradation and preventing the original stream from overfitting to complete modality combinations.

\subsection{Theorem 2: Directionally Selective ERF Anisotropy of GCM}
\label{sec:theorem2}

\textbf{Proposition.} Through key modulation, GCM introduces a direction-dependent multiplicative gain factor $\Gamma_\tau = e^{\delta a_\tau k_\tau}$ for each token, transforming the effective receptive field of WKV from isotropic to anisotropic with selective enhancement along crack directions. For crack-aligned tokens, the gain factor satisfies $\Gamma = e^{\delta\bar{k}_{\text{crack}}} \gg 1$, while for background tokens the gain factor satisfies $\Gamma \approx 1$. Since tokens along the crack direction are crack-aligned with high probability and tokens in the perpendicular direction are background tokens with high probability, the equal-sensitivity contour of the two-dimensional effective receptive field exhibits an elliptical shape elongated along the crack direction.

\textbf{Derivation.} Define the effective receptive field sensitivity of the output $o_t$ at position $t$ with respect to the input $x_\tau$ at position $\tau$ as:
\begin{equation}
\text{ERF}(t, \tau) = \left\| \frac{\partial o_t}{\partial x_\tau} \right\|,
\end{equation}
where $o_t$ is the WKV output at position $t$, $x_\tau$ is the input feature at position $\tau$, and $\|\cdot\|$ denotes the Euclidean norm. In the standard WKV recurrence, the unnormalized aggregation weight of position $\tau$ contributing to position $t$ is proportional to:
\begin{equation}
\omega_{\text{std}}(t, \tau) \propto \exp\!\left(-(t\!-\!1\!-\!\tau)w + k_\tau\right),
\end{equation}
where $w > 0$ is a global decay parameter controlling the forgetting rate of historical information, and $k_\tau$ is the key value at position $\tau$ characterizing the semantic salience of that token. With the introduction of GCM, the key is modulated to $k'_\tau = k_\tau \cdot (1 + \delta \cdot a_\tau)$, where $\delta = \sigma(\gamma) \in (0,1)$ is the learnable modulation strength normalized by the sigmoid function, $\gamma$ is the raw learnable parameter, and $a_\tau = \langle \mathbf{d}_\tau, \mathbf{c}_\tau \rangle \in [-1, 1]$ is the inner product between the crack direction unit vector $\mathbf{d}_\tau$ and the cross-modal context vector $\mathbf{c}_\tau$. Under GCM, the aggregation weight becomes:
\begin{equation}
\omega_{\text{gcm}}(t, \tau) \propto \exp\big( -(t\!-\!1\!-\!\tau)w  + k_\tau + \delta a_\tau k_\tau\big),
\end{equation}

The weight ratio between GCM and standard WKV yields a per-token multiplicative gain factor:
\begin{equation}
\frac{\omega_{\text{gcm}}(t, \tau)}{\omega_{\text{std}}(t, \tau)} = \exp(\delta a_\tau k_\tau) \triangleq \Gamma_\tau,
\end{equation}
It should be noted that $\Gamma_\tau$ depends solely on the directional alignment and key value of token $\tau$ itself and does not alter the step-wise decay ratio $e^{-w}$ of WKV. For crack-direction-aligned tokens, $a_\tau \approx 1$ and semantic activity ensures $k_\tau > 0$, yielding $\Gamma_\tau = e^{\delta k_\tau} \gg 1$; for background tokens, $a_\tau \approx 0$, yielding $\Gamma_\tau \approx 1$.

We now analyze the anisotropy of the two-dimensional effective receptive field. Consider a horizontal crack with strike angle $\varphi = 0$, where the query position is a token on the crack. In the horizontal scan, a token at distance $d$ along the crack direction is a crack-aligned token, whose aggregation weight is:
\begin{equation}
\omega_h^{\text{crack}}(d) = \exp\!\left(-dw + (1\!+\!\delta)\bar{k}_{\text{crack}}\right),
\end{equation}
where $\bar{k}_{\text{crack}} > 0$ is the mean key value of crack tokens. In the vertical scan, a token at distance $d$ in the perpendicular direction is a background token, whose aggregation weight is:
\begin{equation}
\omega_v^{\text{bg}}(d) = \exp\!\left(-dw + \bar{k}_{\text{bg}}\right),
\end{equation}
where $\bar{k}_{\text{bg}}$ is the mean key value of background tokens, satisfying $\bar{k}_{\text{bg}} < \bar{k}_{\text{crack}}$. The equal-sensitivity contour is defined as the spatial locus where $\omega = \epsilon$. Solving for the effective range in each direction gives:
\begin{equation}
L_\parallel = \frac{(1\!+\!\delta)\bar{k}_{\text{crack}} - \ln\epsilon}{w},\; L_\perp = \frac{\bar{k}_{\text{bg}} - \ln\epsilon}{w},
\end{equation}
where $L_\parallel$ is the effective range along the crack direction, $L_\perp$ is the effective range in the perpendicular direction, and $\epsilon > 0$ is the sensitivity threshold. The difference between the two is:
\begin{equation}
L_\parallel - L_\perp = \frac{(1\!+\!\delta)\bar{k}_{\text{crack}} - \bar{k}_{\text{bg}}}{w} = \frac{\Delta k + \delta\bar{k}_{\text{crack}}}{w},
\end{equation}
where $\Delta k = \bar{k}_{\text{crack}} - \bar{k}_{\text{bg}} > 0$ is the intrinsic key value difference between crack and background tokens, and $\delta\bar{k}_{\text{crack}}$ is the additional gain introduced by GCM. The ratio of effective ranges is:
\begin{equation}
R = \frac{L_\parallel}{L_\perp} = \frac{(1\!+\!\delta)\bar{k}_{\text{crack}} - \ln\epsilon}{\bar{k}_{\text{bg}} - \ln\epsilon},
\end{equation}

When the sensitivity threshold $\epsilon$ is small, $-\ln\epsilon \gg \bar{k}$, and both the numerator and the denominator are dominated by $-\ln\epsilon$. In this regime:
\begin{equation}
R \approx 1 + \frac{(1\!+\!\delta)\bar{k}_{\text{crack}} - \bar{k}_{\text{bg}}}{-\ln\epsilon + \bar{k}_{\text{bg}}} > 1,
\end{equation}
The contribution of GCM is reflected in the additional $\delta\bar{k}_{\text{crack}}$ term in the numerator. When $\delta = 0$, $R$ reduces to the baseline anisotropy that depends only on the intrinsic key value difference; when $\delta > 0$, GCM systematically enhances the effective range along the crack direction, further increasing the anisotropy ratio $R$. For a crack at an arbitrary strike angle $\varphi$, the crack direction unit vector is $\mathbf{d} = (\cos\varphi,\, \sin\varphi)$. In the horizontal scan, the displacement vector between adjacent tokens is aligned with the $x$-axis, so the cross-modal context vector satisfies $\mathbf{c}_\tau \propto (1, 0)$ for tokens along the scan direction. The directional alignment score is therefore $a_h = \langle \mathbf{d}, \mathbf{c}_\tau \rangle \propto \cos\varphi$, and since the gain factor depends on $a_\tau k_\tau$ and the effective weight is quadratic in the alignment through the exponential, the net directional selectivity scales as $\cos^2\varphi$. By the same reasoning, in the vertical scan $a_v \propto \sin\varphi$ and the selectivity scales as $\sin^2\varphi$. The equal-sensitivity contour of the bidirectional fusion $o = (o_h + o_v)/2$ in two-dimensional space therefore exhibits an elliptical shape elongated along the crack direction.

\textbf{Conclusion.} GCM selectively amplifies the contribution weight of crack-aligned tokens in WKV aggregation through a per-token directionally selective multiplicative gain $\Gamma_\tau = e^{\delta a_\tau k_\tau}$, while leaving the weights of background tokens unchanged. Since tokens along the crack direction are crack-aligned with high probability, GCM causes the effective receptive field range in that direction to be systematically larger than in the perpendicular direction, and the two-dimensional equal-sensitivity contour exhibits an elliptical shape elongated along the crack direction. When $\delta = 0$, the result reduces to the baseline anisotropy of standard WKV; as $\delta$ increases, the anisotropy ratio $R$ monotonically increases, achieving controllable directionally selective enhancement without altering the linear computational complexity of WKV.

\subsection{Theorem 3: Monotonic Uncertainty Decrease under DS Combination}
\label{sec:theorem3}

\textbf{Proposition.} When two conditions are simultaneously satisfied, namely that the conflict coefficient satisfies $C < 1$ and that each modality possesses nonzero evidence (i.e., $u_m < 1$), the Dempster-Shafer combination rule guarantees that the fused uncertainty is strictly less than the uncertainty of any single modality, i.e., $u_c < \min(u_1, u_2)$. This property can be extended to the sequential pairwise combination of $M$ modalities via mathematical induction, guaranteeing that the uncertainty is strictly monotonically decreasing with respect to the number of available modalities.

\textbf{Derivation.} Consider the evidence framework for two modalities. Following the definitions in the Methodology section, the Dirichlet parameters for each modality $m$ are $\boldsymbol{\alpha}_m = \mathbf{e}_m + 1$, where $\mathbf{e}_m$ is the nonnegative evidence vector. The total strength is $S_m = \sum_k \alpha_{m,k}$, where $k$ indexes the categories. The belief mass is $b_{m,k} = e_{m,k} / S_m$, representing the degree of support that modality $m$ assigns to category $k$. The uncertainty is $u_m = K / S_m$, where $K$ is the number of categories, representing the overall degree of evidential insufficiency for modality $m$. These quantities satisfy the normalization constraint:
\begin{equation}
\sum_{k=1}^{K} b_{m,k} + u_m = 1,
\end{equation}

The condition $u_m < 1$ is equivalent to modality $m$ providing nonzero evidence for at least one category, i.e., $\exists k, e_{m,k} > 0$. According to the DS combination rule, the fused uncertainty is:
\begin{equation}
u_c = \frac{u_1 \cdot u_2}{1 - C},
\end{equation}
where $C = \sum_{k \neq l} b_{1,k} \cdot b_{2,l}$ is the conflict coefficient between the two modalities, measuring the degree of contradiction in their categorical judgments, and $1 - C > 0$ is the normalization constant.

We now prove $u_c < u_1$, which is equivalent to proving $u_2 < 1 - C$. Expanding $1 - C$, since the sum of all agreement masses, residual masses, and conflict masses equals unity, we have:
\begin{equation}
 \sum_k b_{1,k} b_{2,k} + \sum_k b_{1,k} u_2 
 + \sum_k u_1 b_{2,k} + u_1 u_2 + C = 1,
\end{equation}

Rearranging yields:
\begin{equation}
1 - C = \sum_k b_{1,k} b_{2,k} + u_2 \sum_k b_{1,k} + u_1 \sum_k b_{2,k} + u_1 u_2,
\end{equation}

Substituting the normalization constraint $\sum_k b_{m,k} = 1 - u_m$ gives:
\begin{equation}
\begin{split}
1 - C & = \sum_k b_{1,k} b_{2,k} + u_2(1 - u_1) \\
& \quad + u_1(1 - u_2) + u_1 u_2 \\
& = \sum_k b_{1,k} b_{2,k} + u_1 + u_2 - u_1 u_2,
\end{split}
\end{equation}

Substituting the target inequality $u_2 < 1 - C$ into the expression above and simplifying, we need to verify:
\begin{equation}
0 < \sum_k b_{1,k} b_{2,k} + u_1(1 - u_2),
\end{equation}
Since $u_1 > 0$ (because $K \geq 1$ and $S_1$ is finite) and $u_2 < 1$ (by the nonzero evidence condition), we have $u_1(1 - u_2) > 0$. At the same time, $\sum_k b_{1,k} b_{2,k} \geq 0$. Therefore, the inequality holds unconditionally, and $u_c < u_1$ is established. By the symmetry of the two modalities, $u_c < u_2$ follows analogously. Combining the two results gives:
\begin{equation}
u_c < \min(u_1, u_2),
\end{equation}

The above conclusion is extended to $M$ modalities by mathematical induction. The base case $M = 2$ has been proved. Suppose that after sequential pairwise DS combination of the first $m$ modalities, the fused uncertainty satisfies $u_{1:m} < \min(u_1, \ldots, u_m)$, and that the pairwise conflict coefficient at each combination step satisfies $C < 1$. Since $u_{1:m} < u_i < 1$ holds for all $i$, the nonzero evidence condition $u_{1:m} < 1$ is still satisfied. Upon introducing the $(m+1)$-th modality, treating $u_{1:m}$ as the uncertainty of an integrated evidence framework and combining it with $u_{m+1}$ via DS combination, provided that the conflict coefficient between the integrated framework and the $(m+1)$-th modality also satisfies $C < 1$, the base case yields:
\begin{equation}
u_{1:m+1}  < \min(u_{1:m},\, u_{m+1})
 \leq \min(u_1, \ldots, u_m, u_{m+1}),
\end{equation}
This completes the induction. Therefore, provided that the pairwise conflict coefficient remains strictly less than one at every combination step, the uncertainty after sequential pairwise DS combination of $M$ modalities is strictly less than the minimum of all individual modality uncertainties, and it is strictly monotonically decreasing as the number of available modalities $M$ increases.

\textbf{Conclusion.} The Dempster-Shafer evidence combination in ETPF guarantees the strict monotonic decrease of uncertainty with respect to the number of available modalities: each additional modality that provides nonzero evidence and whose conflict with the existing fused evidence satisfies $C < 1$ necessarily reduces the fused decision uncertainty. This establishes a theoretical lower bound on performance degradation under modality-missing conditions. Missing modalities merely reduce the number of evidence sources participating in the fusion and thus slow down the rate at which uncertainty decreases, but they do not cause uncertainty to rebound or the fusion quality to deteriorate. Specifically, if the fused uncertainty with the full set of $M$ modalities is $u_{1:M}$ and the uncertainty after losing one modality rises to $u_{1:M-1}$, the relation $u_{1:M-1} < \min(u_1, \ldots, u_{M-1})$ still holds, guaranteeing a reliability lower bound for fused decisions under missing-modality scenarios.

\newcolumntype{G}{c} 
\begin{table*}[!t]
    \centering
    \scriptsize
    \setlength{\tabcolsep}{2pt}
    \renewcommand{\arraystretch}{0.9}
    \caption{Performance comparison under dual-modal and multi missing settings. {\setlength{\fboxsep}{1pt}\colorbox{cyan!10}{\textbf{Blue}}} and {\setlength{\fboxsep}{1pt}\colorbox{orange!12}{\underline{Orange}}} denote best and second-best.}
    \resizebox{\textwidth}{!}{
    \begin{tabular}{lcc|*{8}{cc}cc}
    \toprule
    \multirow{2}{*}{\rotatebox{90}{}} & \multicolumn{2}{c|}{Missing Ratio}
    & \multicolumn{2}{c}{CMX \cite{Zhang2023CMX}} & \multicolumn{2}{c}{PrimKD \cite{Hao2024PrimKD}} & \multicolumn{2}{c}{Sigma \cite{Wan2025Sigma}} & \multicolumn{2}{c}{CMNeXT \cite{Zhang2023Delivering}} & \multicolumn{2}{c}{GminiFu \cite{Jia2024GeminiFusion}} & \multicolumn{2}{c}{PWRF  \cite{Liu2025Part}} & \multicolumn{2}{c}{LIDAR \cite{Liu2025LIDAR}} & \multicolumn{2}{c}{VRWKV \cite{Duan2025Vision}} & \multicolumn{2}{c}{\textbf{Compass}} \\
    \midrule

    \multirow{12}{*}{\rotatebox{90}{CrackPolar \cite{Liu2025LIDAR}}}
        & RGB & DoP & F1 & mIoU & F1 & mIoU & F1 & mIoU & F1 & mIoU & F1 & mIoU & F1 & mIoU & F1 & mIoU & F1 & mIoU & F1 & mIoU \\ \midrule
        & 10\% & Full & 0.7197 & 0.7923 & 0.7233 & \cellcolor{orange!12}\underline{0.7963} & \cellcolor{orange!12}\underline{0.7242} & 0.7953 & 0.7043 & 0.7854 & 0.6714 & 0.7637 & 0.6719 & 0.7742 & 0.6935 & 0.7751 & 0.6928 & 0.7788 & \cellcolor{cyan!10}\textbf{0.7463} & \cellcolor{cyan!10}\textbf{0.8029} \\
        & 30\% & Full & 0.7036 & \cellcolor{orange!12}\underline{0.7870} & \cellcolor{orange!12}\underline{0.7164} & 0.7841 & 0.6873 & 0.7746 & 0.6728 & 0.7677 & 0.6449 & 0.7487 & 0.6404 & 0.7546 & 0.6430 & 0.7410 & 0.6500 & 0.7569 & \cellcolor{cyan!10}\textbf{0.7277} & \cellcolor{cyan!10}\textbf{0.7946} \\
        & 50\% & Full & 0.6866 & 0.7779 & \cellcolor{orange!12}\underline{0.6957} & \cellcolor{orange!12}\underline{0.7831} & 0.6793 & 0.7771 & 0.6635 & 0.7649 & 0.6332 & 0.7432 & 0.6128 & 0.7434 & 0.6345 & 0.7372 & 0.6343 & 0.7504 & \cellcolor{cyan!10}\textbf{0.7137} & \cellcolor{cyan!10}\textbf{0.7872} \\
        & 70\% & Full & 0.6757 & 0.7655 & \cellcolor{orange!12}\underline{0.6882} & 0.7660 & 0.6724 & \cellcolor{orange!12}\underline{0.7691} & 0.6502 & 0.7524 & 0.6233 & 0.7407 & 0.6105 & 0.7355 & 0.6366 & 0.7388 & 0.6235 & 0.7440 & \cellcolor{cyan!10}\textbf{0.7155} & \cellcolor{cyan!10}\textbf{0.7835} \\
        & 90\% & Full & 0.6738 & 0.7596 & \cellcolor{orange!12}\underline{0.6791} & 0.7600 & 0.6703 & \cellcolor{orange!12}\underline{0.7628} & 0.6414 & 0.7533 & 0.6018 & 0.7195 & 0.5960 & 0.7286 & 0.6266 & 0.7245 & 0.5959 & 0.7233 & \cellcolor{cyan!10}\textbf{0.7168} & \cellcolor{cyan!10}\textbf{0.7871} \\
        & Full & 10\% & 0.7197 & 0.7923 & 0.7169 & 0.7958 & \cellcolor{orange!12}\underline{0.7271} & \cellcolor{orange!12}\underline{0.7970} & 0.7026 & 0.7886 & 0.6656 & 0.7587 & 0.6891 & 0.7744 & 0.7049 & 0.7818 & 0.7002 & 0.7774 & \cellcolor{cyan!10}\textbf{0.7397} & \cellcolor{cyan!10}\textbf{0.8023} \\
        & Full & 30\% & \cellcolor{orange!12}\underline{0.7160} & \cellcolor{orange!12}\underline{0.7909} & 0.7106 & 0.7844 & 0.7101 & 0.7826 & 0.6987 & 0.7890 & 0.6424 & 0.7523 & 0.6674 & 0.7721 & 0.6887 & 0.7725 & 0.6661 & 0.7568 & \cellcolor{cyan!10}\textbf{0.7385} & \cellcolor{cyan!10}\textbf{0.7993} \\
        & Full & 50\% & 0.6947 & 0.7884 & \cellcolor{orange!12}\underline{0.6981} & \cellcolor{orange!12}\underline{0.7897} & 0.6962 & 0.7814 & 0.6912 & 0.7863 & 0.6424 & 0.7523 & 0.6630 & 0.7713 & 0.6810 & 0.7672 & 0.6350 & 0.7452 & \cellcolor{cyan!10}\textbf{0.7357} & \cellcolor{cyan!10}\textbf{0.8003} \\
        & Full & 70\% & \cellcolor{orange!12}\underline{0.6944} & 0.7844 & 0.6920 & \cellcolor{orange!12}\underline{0.7868} & 0.6938 & 0.7806 & 0.6782 & 0.7802 & 0.6389 & 0.7538 & 0.6585 & 0.7682 & 0.6757 & 0.7593 & 0.6327 & 0.7472 & \cellcolor{cyan!10}\textbf{0.7274} & \cellcolor{cyan!10}\textbf{0.8002} \\
        & Full & 90\% & 0.6759 & 0.7784 & 0.6811 & 0.7794 & \cellcolor{orange!12}\underline{0.6846} & \cellcolor{orange!12}\underline{0.7893} & 0.6740 & 0.7751 & 0.6179 & 0.7399 & 0.6224 & 0.7530 & 0.6452 & 0.7441 & 0.6120 & 0.7287 & \cellcolor{cyan!10}\textbf{0.7241} & \cellcolor{cyan!10}\textbf{0.7963} \\
        & Full & Full & 0.7237 & \cellcolor{orange!12}\underline{0.7986} & 0.7255 & 0.7966 & 0.7289 & 0.7979 & 0.7246 & 0.7953 & 0.6946 & 0.7776 & 0.6988 & 0.7788 & \cellcolor{orange!12}\underline{0.7290} & 0.7983 & 0.7278 & 0.7970 & \cellcolor{cyan!10}\textbf{0.7452} & \cellcolor{cyan!10}\textbf{0.8075} \\
    \midrule

    \multirow{12}{*}{\rotatebox{90}{CrackPolar \cite{Liu2025LIDAR}}}
        & RGB & AoP & F1 & mIoU & F1 & mIoU & F1 & mIoU & F1 & mIoU & F1 & mIoU & F1 & mIoU & F1 & mIoU & F1 & mIoU & F1 & mIoU \\ \midrule
        & 10\% & Full & \cellcolor{orange!12}\underline{0.7175} & 0.7912 & 0.7140 & \cellcolor{orange!12}\underline{0.7958} & 0.7049 & 0.7928 & 0.7030 & 0.7854 & 0.6718 & 0.7678 & 0.6859 & 0.7757 & 0.6704 & 0.7582 & 0.6789 & 0.7744 & \cellcolor{cyan!10}\textbf{0.7362} & \cellcolor{cyan!10}\textbf{0.7982} \\
        & 30\% & Full & 0.6958 & 0.7854 & 0.6917 & 0.7814 & 0.6935 & \cellcolor{orange!12}\underline{0.7858} & \cellcolor{orange!12}\underline{0.6971} & 0.7826 & 0.6499 & 0.7476 & 0.6593 & 0.7554 & 0.5756 & 0.6877 & 0.6534 & 0.7551 & \cellcolor{cyan!10}\textbf{0.7327} & \cellcolor{cyan!10}\textbf{0.7968} \\
        & 50\% & Full & 0.6831 & \cellcolor{orange!12}\underline{0.7810} & \cellcolor{orange!12}\underline{0.6869} & 0.7711 & 0.6736 & 0.7768 & 0.6751 & 0.7730 & 0.6317 & 0.7445 & 0.6204 & 0.7482 & 0.5331 & 0.6908 & 0.6434 & 0.7480 & \cellcolor{cyan!10}\textbf{0.7203} & \cellcolor{cyan!10}\textbf{0.7991} \\
        & 70\% & Full & 0.6631 & 0.7550 & 0.6612 & \cellcolor{orange!12}\underline{0.7604} & 0.6605 & 0.7591 & \cellcolor{orange!12}\underline{0.6631} & 0.7559 & 0.6375 & 0.7438 & 0.6079 & 0.7281 & 0.5353 & 0.6748 & 0.6047 & 0.7339 & \cellcolor{cyan!10}\textbf{0.7119} & \cellcolor{cyan!10}\textbf{0.7852} \\
        & 90\% & Full & 0.6499 & 0.7548 & 0.6415 & 0.7519 & 0.6540 & \cellcolor{orange!12}\underline{0.7564} & \cellcolor{orange!12}\underline{0.6593} & 0.7563 & 0.6117 & 0.7326 & 0.5762 & 0.7112 & 0.4989 & 0.6660 & 0.5998 & 0.7226 & \cellcolor{cyan!10}\textbf{0.7013} & \cellcolor{cyan!10}\textbf{0.7815} \\
        & Full & 10\% & 0.7118 & 0.7890 & 0.7162 & \cellcolor{orange!12}\underline{0.7917} & \cellcolor{orange!12}\underline{0.7227} & 0.7916 & 0.7025 & 0.7883 & 0.6704 & 0.7685 & 0.6822 & 0.7739 & 0.7154 & 0.7880 & 0.7106 & 0.7841 & \cellcolor{cyan!10}\textbf{0.7459} & \cellcolor{cyan!10}\textbf{0.8056} \\
        & Full & 30\% & 0.7022 & 0.7914 & 0.7086 & 0.7892 & \cellcolor{orange!12}\underline{0.7094} & \cellcolor{orange!12}\underline{0.7931} & 0.6998 & 0.7843 & 0.6574 & 0.7628 & 0.6616 & 0.7696 & 0.6941 & 0.7770 & 0.6881 & 0.7738 & \cellcolor{cyan!10}\textbf{0.7449} & \cellcolor{cyan!10}\textbf{0.8057} \\
        & Full & 50\% & 0.6938 & 0.7890 & \cellcolor{orange!12}\underline{0.7038} & 0.7845 & 0.6926 & \cellcolor{orange!12}\underline{0.7912} & 0.6901 & 0.7807 & 0.6327 & 0.7543 & 0.6644 & 0.7707 & 0.6897 & 0.7752 & 0.6762 & 0.7685 & \cellcolor{cyan!10}\textbf{0.7420} & \cellcolor{cyan!10}\textbf{0.8047} \\
        & Full & 70\% & 0.6927 & \cellcolor{orange!12}\underline{0.7861} & \cellcolor{orange!12}\underline{0.6981} & 0.7772 & 0.6922 & 0.7857 & 0.6805 & 0.7794 & 0.6394 & 0.7552 & 0.6481 & 0.7683 & 0.6774 & 0.7683 & 0.6610 & 0.7633 & \cellcolor{cyan!10}\textbf{0.7378} & \cellcolor{cyan!10}\textbf{0.8019} \\
        & Full & 90\% & 0.6824 & 0.7795 & 0.6800 & 0.7863 & \cellcolor{orange!12}\underline{0.6854} & \cellcolor{orange!12}\underline{0.7894} & 0.6708 & 0.7685 & 0.6105 & 0.7486 & 0.6236 & 0.7537 & 0.6619 & 0.7570 & 0.6495 & 0.7584 & \cellcolor{cyan!10}\textbf{0.7349} & \cellcolor{cyan!10}\textbf{0.7990} \\
        & Full & Full & 0.7240 & 0.7933 & 0.7273 & 0.7906 & 0.7278 & 0.7982 & 0.7219 & 0.7916 & 0.6846 & 0.7746 & 0.6929 & 0.7796 & \cellcolor{orange!12}\underline{0.7354} & \cellcolor{orange!12}\underline{0.8010} & 0.7309 & 0.7995 & \cellcolor{cyan!10}\textbf{0.7499} & \cellcolor{cyan!10}\textbf{0.8061} \\
    \bottomrule
    \end{tabular}
    }
    \label{tab:double_modals_crackpolar}
    \vspace{-0.3cm}
\end{table*}

\begin{figure*}[!t]
  \centering
  \includegraphics[width=\textwidth]{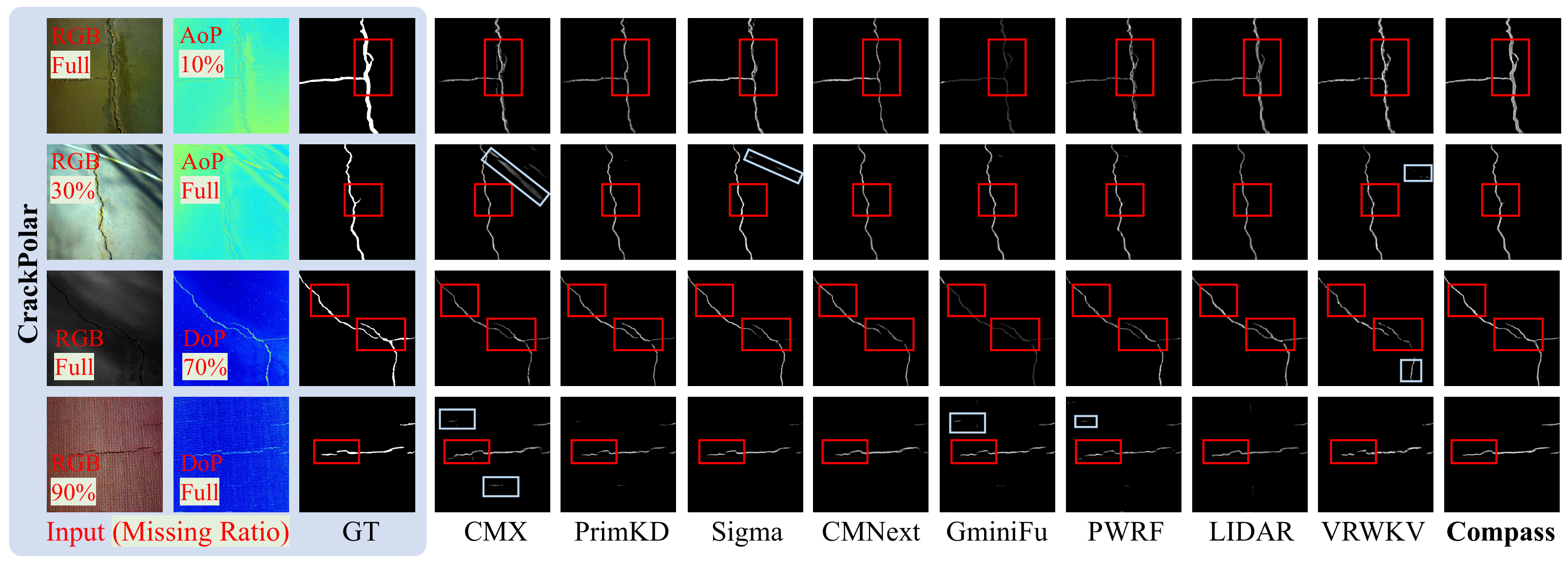}
  \caption{Visual comparison under dual-modal and multi missing settings on the CrackPolar dataset. Red boxes denote key regions; blue denote misidentifications.}
  \label{fig:Crackpolar_dual_modals_results}
  \vspace{-0.4cm}
\end{figure*}

\section{Objective Function Details}

Training proceeds in two phases. The warmup phase uses only the original stream for standard segmentation training to establish stable Backbone feature representations. After warmup, the dual-stream phase is activated with the degradation simulation stream and all distillation losses.

The segmentation loss combines binary cross-entropy loss and Dice loss \cite{Liu2025SCSegamba}. Binary cross-entropy measures the per-pixel discrepancy between predicted probabilities and ground truth labels:
\begin{equation}
\mathcal{L}_{BCE} = -\frac{1}{N}\sum_{j=1}^{N}\left[y_j \log \hat{y}_j + \left(1 - y_j\right) \log\left(1 - \hat{y}_j\right)\right],
\end{equation}
where $N$ is the total number of pixels, $y_j \in \{0, 1\}$ is the ground truth label of the $j$-th pixel, and $\hat{y}_j \in [0, 1]$ is the predicted probability. Dice loss measures the overlap between predicted and ground truth regions from a set similarity perspective, which is particularly important for crack segmentation where the foreground ratio is extremely low:
\begin{equation}
\mathcal{L}_{Dice} = 1 - \frac{2\sum_{j=1}^{N} y_j \hat{y}_j + \epsilon}{\sum_{j=1}^{N} y_j + \sum_{j=1}^{N} \hat{y}_j + \epsilon},
\end{equation}
where $\epsilon$ is a smoothing constant preventing division by zero. Their combination $\mathcal{L}_{seg} = \mathcal{L}_{BCE} + \mathcal{L}_{Dice}$ is computed separately for the original and degradation simulation streams, denoted as $\mathcal{L}_{seg}^{o}$ and $\mathcal{L}_{seg}^{d}$ respectively. The total loss is the sum of all loss terms:
\begin{equation}
\mathcal{L} = \mathcal{L}_{seg}^{o} + \mathcal{L}_{seg}^{d} + \mathcal{L}_{IMD} + \mathcal{L}_{MGC} + \mathcal{L}_{FC},
\end{equation}
where $\mathcal{L}_{IMD}$, $\mathcal{L}_{MGC}$, and $\mathcal{L}_{FC}$ are the interactive mutual distillation loss, multi-granularity consistency loss, and feature completion loss detailed in Section~\ref{sec:dsd}. The warmup phase uses only $\mathcal{L}_{seg}^{o}$. Gradient clipping is applied to ensure numerical stability during dual-stream training.

\begin{table*}[!t]
    \centering

    \scriptsize
    \setlength{\tabcolsep}{2pt}
    \renewcommand{\arraystretch}{0.9}
    \caption{Performance comparison under all seven-modal inputs and different missing ratio settings on the CrackPolar dataset. {\setlength{\fboxsep}{1pt}\colorbox{cyan!10}{\textbf{Blue}}} and {\setlength{\fboxsep}{1pt}\colorbox{orange!12}{\underline{Orange}}} denote best and second-best.}
    \resizebox{\textwidth}{!}{
    \begin{tabular}{c|c|c|c|*{6}{cc}cc}
    \toprule
    \multicolumn{4}{c|}{Missing Ratio} 
      & \multicolumn{2}{c}{CMNeXT \cite{Zhang2023Delivering}} & \multicolumn{2}{c}{GminiFusion \cite{Jia2024GeminiFusion}} & \multicolumn{2}{c}{mmsFormer \cite{Reza2024MMSFormer}} & \multicolumn{2}{c}{PWRF \cite{Liu2025Part}} & \multicolumn{2}{c}{LIDAR \cite{Liu2025LIDAR}} & \multicolumn{2}{c}{VRWKV \cite{Duan2025Vision}} & \multicolumn{2}{c}{\textbf{Compass}} \\
    \midrule
        RGB & AoP & DoP & [$0^\circ$, $45^\circ$, $90^\circ$, $135^\circ$] 
        & F1 & mIoU & F1 & mIoU & F1 & mIoU & F1 & mIoU & F1 & mIoU & F1 & mIoU & F1 & mIoU \\ 
    \midrule
        Full & 10\% & 30\% & 70\% & \cellcolor{orange!12}\underline{0.7317} & \cellcolor{orange!12}\underline{0.8018} & 0.6402 & 0.7512 & 0.7213 & 0.7968 & 0.7122 & 0.7849 & 0.7178 & 0.7897 & 0.7118 & 0.7844 & \cellcolor{cyan!10}\textbf{0.7538} & \cellcolor{cyan!10}\textbf{0.8089} \\
        Full & 30\% & 50\% & 90\% & \cellcolor{orange!12}\underline{0.7213} & 0.7905 & 0.6671 & 0.7540 & 0.7107 & \cellcolor{orange!12}\underline{0.7926} & 0.7117 & 0.7869 & 0.7130 & 0.7863 & 0.7016 & 0.7886 & \cellcolor{cyan!10}\textbf{0.7497} & \cellcolor{cyan!10}\textbf{0.8085} \\
        30\% & Full & 70\% & 10\% & \cellcolor{orange!12}\underline{0.7315} & 0.7984 & 0.7090 & 0.7852 & 0.7175 & \cellcolor{orange!12}\underline{0.8005} & 0.7196 & 0.7903 & 0.7200 & 0.7885 & 0.7184 & 0.7941 & \cellcolor{cyan!10}\textbf{0.7550} & \cellcolor{cyan!10}\textbf{0.8086} \\
        10\% & 70\% & Full & 30\% & \cellcolor{orange!12}\underline{0.7380} & \cellcolor{orange!12}\underline{0.8031} & 0.7055 & 0.7869 & 0.7131 & 0.7981 & 0.7106 & 0.7881 & 0.6959 & 0.7746 & 0.7143 & 0.7863 & \cellcolor{cyan!10}\textbf{0.7526} & \cellcolor{cyan!10}\textbf{0.8075} \\
        70\% & 30\% & 10\% & Full & \cellcolor{orange!12}\underline{0.7233} & \cellcolor{orange!12}\underline{0.7991} & 0.7031 & 0.7810 & 0.7190 & 0.7986 & 0.7192 & 0.7886 & 0.7152 & 0.7880 & 0.7127 & 0.7919 & \cellcolor{cyan!10}\textbf{0.7507} & \cellcolor{cyan!10}\textbf{0.8086} \\
        Full & Full & 30\% & 30\% & \cellcolor{orange!12}\underline{0.7303} & \cellcolor{orange!12}\underline{0.8058} & 0.6978 & 0.7771 & 0.7207 & 0.7970 & 0.7162 & 0.7891 & 0.7173 & 0.7899 & 0.7183 & 0.7937 & \cellcolor{cyan!10}\textbf{0.7533} & \cellcolor{cyan!10}\textbf{0.8082} \\
        Full & 30\% & Full & 30\% & 0.7195 & 0.7895 & 0.7000 & 0.7757 & \cellcolor{orange!12}\underline{0.7233} & \cellcolor{orange!12}\underline{0.7992} & 0.7165 & 0.7903 & 0.7187 & 0.7893 & 0.7190 & 0.7932 & \cellcolor{cyan!10}\textbf{0.7515} & \cellcolor{cyan!10}\textbf{0.8065} \\
        30\% & Full & Full & 30\% & \cellcolor{orange!12}\underline{0.7242} & 0.7948 & 0.7179 & 0.7919 & 0.7206 & 0.7947 & 0.7171 & 0.7899 & 0.7152 & 0.7839 & 0.7184 & \cellcolor{orange!12}\underline{0.7968} & \cellcolor{cyan!10}\textbf{0.7512} & \cellcolor{cyan!10}\textbf{0.8077} \\
        90\% & Full & Full & Full & \cellcolor{orange!12}\underline{0.7340} & \cellcolor{orange!12}\underline{0.8016} & 0.7248 & 0.7966 & 0.7101 & 0.7947 & 0.7156 & 0.7884 & 0.7184 & 0.7879 & 0.7059 & 0.7861 & \cellcolor{cyan!10}\textbf{0.7539} & \cellcolor{cyan!10}\textbf{0.8093} \\
        Full & Full & 90\% & Full & \cellcolor{orange!12}\underline{0.7214} & \cellcolor{orange!12}\underline{0.7993} & 0.6096 & 0.7424 & 0.7190 & 0.7978 & 0.7178 & 0.7900 & 0.7059 & 0.7799 & 0.7210 & 0.7983 & \cellcolor{cyan!10}\textbf{0.7544} & \cellcolor{cyan!10}\textbf{0.8080} \\
        10\% & 30\% & 70\% & 90\% & \cellcolor{orange!12}\underline{0.7118} & \cellcolor{orange!12}\underline{0.7972} & 0.6858 & 0.7711 & 0.7048 & 0.7888 & 0.6672 & 0.7694 & 0.6804 & 0.7546 & 0.6917 & 0.7780 & \cellcolor{cyan!10}\textbf{0.7445} & \cellcolor{cyan!10}\textbf{0.8021} \\
        30\% & 50\% & 90\% & 10\% & \cellcolor{orange!12}\underline{0.7235} & \cellcolor{orange!12}\underline{0.7923} & 0.6916 & 0.7798 & 0.6954 & 0.7890 & 0.6746 & 0.7762 & 0.6959 & 0.7781 & 0.6846 & 0.7744 & \cellcolor{cyan!10}\textbf{0.7486} & \cellcolor{cyan!10}\textbf{0.8073} \\
        90\% & 70\% & 30\% & 10\% & \cellcolor{orange!12}\underline{0.7256} & 0.7958 & 0.6899 & 0.7771 & 0.7167 & \cellcolor{orange!12}\underline{0.7986} & 0.7205 & 0.7885 & 0.7168 & 0.7888 & 0.6894 & 0.7781 & \cellcolor{cyan!10}\textbf{0.7494} & \cellcolor{cyan!10}\textbf{0.8094} \\
        Full & Full & Full & Full & 0.7391 & 0.8029 & 0.7051 & 0.7848 & 0.7264 & 0.7998 & 0.7219 & 0.7921 & \cellcolor{orange!12}\underline{0.7451} & \cellcolor{orange!12}\underline{0.8037} & 0.7361 & 0.8016 & \cellcolor{cyan!10}\textbf{0.7591} & \cellcolor{cyan!10}\textbf{0.8105} \\
    \bottomrule
    \end{tabular}
    }
    \label{tab:CrackPolar_seven_modal_results}
\end{table*}

\begin{figure*}[!t]
  \centering
  \includegraphics[width=\textwidth]{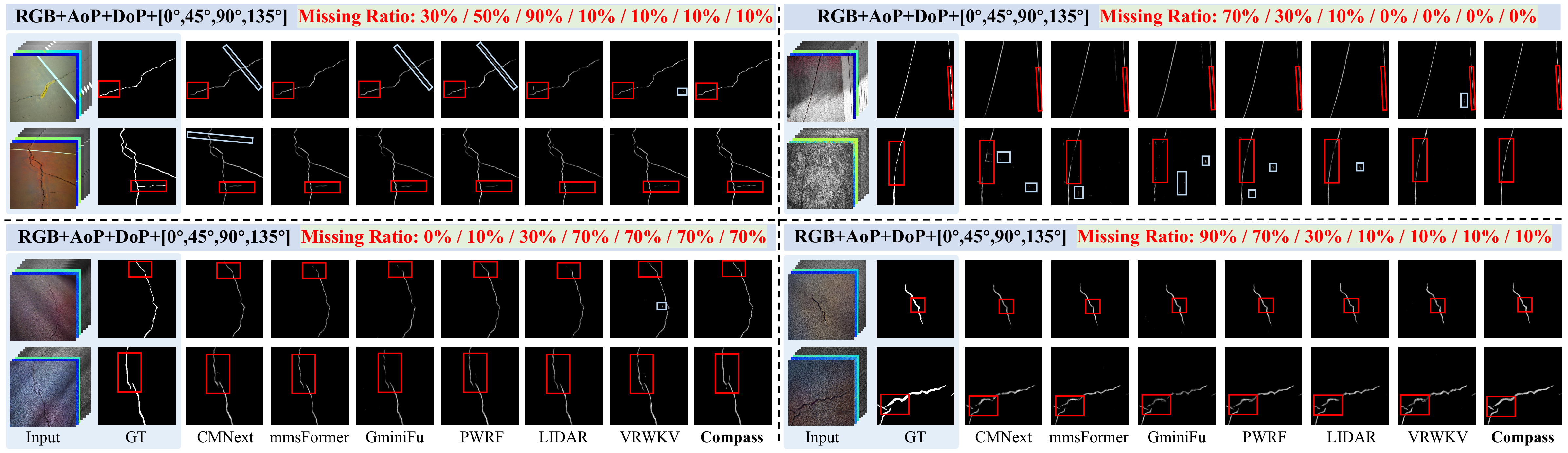}
  \caption{Visual comparison under seven-modal and multimissing settings on the CrackPolar dataset. Red boxes denote key regions; blue denote misidentifications.}
  \label{fig:Crackpolar_seven_modals_results}
  \vspace{-0.4cm}
\end{figure*}

\vspace{-0.2cm}
\section{More Multimodal Comparative Experiments}

\subsection{CrackPolar Dual-Modal Experiments}

To more comprehensively evaluate the generalization capability of Compass under different input conditions, we conduct experiments on the CrackPolar \cite{Liu2025LIDAR} dataset under two dual-modal input settings, namely RGB with DoP and RGB with AoP. As shown in Table \ref{tab:double_modals} and Figure \ref{fig:dual_results}, Compass achieves superior results under nearly all missing-modality configurations.

In the RGB and DoP combination, when the RGB missing rate reaches 90\%, Compass achieves an F1 score of 0.7168, surpassing the second-best method PrimKD~\cite{Hao2024PrimKD} by 5.54\%. Under the condition where the auxiliary modality DoP is missing at 90\%, Compass achieves an F1 score of 0.7241, outperforming Sigma~\cite{Wan2025Sigma} by 5.76\%. As illustrated in Figure \ref{fig:dual_results}, when DoP is missing at 70\%, most competing methods exhibit crack-tip omissions, whereas Compass accurately preserves the complete crack trajectory. Even under the no-missing setting, Compass surpasses LIDAR~\cite{Liu2025LIDAR} by 2.21\% and 1.15\% in F1 and mIoU, respectively.

In the RGB and AoP combination, the advantages of Compass are even more pronounced. When the RGB missing rate is 70\%, Compass outperforms the second-best method CMNeXT~\cite{Zhang2023Delivering} by 7.34\% and 3.87\% in F1 and mIoU, respectively. Under the extreme 90\% missing condition, Compass exceeds LIDAR~\cite{Liu2025LIDAR} by 40.57\% in F1, indicating that LIDAR suffers catastrophic performance collapse when the primary modality is severely degraded, whereas Compass maintains robust performance at inference time owing to the degradation-simulated reciprocal learning paradigm of DSD, which actively exposes the model to comparable or even more severe degradation conditions during training. When the AoP missing rate is 90\%, Compass outperforms Sigma~\cite{Liu2025LIDAR} by 7.21\% in F1. As shown in Figure \ref{fig:dual_results}, when RGB is missing at 90\%, the segmentation quality of CMX~\cite{Zhang2023CMX} and PrimKD~\cite{Hao2024PrimKD} degrades severely, while Compass still produces topologically connected segmentation results. This is attributed to FAPT reconstructing semantically complete feature representations from residual AoP polarization information via semantic prototypes, and the uncertainty-gated decoder of ETPF automatically suppressing noise propagation in highly degraded regions. Under the no-missing setting, Compass outperforms LIDAR~\cite{Liu2025LIDAR} by 1.97\% and 0.64\% in F1 and mIoU, further demonstrating that the GCM within Needle remains superior to existing backbones under complete data conditions.

\begin{table}[!t]
    \centering
    \tiny
    \setlength{\tabcolsep}{4pt}
    \renewcommand{\arraystretch}{1}
    \caption{Complexity comparison of seven-modal 512  times 512 input. {\setlength{\fboxsep}{1pt}\colorbox{cyan!10}{\textbf{Blue}}} and {\setlength{\fboxsep}{1pt}\colorbox{orange!12}{\underline{Orange}}} denote best and second-best results.}
    \resizebox{\linewidth}{!}{
    \begin{tabular}{ccccc}
    \hline
    Method & Year & FLOPs$\downarrow$ & Params$\downarrow$ & Size$\downarrow$ \\ \hline
    CMNeXT \cite{Zhang2023Delivering} & CVPR 2023 & \cellcolor{cyan!10}\textbf{50.92G} & 57.69M & 660MB \\
    mmsFormer \cite{Reza2024MMSFormer} & OJSP 2024 & 93.50G & 86.68M & 1034MB \\
    GminiFu \cite{Jia2024GeminiFusion} & ICML 2024 & \cellcolor{orange!12}\underline{51.75G} & 29.27M & 358MB \\
    PWRF \cite{Liu2025Part} & IJCV 2025 & 451.58G & 710.67M & 7024MB \\
    LIDAR \cite{Liu2025LIDAR} & MM 2025 & 116.68G & \cellcolor{orange!12}\underline{18.68M} & \cellcolor{orange!12}\underline{270MB} \\
    VRWKV \cite{Duan2025Vision} & ICLR 2025 & 336.34G & 32.84M & 418MB \\
    \textbf{Compass (Ours)} & N/A & 96.59G & \cellcolor{cyan!10}\textbf{8.69M} & \cellcolor{cyan!10}\textbf{58MB} \\ \hline
    \end{tabular}
    }
    \label{tab:Complexity_comparison_under_seven_modals}
    \vspace{-0.6cm}
\end{table}

\begin{figure*}[!t]
  \centering
  \includegraphics[width=0.8\textwidth]{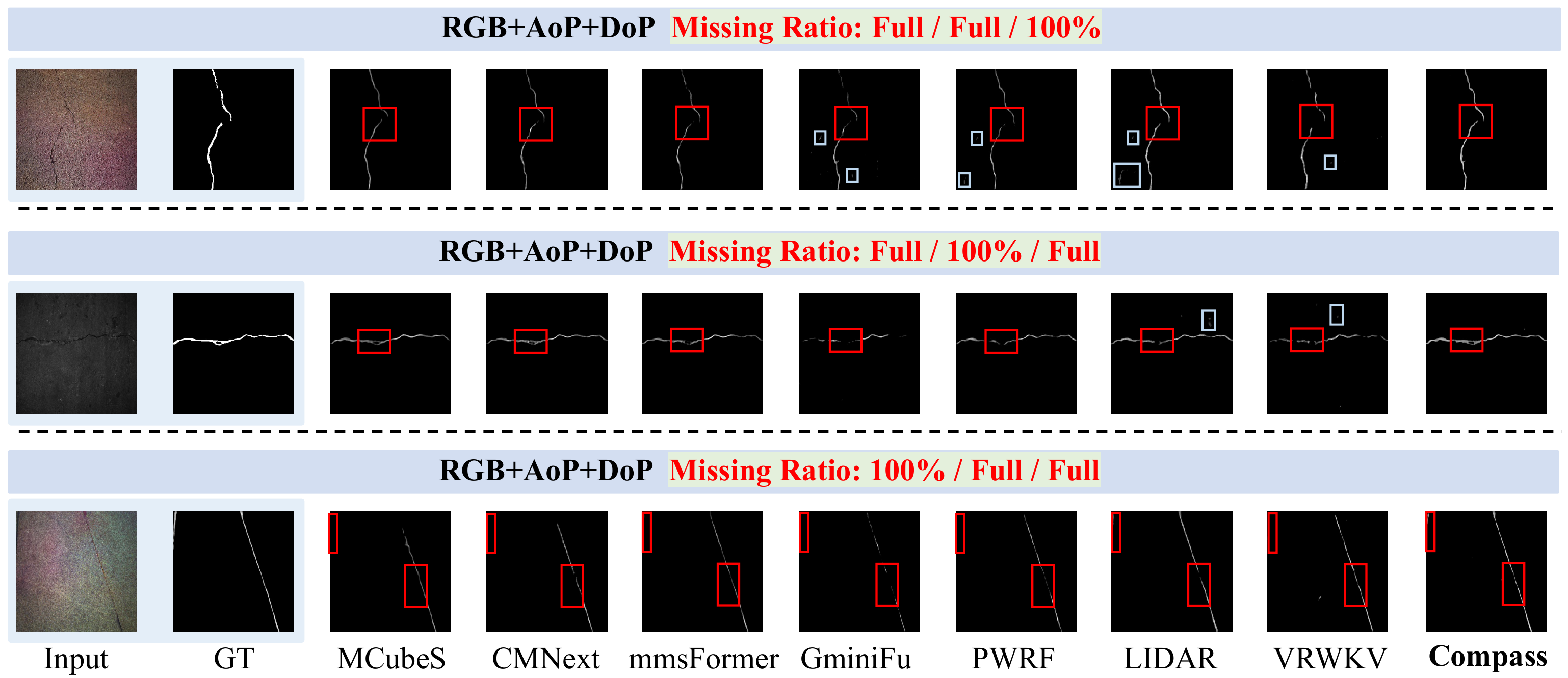}
  \caption{Visual comparison under seven-modal and extreme missing settings on the CrackPolar dataset. Red boxes denote key regions; blue denote misidentifications.}
  \label{fig:Crackpolar_seven_modals_results_100}
  \vspace{-0.3cm}
\end{figure*}

\subsection{CrackPolar Seven-Modal Experiments}

To validate the performance of Compass under large-scale modal inputs, we conduct comparative experiments on the CrackPolar \cite{Liu2025LIDAR} dataset using all seven modalities, comprising RGB, AoP, DoP, and four polarization angle images captured at 0°, 45°, 90°, and 135°. The AoP modality reflects the spatial distribution of light wave oscillation directions and is sensitive to surface microstructure and orientation variations. The DoP modality characterizes the degree of polarization purity of light and is closely related to surface reflection properties and roughness. The four polarization angle images individually record the intensity distribution of light under four polarization orientations, capturing the anisotropic reflection characteristics of the surface with respect to incident light. Comparison methods include CMNeXT~\cite{Zhang2023Delivering}, GeminiFusion~\cite{Jia2024GeminiFusion}, mmsFormer~\cite{Reza2024MMSFormer}, PWRF~\cite{Liu2025Part}, LIDAR~\cite{Liu2025LIDAR}, and Vision RWKV~\cite{Duan2025Vision}.

\begin{table*}[!t]
    \centering
    \setlength{\heavyrulewidth}{0.5pt}
    \setlength{\lightrulewidth}{0.3pt}
    \setlength{\cmidrulewidth}{0.2pt}
    \setlength{\tabcolsep}{4pt}
    \renewcommand{\arraystretch}{0.88}
    \caption{Performance comparison under tri-modal and extreme missing. {\setlength{\fboxsep}{1pt}\colorbox{cyan!10}{\textbf{Blue}}} and {\setlength{\fboxsep}{1pt}\colorbox{orange!12}{\underline{Orange}}} denote best and second-best results.}
    \resizebox{\linewidth}{!}{
    \begin{tabular}{ccc|*{6}{cc}cc}
    \toprule
    \multicolumn{3}{c|}{Missing Ratio}
      & \multicolumn{2}{c}{GminiFusion \cite{Jia2024GeminiFusion}} & \multicolumn{2}{c}{mmsFormer \cite{Reza2024MMSFormer}} & \multicolumn{2}{c}{PWRF \cite{Liu2025Part}} & \multicolumn{2}{c}{CMNeXT \cite{Zhang2023Delivering}} & \multicolumn{2}{c}{LIDAR \cite{Liu2025LIDAR}} & \multicolumn{2}{c}{VRWKV \cite{Duan2025Vision}} & \multicolumn{2}{c}{\textbf{Compass}} \\
    \midrule
    RGB & AoP & DoP & F1 & mIoU & F1 & mIoU & F1 & mIoU & F1 & mIoU & F1 & mIoU & F1 & mIoU & F1 & mIoU \\
    \midrule
    Full & Full & 100\% & 0.6406 & 0.7453 & 0.6978 & 0.7892 & 0.6752 & 0.7740 & 0.7112 & \cellcolor{orange!12}\underline{0.7928} & 0.7065 & 0.7840 & \cellcolor{orange!12}\underline{0.7127} & 0.7883 & \cellcolor{cyan!10}\textbf{0.7317} & \cellcolor{cyan!10}\textbf{0.7956} \\
    Full & 100\% & Full & 0.6368 & 0.7470 & 0.6970 & 0.7835 & 0.6827 & 0.7761 & \cellcolor{orange!12}\underline{0.7139} & \cellcolor{orange!12}\underline{0.7944} & 0.7097 & 0.7858 & 0.7103 & 0.7892 & \cellcolor{cyan!10}\textbf{0.7383} & \cellcolor{cyan!10}\textbf{0.8011} \\
    100\% & Full & Full & 0.6131 & 0.7424 & 0.6839 & 0.7822 & 0.6766 & 0.7753 & 0.6926 & \cellcolor{orange!12}\underline{0.7854} & 0.6886 & 0.7717 & \cellcolor{orange!12}\underline{0.7052} & 0.7823 & \cellcolor{cyan!10}\textbf{0.7303} & \cellcolor{cyan!10}\textbf{0.7981} \\
    Full & Full & Full & 0.6920 & 0.7794 & 0.7265 & 0.7952 & 0.7063 & 0.7837 & 0.7124 & 0.7875 & \cellcolor{orange!12}\underline{0.7353} & \cellcolor{orange!12}\underline{0.7976} & 0.7291 & 0.7919 & \cellcolor{cyan!10}\textbf{0.7503} & \cellcolor{cyan!10}\textbf{0.8055} \\
    \bottomrule
    \end{tabular}
    }
    \label{tab:tri_modals_100}
    \vspace{-0.3cm}
\end{table*}

As shown in Table \ref{tab:CrackPolar_seven_modal_results} and Figure \ref{fig:Crackpolar_seven_modals_results}, Compass achieves the best results under all missing configurations. When only RGB is complete and the remaining six modalities are missing at rates of 10\%, 30\%, and 70\%, respectively, Compass achieves an F1 of 0.7538 and an mIoU of 0.8089, outperforming CMNeXT~\cite{Zhang2023Delivering} by 3.03\% and 0.89\%. When the RGB missing rate reaches 90\% while the remaining modalities decrease in missing rate at 70\%, 30\%, and 10\%, Compass outperforms CMNeXT~\cite{Zhang2023Delivering} and mmsFormer~\cite{Reza2024MMSFormer} by 3.28\% and 4.56\% in F1, respectively. Under the extreme condition where all seven modalities suffer from varying degrees of missing at rates of 10\%, 30\%, 70\%, and 90\%, Compass achieves an F1 of 0.7445, surpassing CMNeXT~\cite{Zhang2023Delivering} by 4.60\% and LIDAR~\cite{Liu2025LIDAR} by 9.42\%. As illustrated in Figure \ref{fig:Crackpolar_seven_modals_results}, under this extreme condition most methods produce severe structural discontinuities and omissions, while Compass still generates continuous segmentation results in critical regions. This demonstrates that the degradation-simulated reciprocal learning paradigm of DSD enables the model to proactively adapt to degradation patterns arising from arbitrary modality combinations during training, that the semantic prototype mechanism of FAPT can effectively reconstruct semantically complete features from residual information in the seven-modal setting, and that the Dempster-Shafer evidential combination in ETPF performs pairwise reliability assessment and fusion across all seven modalities, ensuring fusion stability when a large number of modalities are simultaneously degraded. Under the no-missing setting, Compass outperforms LIDAR~\cite{Liu2025LIDAR} by 1.88\% and 0.85\% in F1 and mIoU, respectively.

\begin{table*}[!t]
    \centering
    \scriptsize
    \setlength{\tabcolsep}{4pt}
    \renewcommand{\arraystretch}{0.9}
    \caption{Performance comparison under single RGB modal input. {\setlength{\fboxsep}{1pt}\colorbox{cyan!10}{\textbf{Blue}}} and {\setlength{\fboxsep}{1pt}\colorbox{orange!12}{\underline{Orange}}} denote best and second-best results.}
    \resizebox{0.8\textwidth}{!}{
    \begin{tabular}{c|cc|cc|cc|ccc}
    \toprule
        \multirow{2}{*}{Method} 
        & \multicolumn{2}{c|}{CrackPolar \cite{Liu2025LIDAR}} 
        & \multicolumn{2}{c|}{IRTCrack \cite{liu2022asphalt}} 
        & \multicolumn{2}{c|}{CrackDepth \cite{Liu2025LIDAR}} 
        & \multirow{2}{*}{Params$\downarrow$} 
        & \multirow{2}{*}{FLOPs$\downarrow$} 
        & \multirow{2}{*}{Size$\downarrow$} \\ 
        \cmidrule(lr){2-7}
        & F1 & mIoU & F1 & mIoU & F1 & mIoU &  &  &  \\ 
    \midrule
        CrackFormer \cite{Liu2021CrackFormer} & 0.7104 & 0.7729 & 0.8204 & 0.8185 & \cellcolor{orange!12}\underline{0.8059} & \cellcolor{orange!12}\underline{0.8393} & 4.55M & 81.85G & 55MB \\
        SimCrack \cite{Jaziri2024Designing} & 0.7125 & 0.7889 & 0.8142 & 0.8242 & 0.7940 & 0.8291 & 29.58M & 286.62G & 225MB \\
        MambaIR \cite{Guo2024MambaIR} & 0.7063 & 0.7777 & \cellcolor{orange!12}\underline{0.8297} & \cellcolor{orange!12}\underline{0.8294} & 0.7912 & 0.8254 & 10.34M & 47.32G & 79MB \\
        CSMamba \cite{liu2024cmunet} & 0.6428 & 0.7342 & 0.7909 & 0.7925 & 0.7238 & 0.7753 & 35.95M & 145.85G & 233MB \\
        PlainMamba \cite{Yang2024PlainMamba} & 0.7161 & 0.7824 & 0.8294 & 0.8277 & 0.8013 & 0.8306 & 16.72M & 73.86G & 96MB \\
        SCSegamba \cite{Liu2025SCSegamba} & \cellcolor{orange!12}\underline{0.7173} & \cellcolor{orange!12}\underline{0.7925} & 0.8214 & 0.8284 & 0.8008 & 0.8342 & \cellcolor{orange!12}\underline{2.80M} & \cellcolor{orange!12}\underline{18.16G} & \cellcolor{orange!12}\underline{37MB} \\
        \textbf{Compass (Ours)} & \cellcolor{cyan!10}\textbf{0.7381} & \cellcolor{cyan!10}\textbf{0.8007} & \cellcolor{cyan!10}\textbf{0.8446} & \cellcolor{cyan!10}\textbf{0.8468} & \cellcolor{cyan!10}\textbf{0.8289} & \cellcolor{cyan!10}\textbf{0.8488} & \cellcolor{cyan!10}\textbf{1.35M} & \cellcolor{cyan!10}\textbf{15.94G} & \cellcolor{cyan!10}\textbf{21MB} \\
    \bottomrule
    \end{tabular}
    }
    \label{tab:single_modal_compare}
\end{table*}

\begin{figure*}[!t]
  \centering
  \includegraphics[width=0.85\textwidth]{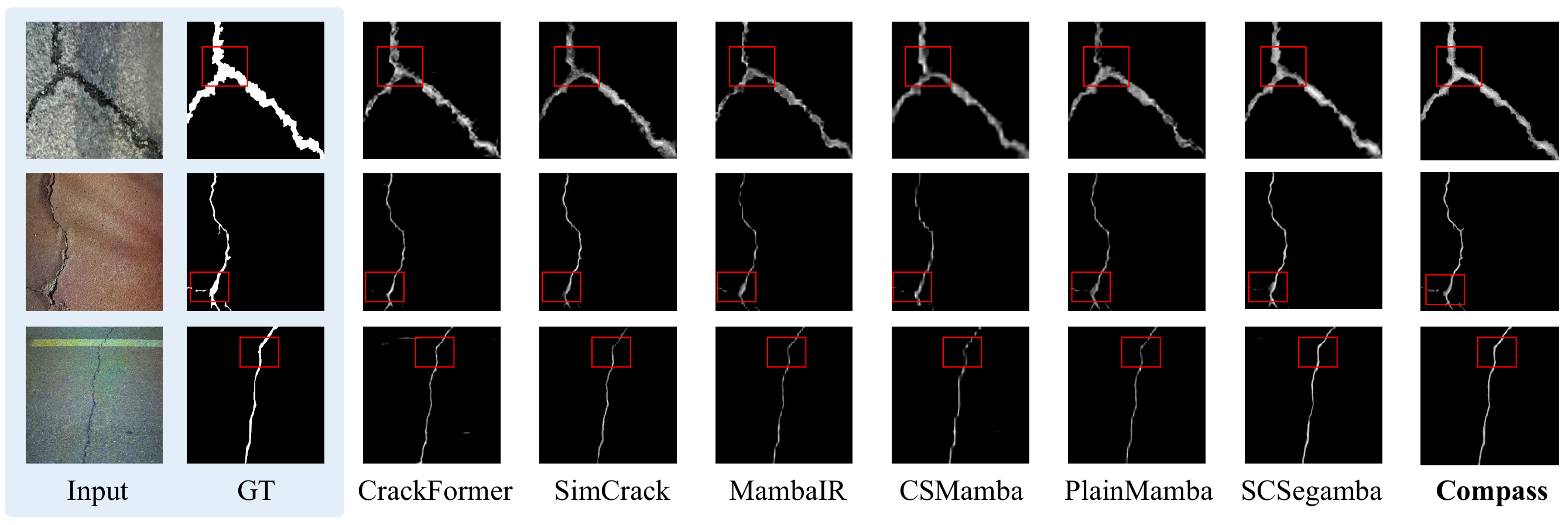}
  \caption{Visual comparison under signle RGB modal. Red boxes denote key regions.}
  \label{fig:signle_modal_results}
  \vspace{-0.3cm}
\end{figure*}

As shown in Table \ref{tab:Complexity_comparison_under_seven_modals}, even under seven-modal input, Compass requires only 8.69M parameters and a model size of 58MB, which is 53.48\% and 78.52\% lower than LIDAR~\cite{Liu2025LIDAR}, and reduces the parameter count and model size by 82 times and 121 times compared to PWRF~\cite{Liu2025Part}. This efficiency stems from the lightweight architecture design of Needle, which performs sequential modeling in extremely low-dimensional subspaces at each layer. The model size remains constantly at 58MB as the number of modalities scales from dual-modal to seven-modal, with only the Needle branches and FAPT completion channels growing linearly with the number of modalities, demonstrating excellent scalability and potential for edge deployment.

\vspace{-0.5cm}
\subsection{Experiments with Extreme Missing Modalities on CrackPolar}
To evaluate Compass's robustness when an entire modality is unavailable, we conduct experiments under three single-modality complete dropout settings on CrackPolar. As shown in Table~\ref{tab:tri_modals_100} and Figure~\ref{fig:Crackpolar_seven_modals_results_100}, Compass achieves the best performance across all settings.

With RGB and AoP complete but DoP entirely missing, Compass achieves F1 of 0.7317 and mIoU of 0.7956, surpassing Vision RWKV~\cite{Duan2025Vision} by 1.90\% and CMNeXT~\cite{Zhang2023Delivering} by 2.05\% in F1. When AoP is entirely missing with RGB and DoP complete, Compass reaches F1 of 0.7383 and mIoU of 0.8011, exceeding CMNeXT~\cite{Zhang2023Delivering} by 2.44\% in F1 and 0.67\% in mIoU. These results confirm that DSD's degradation simulation reciprocal learning has exposed the model to comparable or even harsher missing conditions during training, enabling FAPT to reconstruct the missing polarization semantics from the remaining two modalities through modality-agnostic prototypes.

In the most challenging setting where the primary RGB modality is entirely missing with only AoP and DoP retained, Compass still attains F1 of 0.7303 and mIoU of 0.7981, surpassing CMNeXT~\cite{Zhang2023Delivering} by 3.77\% and LIDAR~\cite{Liu2025LIDAR} by 4.17\% in F1. As shown in Figure~\ref{fig:Crackpolar_seven_modals_results_100}, GeminiFusion~\cite{Jia2024GeminiFusion} and LIDAR~\cite{Liu2025LIDAR} suffer severe crack fragmentation and omission under this condition, while CMNeXT~\cite{Zhang2023Delivering} and mmsFormer~\cite{Reza2024MMSFormer} exhibit structural loss at crack endpoints; in contrast, Compass still generates clear and continuous segmentation results. This demonstrates that ETPF's Dempster-Shafer evidential combination accurately assesses polarization modality reliability to regulate fusion weights, anisotropic evidence propagation diffuses belief along crack directions to preserve topological connectivity, and the uncertainty-gated decoder suppresses high-uncertainty regions caused by complete RGB absence. Under full modalities, Compass achieves F1 of 0.7503 and mIoU of 0.8055, exceeding LIDAR~\cite{Liu2025LIDAR} by 1.50\% and 0.79\% respectively.

\vspace{-0.3cm}
\subsection{Experiments on Single-Modal Input}

To validate the performance of Compass under single RGB  input, we compare it against SOTA single-modal crack segmentation methods on CrackPolar, IRTCrack, and CrackDepth, including CrackFormer~\cite{Liu2021CrackFormer}, SimCrack~\cite{Jaziri2024Designing}, MambaIR~\cite{Guo2024MambaIR}, CSMamba~\cite{liu2024cmunet}, PlainMamba~\cite{Yang2024PlainMamba}, and SCSegamba~\cite{Liu2025SCSegamba}. As shown in Table \ref{tab:single_modal_compare} and Figure \ref{fig:signle_modal_results}, when only the RGB modality is available, Compass achieves the best segmentation performance on all three datasets while incurring the lowest computational overhead.

On the CrackPolar and CrackDepth datasets \cite{Liu2025LIDAR}, which feature diverse scenes and heavy background interference, Compass achieves F1 of 0.7381 and 0.8289, outperforming SCSegamba~\cite{Liu2025SCSegamba}, which is specifically designed for crack segmentation, by 2.90\% and 3.50\% in F1 and by 1.03\% and 1.74\% in mIoU, respectively. As illustrated in Figure \ref{fig:signle_modal_results}, in regions with subtle cracks and severe interference, Compass produces more continuous and clearer segmentation results than competing methods, indicating that the orientation-aware key modulation of GCM and the connectivity gating of CohGate within Needle can effectively perceive the anisotropic topological structure of cracks even under single-modal conditions, enhancing information propagation along crack orientations to maintain topological connectivity. On the IRTCrack \cite{liu2022asphalt} dataset, where crack regions are relatively large and boundaries are blurred, Compass achieves F1 and mIoU scores of 0.8446 and 0.8468, outperforming SCSegamba~\cite{Liu2025SCSegamba} by 2.83\% and 2.22\% and MambaIR~\cite{Guo2024MambaIR} by 1.80\% and 2.09\%, demonstrating that the AGP module of Needle adaptively selects the optimal receptive field through anisotropic context probing to effectively capture the boundary information of wide cracks.

In terms of computational complexity, Compass requires only 1.35M parameters, 15.94G FLOPs, and a model size of 21MB, reducing the parameter count, FLOPs, and model size by 51.79\%, 12.22\%, and 43.24\% compared to SCSegamba~\cite{Liu2025SCSegamba}, and reducing the parameter count and FLOPs by 22 times and 18 times compared to SimCrack~\cite{Jaziri2024Designing}. These results demonstrate that the lightweight architecture design of Needle not only applies to multimodal scenarios but also exhibits outstanding efficiency advantages under single-modal conditions.

\vspace{-0.3cm}
\section{More Ablation and Sensitivity Studies}

\subsection{Ablation Study on Individual Components}

As shown in Table \ref{tab:abl_component}, we evaluate the contribution of each core component by sequentially adding and removing FAPT, AGP, and CohGate. When all three components are enabled, Compass achieves the best performance in both F1 and mIoU.

Removing FAPT leads to a decrease of 1.59\% in F1 and 0.22\% in mIoU, while the introduction of FAPT incurs only 0.01M additional parameters and 0.01G FLOPs. This demonstrates that FAPT achieves effective semantic completion of missing modalities at an extremely low computational cost, and its modality-agnostic semantic prototype mechanism provides critical support for feature reconstruction under degraded conditions. Removing AGP results in a decrease of 1.62\% in F1 and 0.55\% in mIoU, indicating that the anisotropic context probing of AGP plays an important role in adaptively selecting the optimal receptive field and capturing the directional structural information of cracks across different scales. Removing CohGate leads to a decrease of 1.05\% in F1, demonstrating that the connectivity-aware gating of CohGate effectively enhances the cooperative activation of connected crack regions by taking into account the spatial consistency of local neighborhoods, thereby suppressing isolated noise.

When only a single component is used, performance degrades more significantly. The variant using only AGP achieves the lowest F1 of 0.7290, a decrease of 2.46\% compared to the full model, indicating that isolated context probing without missing-modality completion and connectivity gating is insufficient to handle complex degradation scenarios. The three components operate synergistically across the dimensions of modality completion, orientation awareness, and connectivity constraint, jointly ensuring the segmentation accuracy and topological integrity of Compass under various missing conditions. The full model requires only 3.80M parameters and 42.82G FLOPs, maintaining extremely low computational overhead.

\begin{table}[!t]
\setlength{\tabcolsep}{2.5pt}
\caption{Performance comparison of different components in ETPF. {\setlength{\fboxsep}{1pt}\colorbox{cyan!10}{\textbf{Blue}}} and {\setlength{\fboxsep}{1pt}\colorbox{orange!12}{\underline{Orange}}} denote best and second-best results.}
\begin{tabular}{ccc|ccccc}
\hline
AEP & DS & MSEC & F1 & mIoU & Params$\downarrow$ & FLOPs$\downarrow$ & Size$\downarrow$ \\ \hline
\usym{2713} & \usym{2717} & \usym{2717} & 0.7297 & 0.7954 & \cellcolor{orange!12}\underline{3.79M} & \cellcolor{orange!12}\underline{42.61G} & \cellcolor{cyan!10}\textbf{57MB} \\
\usym{2717} & \usym{2713} & \usym{2717} & 0.7336 & 0.7987 & \cellcolor{orange!12}\underline{3.79M} & 42.76G & \cellcolor{cyan!10}\textbf{57MB} \\
\usym{2717} & \usym{2717} & \usym{2713} & 0.7266 & 0.7963 & \cellcolor{cyan!10}\textbf{3.78M} & \cellcolor{cyan!10}\textbf{42.55G} & \cellcolor{cyan!10}\textbf{57MB} \\
\usym{2713} & \usym{2713} & \usym{2717} & \cellcolor{orange!12}\underline{0.7407} & \cellcolor{orange!12}\underline{0.8021} & 3.80M & 42.82G & \cellcolor{orange!12}\underline{58MB} \\
\usym{2713} & \usym{2717} & \usym{2713} & 0.7335 & 0.7977 & \cellcolor{orange!12}\underline{3.79M} & \cellcolor{orange!12}\underline{42.61G} & \cellcolor{orange!12}\underline{58MB} \\
\usym{2717} & \usym{2713} & \usym{2713} & 0.7395 & 0.7990 & \cellcolor{orange!12}\underline{3.79M} & 42.76G & \cellcolor{orange!12}\underline{58MB} \\
\usym{2713} & \usym{2713} & \usym{2713} & \cellcolor{cyan!10}\textbf{0.7469} & \cellcolor{cyan!10}\textbf{0.8031} & 3.80M & 42.82G & \cellcolor{orange!12}\underline{58MB} \\
\hline
\end{tabular}
\label{tab:abl_etpf_component}
\end{table}

\begin{figure}[!t]
  \centering
  \includegraphics[width=0.47\textwidth]{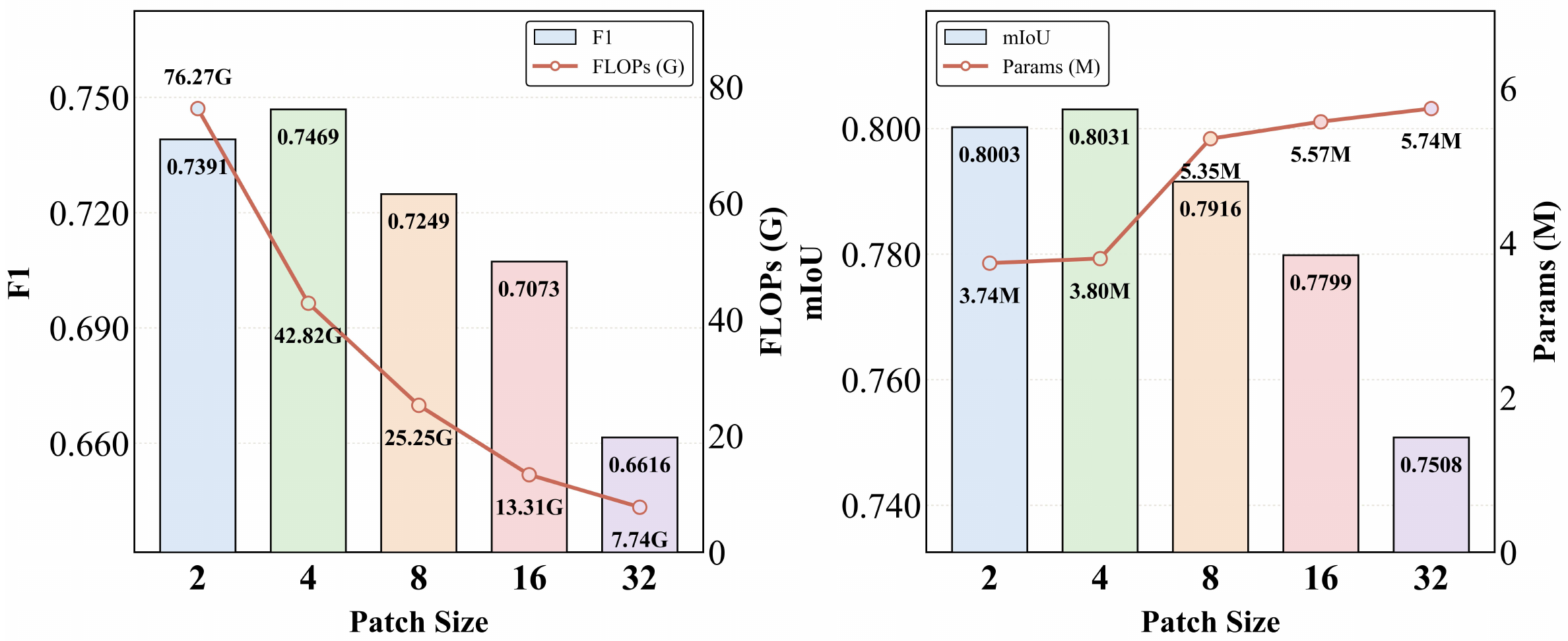}
  \caption{Comparison of performance when the patch size increases.}
  \label{fig:supp_abl_patch}
\end{figure}

\begin{table}[!t]
    \centering
    \scriptsize
    \setlength{\tabcolsep}{2pt}
    \renewcommand{\arraystretch}{0.9}
    \caption{Sensitivity study on different patch sizes. {\setlength{\fboxsep}{1pt}\colorbox{cyan!10}{\textbf{Blue}}} and {\setlength{\fboxsep}{1pt}\colorbox{orange!12}{\underline{Orange}}} denote best and second-best results.}
    \resizebox{0.45\textwidth}{!}{
    \begin{tabular}{c|ccccc}
    \toprule
        Patch Size & F1 & mIoU & Params$\downarrow$ & FLOPs$\downarrow$ & Size$\downarrow$ \\ 
    \midrule
        2 & \cellcolor{orange!12}\underline{0.7442} & \cellcolor{orange!12}\underline{0.7991} & \cellcolor{cyan!10}\textbf{3.79M} & 83.34G & \cellcolor{orange!12}\underline{64MB} \\
        4 & \cellcolor{cyan!10}\textbf{0.7443} & \cellcolor{cyan!10}\textbf{0.8049} & \cellcolor{orange!12}\underline{3.86M} & 44.77G & \cellcolor{cyan!10}\textbf{56MB} \\
        8 & 0.7185 & 0.7899 & 5.41M & 25.98G & 69MB \\
        16 & 0.6847 & 0.7638 & 5.63M & \cellcolor{orange!12}\underline{13.72G} & 69MB \\
        32 & 0.6509 & 0.7436 & 5.60M & \cellcolor{cyan!10}\textbf{8.07G} & 70MB \\
    \bottomrule
    \end{tabular}
    }
    \label{tab:ablation_patchsize}
\end{table}

\begin{figure}[!t]
  \centering
  \includegraphics[width=0.47\textwidth]{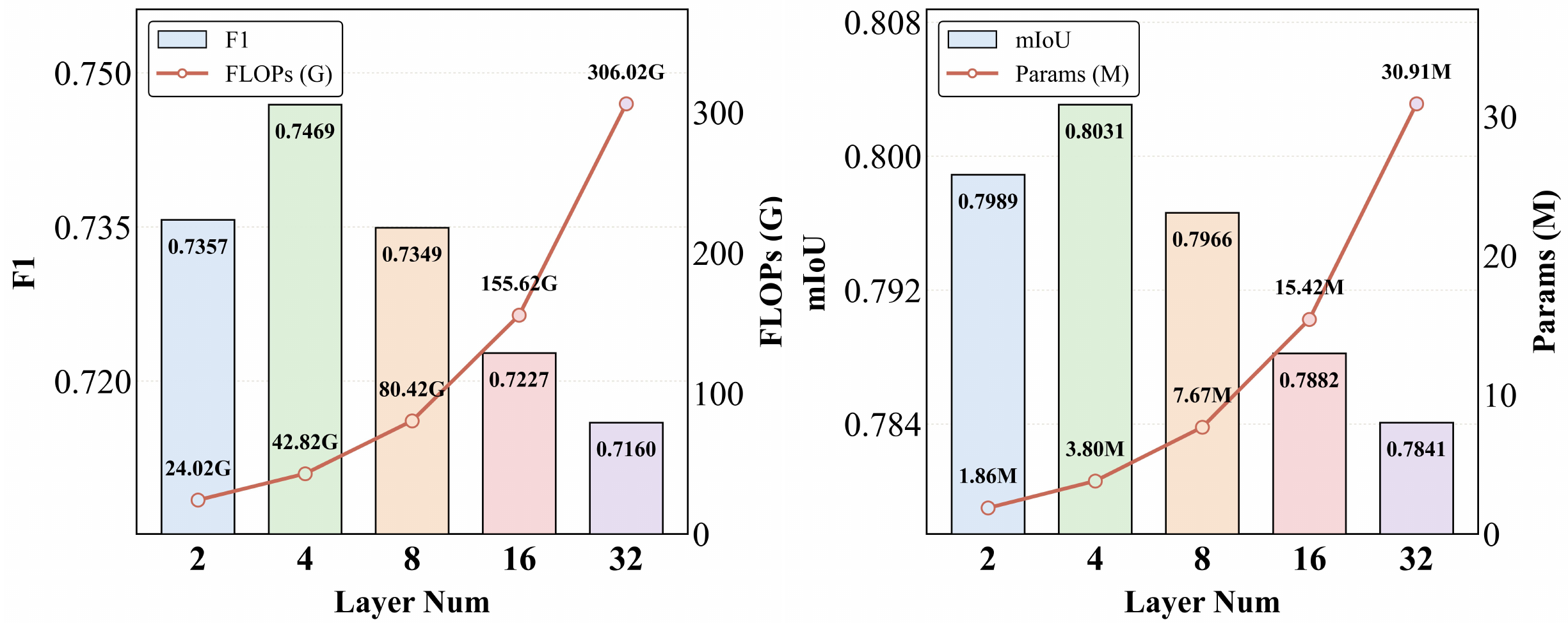}
  \caption{Comparison of performance when the number of Needld blocks increases.}
  \label{fig:supp_abl_layer}
\end{figure}

\begin{table}[!t]
    \centering
    \tiny
    \setlength{\tabcolsep}{2pt}
    \renewcommand{\arraystretch}{0.9}
    \caption{Sensitivity study on different layer numbers. {\setlength{\fboxsep}{1pt}\colorbox{cyan!10}{\textbf{Blue}}} and {\setlength{\fboxsep}{1pt}\colorbox{orange!12}{\underline{Orange}}} denote best and second-best results.}
    \resizebox{0.47\textwidth}{!}{
    \begin{tabular}{c|ccccc}
    \toprule
        Layer Num & F1 & mIoU & Params$\downarrow$ & FLOPs$\downarrow$ & Size$\downarrow$ \\ 
    \midrule
        2 & 0.7381 & 0.7980 & \cellcolor{cyan!10}\textbf{1.27M} & \cellcolor{cyan!10}\textbf{17.69G} & \cellcolor{cyan!10}\textbf{27MB} \\
        4 & \cellcolor{cyan!10}\textbf{0.7443} & \cellcolor{cyan!10}\textbf{0.8049} & \cellcolor{orange!12}\underline{3.86M} & \cellcolor{orange!12}\underline{44.77G} & \cellcolor{orange!12}\underline{56MB} \\
        8 & \cellcolor{orange!12}\underline{0.7431} & \cellcolor{orange!12}\underline{0.8028} & 7.77M & 82.18G & 112MB \\
        16 & 0.7350 & 0.7947 & 15.59M & 159.31G & 224MB \\
        32 & 0.7179 & 0.7881 & 31.23M & 313.58G & 449MB \\
    \bottomrule
    \end{tabular}
    }
    \label{tab:ablation_layers}
    \vspace{-0.3cm}
\end{table}

\vspace{-0.3cm}
\subsection{Ablation Study within ETPF}

To more thoroughly validate the impact of different components within ETPF on performance, we conduct ablation experiments on its three core constituents, with results reported in Table \ref{tab:abl_etpf_component}. AEP denotes anisotropic evidence propagation, which diffuses confidence along crack orientations via directional convolutions. DS denotes Dempster-Shafer evidential combination, which fuses multimodal evidence through DS rules and learnable mixture gating. MSEC denotes multi-scale evidence consistency, which uses coarse-scale confidence as a prior to constrain fine-scale predictions.

\begin{figure*}[!t]
  \centering
  \includegraphics[width=0.8\textwidth]{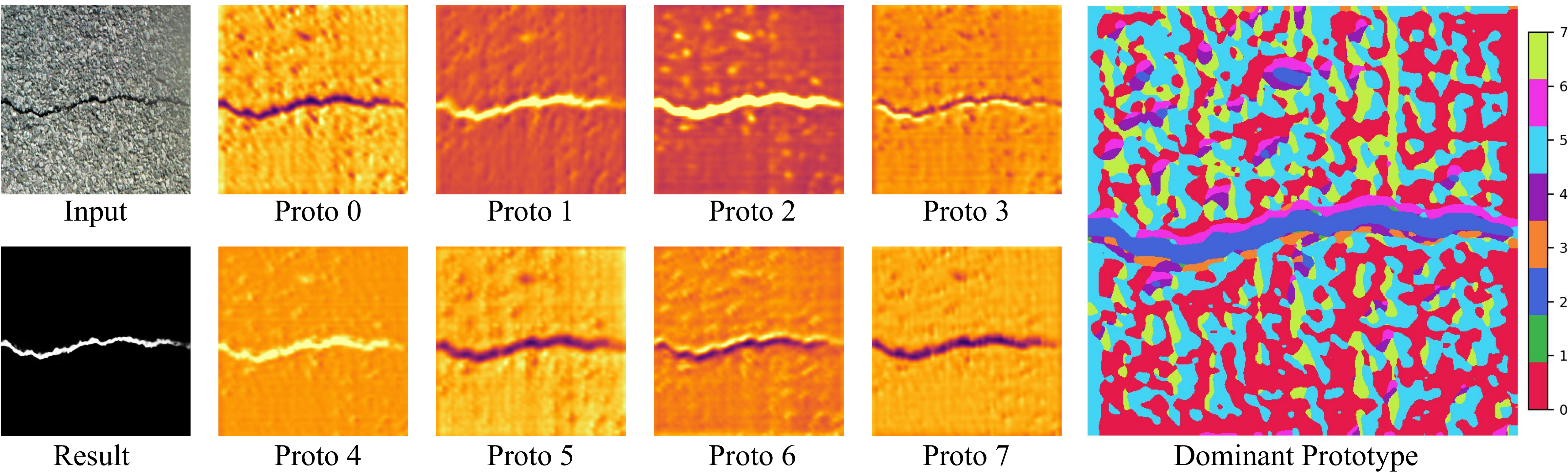}
    \caption{Visual analysis of different prototypes in FAPT.}
    \label{fig:fapt_proto_vis}
    \vspace{-0.4cm}
\end{figure*}

\begin{figure}[!t]
  \centering
  \includegraphics[width=0.47\textwidth]{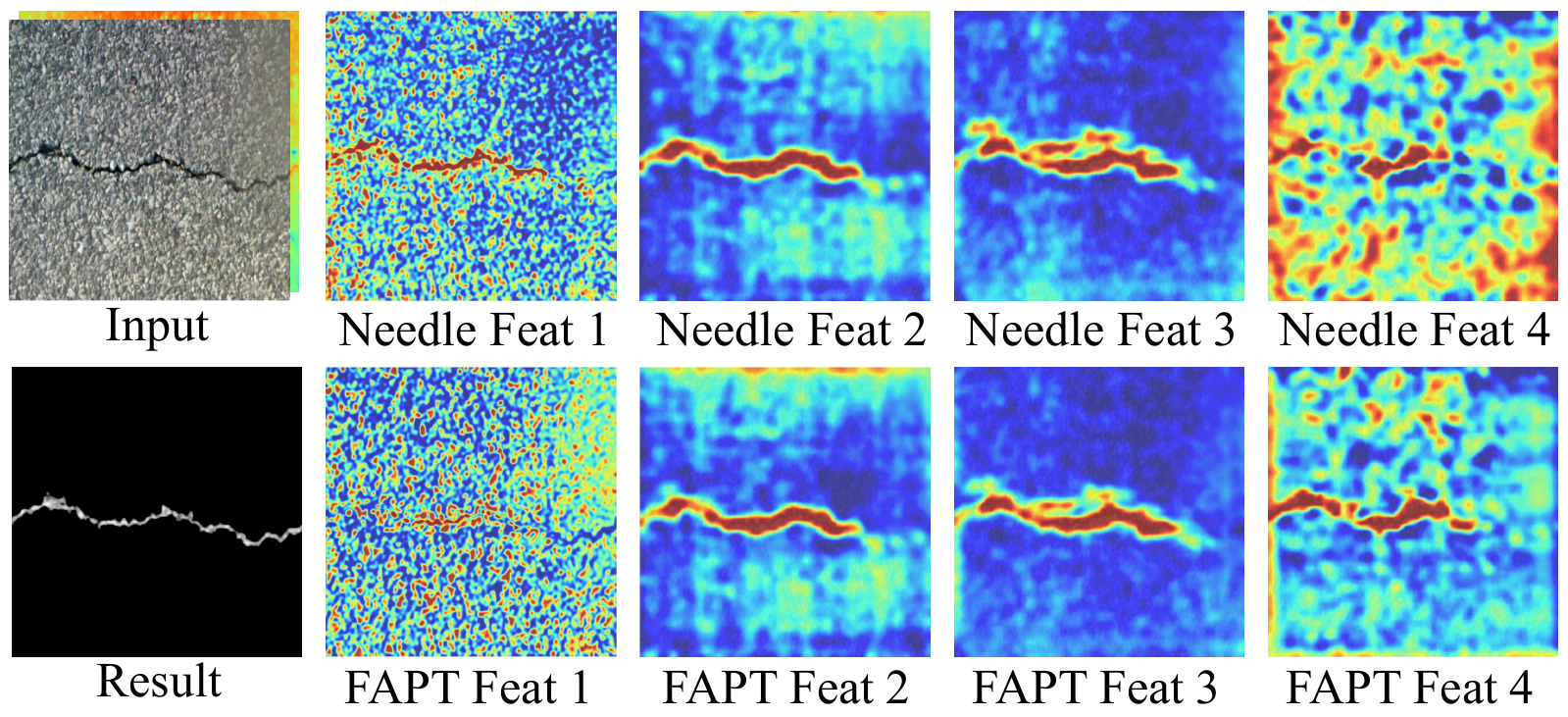}
    \caption{Visualization of the features of Needle and FAPT. L1-L4 represent the 1st-4th layers.}
    \label{fig:fapt_vis}
    \vspace{-0.3cm}
\end{figure}

When all three components are enabled, ETPF achieves F1 and mIoU scores of 0.7469 and 0.8031, respectively. Removing DS leads to the most substantial drop, with F1 and mIoU decreasing by 1.83\% and 0.67\%, respectively, demonstrating that Dempster-Shafer evidential combination is essential for accurately assessing the reliability of each modality under missing-modality conditions and for adjusting fusion weights accordingly. Without an explicit evidential combination mechanism, the fusion process cannot effectively distinguish reliable signals from degradation noise. Removing AEP results in decreases of 1.00\% in F1 and 0.51\% in mIoU, indicating that anisotropic evidence propagation effectively maintains the topological connectivity of cracks by weighting confidence propagation according to directional response strength, and that its absence causes discontinuities in the fused confidence along crack extension directions. Removing MSEC leads to a decrease of 0.84\% in F1, showing that the coarse-to-fine evidence regularization provides auxiliary enhancement for cross-scale confidence consistency.

When only a single component is used, performance degrades more noticeably. The variant using only MSEC achieves the lowest F1 of 0.7266, a decrease of 2.79\% compared to the full model, indicating that multi-scale consistency constraints can only take effect on the basis of reliable evidential combination and directional propagation. Overall, the three components of ETPF jointly constitute a complete evidential reasoning chain spanning the levels of evidence fusion, directional propagation, and scale consistency, and the addition of all components requires only 0.02M extra parameters and 0.27G FLOPs, incurring negligible computational overhead.

\begin{table}[!t]
\setlength{\tabcolsep}{2.3pt}
\caption{Performance comparison of different components. {\setlength{\fboxsep}{1pt}\colorbox{cyan!10}{\textbf{Blue}}} and {\setlength{\fboxsep}{1pt}\colorbox{orange!12}{\underline{Orange}}} denote best and second-best results.}
\begin{tabular}{ccc|ccccc}
\hline
FAPT & AGP & CohGate & F1 & mIoU & Params$\downarrow$ & FLOPs$\downarrow$ & Size$\downarrow$ \\ \hline
\usym{2713} & \usym{2717} & \usym{2717} & 0.7316 & 0.8009 & \cellcolor{cyan!10}\textbf{3.65M} & \cellcolor{cyan!10}\textbf{41.57G} & 56MB \\
\usym{2717} & \usym{2713} & \usym{2717} & 0.7290 & 0.7977 & 3.74M & 42.42G & \cellcolor{cyan!10}\textbf{53MB} \\
\usym{2717} & \usym{2717} & \usym{2713} & 0.7296 & 0.7989 & \cellcolor{orange!12}\underline{3.71M} & \cellcolor{orange!12}\underline{41.97G} & \cellcolor{cyan!10}\textbf{53MB} \\
\usym{2713} & \usym{2713} & \usym{2717} & \cellcolor{orange!12}\underline{0.7391} & 0.8011 & 3.75M & 42.42G & 57MB \\
\usym{2713} & \usym{2717} & \usym{2713} & 0.7350 & 0.7987 & 3.72M & \cellcolor{orange!12}\underline{41.97G} & 56MB \\
\usym{2717} & \usym{2713} & \usym{2713} & 0.7352 & \cellcolor{orange!12}\underline{0.8013} & 3.79M & 42.81G & \cellcolor{orange!12}\underline{54MB} \\
\usym{2713} & \usym{2713} & \usym{2713} & \cellcolor{cyan!10}\textbf{0.7469} & \cellcolor{cyan!10}\textbf{0.8031} & 3.80M & 42.82G & 58MB \\
\hline
\end{tabular}
\label{tab:abl_component}
\vspace{-0.5cm}
\end{table}

\begin{table}[!t]
    \centering
    \tiny
    \setlength{\tabcolsep}{8pt}
    \renewcommand{\arraystretch}{0.5}
    \caption{Sensitivity study on different prototype numbers in FAPT. {\setlength{\fboxsep}{1pt}\colorbox{cyan!10}{\textbf{Blue}}} and {\setlength{\fboxsep}{1pt}\colorbox{orange!12}{\underline{Orange}}} denote best and second-best results.}
    \resizebox{0.38\textwidth}{!}{
    \begin{tabular}{c|cc}
    \toprule
        Proto Num & F1 & mIoU \\
    \midrule
        2  & \cellcolor{orange!12}\underline{0.7431} & 0.8003 \\
        4  & 0.7383 & 0.8016 \\
        8  & \cellcolor{cyan!10}\textbf{0.7469} & \cellcolor{cyan!10}\textbf{0.8031} \\
        16 & 0.7423 & \cellcolor{orange!12}\underline{0.8026} \\
        32 & 0.7382 & 0.8005 \\
        64 & 0.7394 & 0.8015 \\
    \bottomrule
    \end{tabular}
    }
    \label{tab:ablation_proto_num}
    \vspace{-0.6cm}
\end{table}

\subsection{Sensitivity Analysis on Patch Size}

To validate the effect of patch size in Needle on model performance and computational efficiency, we conduct sensitivity experiments on the CrackPolar dataset, with results shown in Table \ref{tab:ablation_patchsize}. The choice of patch size exhibits a pronounced nonlinear influence on model performance. When the patch size is set to 4, Compass achieves the best balance between performance and efficiency, attaining the highest F1 and mIoU scores of 0.7469 and 0.8031 while maintaining a reasonable computational cost of 42.82G FLOPs. When the patch size is reduced to 2, the model can capture finer local textures, yet F1 decreases by 1.05\% and the computational cost surges to 76.27G, an increase of 78.12\% compared to the patch size of 4. This suggests that excessively fine-grained tokenization not only fails to bring performance gains but also leads to aggravated decay of long-range information in the WKV recurrent states due to the sharp increase in sequence length, thereby weakening the effective range of the orientation-aware modulation of GCM.

As shown in Figure \ref{fig:supp_abl_patch}, as the patch size increases, computational costs drop substantially but segmentation performance deteriorates sharply. With a patch size of 8, F1 decreases by 3.04\% and mIoU decreases by 1.45\% relative to the optimal setting. When the patch size is further increased to 32, F1 drops by as much as 12.90\% and mIoU drops by 6.96\%. This indicates that for crack-like targets that are slender and discontinuous, an excessively large patch size causes each token to cover too large a spatial region, causing the per-token orientation predictions of GCM to lose perceptual precision over subtle changes in crack direction, while the neighborhood consistency evaluation of CohGate also fails to effectively distinguish cracks from background due to insufficient spatial resolution. Therefore, a patch size of 4 is selected as the final configuration, ensuring that both GCM and CohGate have sufficient spatial resolution while maintaining low computational overhead.

\begin{figure*}[!t]
  \centering
  \includegraphics[width=\textwidth]{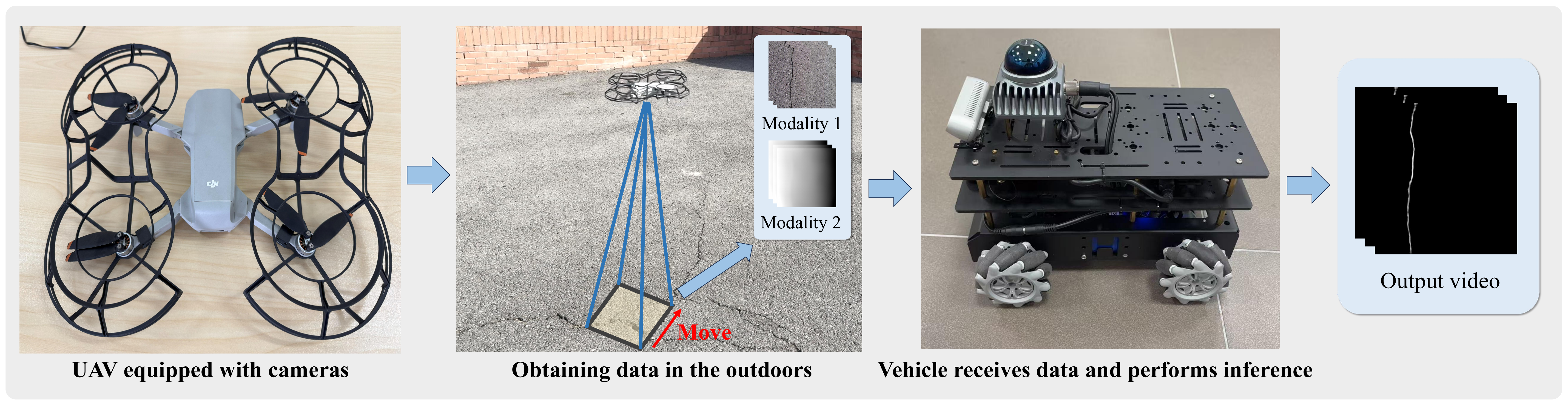}
  \caption{The deployment process in practical application. The UAV is equipped with high-definition cameras, responsible for collecting video stream data; the edge computing module is installed on the intelligent vehicle.}
\label{fig:real_world_deployment_process}
\end{figure*}

\vspace{-0.3cm}
\subsection{Sensitivity Analysis on the Number of Needle Blocks}

We use 4 Needle Block layers in Compass. To validate the rationality of this configuration, we conduct sensitivity experiments with varying numbers of layers on the CrackPolar \cite{Liu2025LIDAR} dataset, with results reported in Table \ref{tab:ablation_layers}. When the number of Needle Block layers is set to 4, Compass achieves the best F1 and mIoU scores of 0.7469 and 0.8031, respectively.

When the number of layers is reduced to 2, F1 and mIoU decrease by 1.53\% and 0.52\%, respectively, while the parameter count and FLOPs drop to 1.86M and 24.02G. This suggests that too few layers limit the ability of GCM to progressively accumulate orientation-aware information across multiple scales, and the anisotropic context probing of AGP also fails to adequately cover cracks of different widths due to insufficient hierarchical feature extraction.

As shown in Figure \ref{fig:supp_abl_layer}, as the number of layers increases, the parameter count, FLOPs, and model size rise substantially, while F1 and mIoU first improve and then decline. With 8 layers, F1 decreases by 1.63\% and mIoU decreases by 0.81\% compared to the optimal setting, while FLOPs increase to 80.42G, nearly twice that of the optimal configuration. When the number of layers is further increased to 32, F1 and mIoU decrease by 4.32\% and 2.42\%, respectively, with the parameter count expanding to 30.91M, FLOPs reaching 306.02G, and model size growing to 442MB. This indicates that an excessive number of layers introduces significant redundancy in the morphological and textural information across modalities, causing critical features to be smoothed and blurred during deep-layer propagation and weakening the model's capacity to represent the spatial hierarchy and morphological details of cracks. Therefore, 4 Needle Block layers represent the optimal balance between performance and efficiency.

\begin{figure*}[!t]
  \centering
  \includegraphics[width=0.85\textwidth]{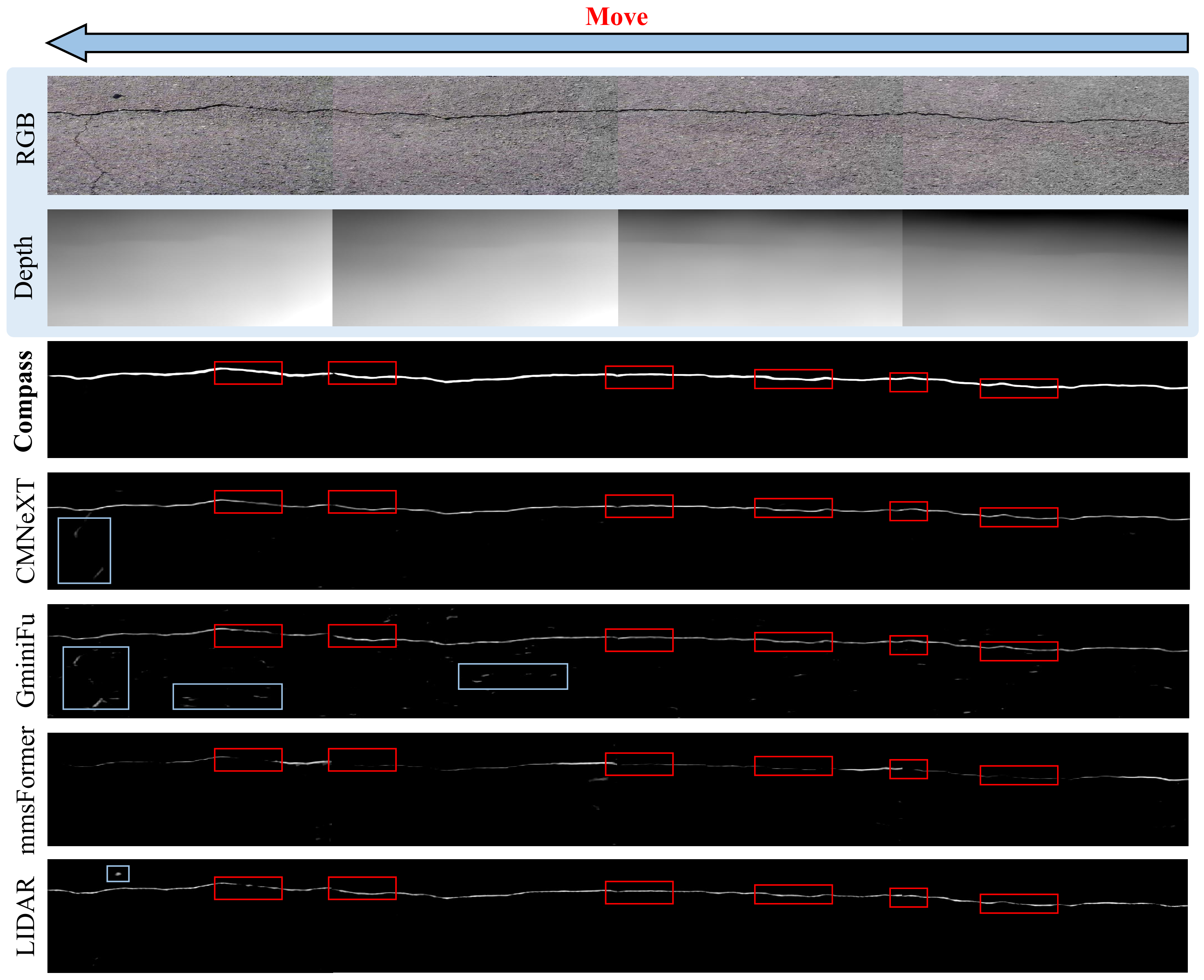}
  \caption{Performance comparison on real videos. Red boxes denote key regions; blue denote misidentifications.}
  \label{fig:application_results}
\end{figure*}

\subsection{Visualization Analysis of FAPT Feature Completion and Prototype Space}

To gain a deeper understanding of the working mechanism of FAPT, we analyze it from three perspectives, namely feature completion effectiveness, prototype space structure, and prototype number sensitivity.

As shown in Figure \ref{fig:fapt_vis}, we visualize the raw features output by the four Needle Block layers alongside the features after FAPT completion. In the raw Needle features, the shallow layers contain considerable noise and crack structures are only faintly discernible, while the crack skeleton gradually becomes clearer as the layer depth increases. At the same layer level, the FAPT-completed features exhibit higher contrast between cracks and background, more coherent activation along crack orientations, and background regions that approach uniformly low values. This demonstrates that the modality-agnostic semantic prototypes of FAPT effectively aggregate dispersed local responses into coherent crack semantic representations at each layer level, providing an intuitive explanation for the 3.09\% F1 improvement of the complete DSD over Single Stream.

To further reveal the underlying mechanism through which FAPT achieves the completion effect described above, we visualize the spatial activation maps of the $K=8$ learnable semantic prototypes as well as the dominant prototype assignment map at each pixel location, as shown in Figure \ref{fig:fapt_proto_vis}. The 8 prototypes spontaneously differentiate into three semantic roles without any explicit supervision. Proto 2 and Proto 5 exhibit strong positive responses in crack body regions, encoding the semantic characteristics of crack cores. Proto 1 and Proto 8 show low activation in crack regions but high activation over background areas, encoding the semantic concepts of background regions. Proto 6 and Proto 7 exhibit differentiated activation near crack boundaries, capturing the structural characteristics of edge transition zones. This spontaneous role differentiation indicates that end-to-end training causes the prototype space to form a compact representation of the semantic structure of crack scenes. This is also the core advantage of FAPT over conventional feature interpolation methods, as interpolation merely fills gaps at the numerical level whereas FAPT performs semantically meaningful reconstruction at the structural level through semantic prototypes.

The dominant prototype assignment map further corroborates these findings. The crack core, crack edge, and background regions are dominated by distinct prototypes, forming clean spatial partitions. This demonstrates that the soft assignment weight matrix $W$ in FAPT has learned effective spatial semantic divisions, with each prototype responsible for reconstructing a specific type of semantic region rather than performing uniform mixing across all regions. Since the 8 prototypes are defined in a feature space shared across all modalities and are not bound to the statistical properties of any specific modality, when a certain modality is missing, the residual modalities can still activate the corresponding prototypes through the same assignment mechanism to complete the reconstruction. This provides a visualization-level explanation for the performance advantage of FAPT over distribution alignment based methods.

\begin{figure*}[!t]
  \centering
  \includegraphics[width=\textwidth]{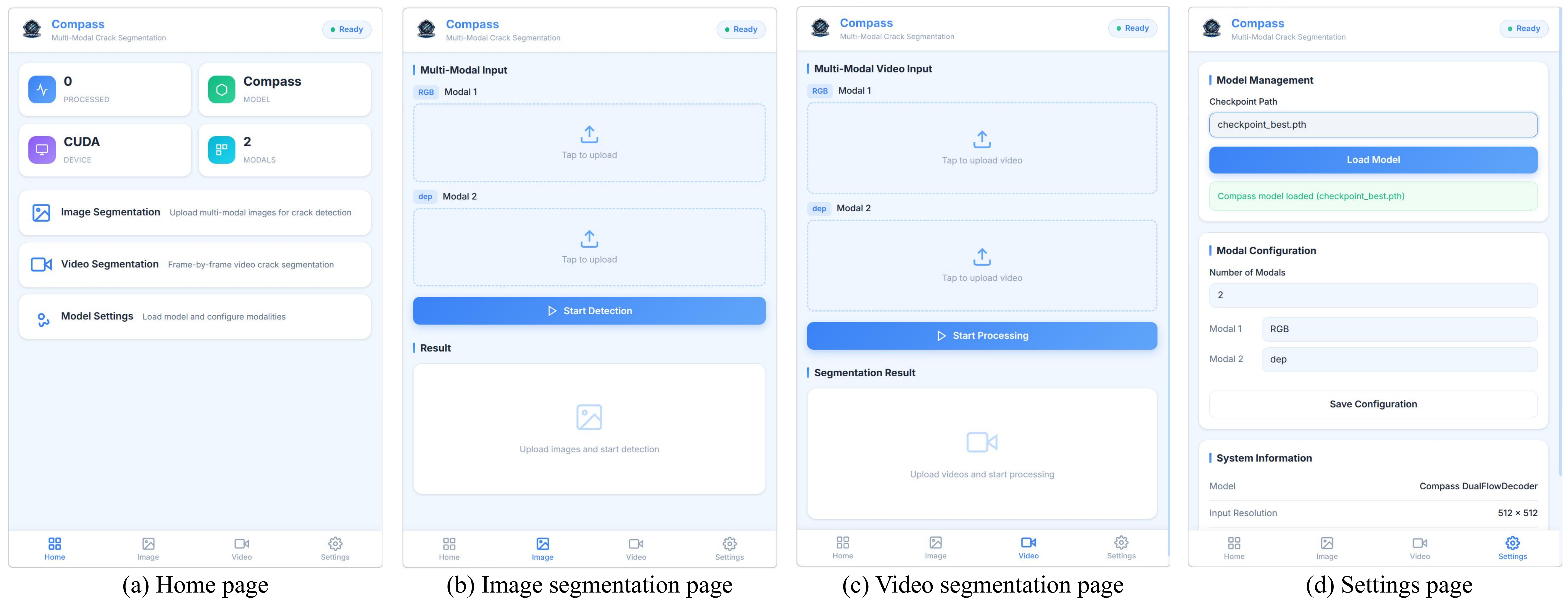}
  \caption{The interface of the Compass UI system comprises a home page,  an image segmentation page, a video segmentation page and a settings page.}
  \label{fig:supp_system_UI_page}
\end{figure*}

To validate the rationality of the prototype number $K$, we conduct sensitivity experiments with varying $K$ values on the CrackPolar dataset, with results shown in Table \ref{tab:ablation_proto_num}. When $K=8$, F1 and mIoU reach their optimal values of 0.7469 and 0.8031, respectively. When $K=2$, F1 and mIoU are 0.7431 and 0.8003, decreasing by 0.38\% and 0.28\% compared to the best setting, indicating that too few prototypes cannot sufficiently cover the three semantic roles of crack core, crack edge, and background, limiting the granularity of the completion. When $K$ increases to 16, 32, and 64, F1 drops to 0.7423, 0.7382, and 0.7394, respectively, showing a declining trend. This is because an excessive number of prototypes leads to over-segmentation of semantic roles, and the insufficient training samples per prototype prevent each prototype from learning a stable semantic representation, which in turn weakens the robustness of the completion. Therefore, $K=8$ achieves the optimal balance between prototype expressiveness and training adequacy.

\vspace{-0.3cm}
\section{Real-World Deployment Experiments}

To validate the deployment capability and inference efficiency of Compass in practical industrial scenarios, as shown in Figrue \ref{fig:real_world_deployment_process}, we construct a real-world deployment system based on the collaboration between an unmanned aerial vehicle and a ground station, and conduct field comparative tests against existing lightweight segmentation methods.

The hardware deployment architecture comprises two core modules, namely an aerial acquisition module and an edge computing module. The aerial acquisition module employs a DJI Mini drone equipped with high-definition visual sensors and depth sensors, responsible for collecting multimodal video stream data in complex scenes. The edge computing module is mounted on an intelligent vehicle, with an NVIDIA Jetson Orin NX Super edge intelligence computing card as the core computational platform running Ubuntu 22.04. During the experiments, the drone and the intelligent vehicle establish a low-latency real-time communication link via a high-bandwidth local area network. The drone maintains a constant cruising speed of 0.2 m/s to perform continuous video acquisition of crack regions on walls and ground surfaces, transmitting the raw video stream in real time to the edge computing module. The Compass model deployed on the Jetson platform performs real-time frame-by-frame inference on the received unseen video stream data, instantly generating crack segmentation results.

To evaluate the operational performance of the model, we simultaneously deploy the pretrained weights of CMNeXT~\cite{Zhang2023Delivering}, GeminiFusion~\cite{Jia2024GeminiFusion}, mmsFormer~\cite{Reza2024MMSFormer}, and LIDAR~\cite{Liu2025LIDAR} on the same edge platform for real-time inference comparison. In the simulated test, the RGB modality is kept complete while the missing rate of the depth modality reaches 90\%.

As shown in Figure \ref{fig:application_results}, Compass demonstrates significant advantages when processing unfamiliar multimodal video stream data. The generated segmentation masks exhibit clear boundaries, accurately capturing subtle crack branches while effectively suppressing background noise in complex environments. This can be attributed to DSD having actively adapted the model during training to degradation conditions comparable to or more severe than those encountered at inference time. FAPT reconstructs semantically complete feature representations from only 10\% of the residual depth information through modality-agnostic semantic prototypes, enabling Compass to maintain stable segmentation quality under extreme missing conditions. The orientation-aware key modulation of GCM and the connectivity gating of CohGate within Needle ensure the topological continuity of cracks across video frames, and the anisotropic context probing of AGP adaptively captures the boundary information of cracks with varying widths. The uncertainty-gated decoder of ETPF automatically suppresses noise propagation in degraded depth modality regions, ensuring the reliability of fusion decisions.

Among the comparison methods, GeminiFusion~\cite{Jia2024GeminiFusion} fails to effectively suppress irrelevant background texture noise, leading to confusion between crack regions and background interference and causing a large number of false detections. mmsFormer~\cite{Reza2024MMSFormer} suffers from severe missed detections, omitting a substantial amount of critical crack texture details, indicating a lack of precise crack perception capability on unseen data. LIDAR~\cite{Liu2025LIDAR} fails to achieve accurate segmentation in critical fine crack junction regions, producing evident discontinuities and missed detections. CMNeXT~\cite{Zhang2023Delivering} preserves the basic crack orientation but exhibits false detections in localized regions.

To further improve the usability of the system, we design a visualization user interface for Compass, as shown in Figure \ref{fig:supp_system_UI_page}. The system contains four core functional modules. The main page serves as the control center, integrating function control buttons, selectable input modality number indicators, and model loading status monitoring, enabling users to quickly grasp the system configuration and operational status. The model settings page provides a flexible configuration interface through which users can load pretrained weight files and specify the number and names of input modalities. The image segmentation page supports receiving multimodal crack images from the drone or uploading them locally, automatically performing inference and displaying the original images alongside the segmentation masks. The video segmentation page is designed for real-time monitoring of dynamic scenes, receiving multimodal video streams transmitted from the drone or user-uploaded video data, feeding them frame by frame into the Compass model for real-time inference, and outputting segmentation results in video stream format. The development of this visualization system substantially improves the convenience and deployability of Compass in practical engineering applications.

\bibliographystyle{ACM-Reference-Format}
\bibliography{sample-base}

@String{Computer = "{IEEE} Computer" }

@String{Springer = "Springer-Verlag" }

@article{Cheng2024Selective,
  author = {X. Cheng and T. He and F. Shi and M. Zhao and X. Liu and S. Chen},
  title = {{Selective Feature Fusion and Irregular-Aware Network for Pavement Crack Detection}},
  journal = {IEEE Transactions on Intelligent Transportation Systems},
  volume = {25},
  number = {5},
  pages = {3445--3456},
  year = {2024},
}

@inproceedings{Chen2024Mind,
  author = {Z. Chen and Z. Lai and J. Chen and J. Li},
  title = {{Mind Marginal Non-Crack Regions: Clustering-Inspired Representation Learning for Crack Segmentation}},
  booktitle = {Proceedings of the IEEE/CVF Conference on Computer Vision and Pattern Recognition},
  pages = {12698--12708},
  year = {2024},
}

@inproceedings{Benz2024Omni,
  author = {C. Benz and V. Rodehorst},
  title = {{Omni-Crack30k: A Benchmark for Crack Segmentation and the Reasonable Effectiveness of Transfer Learning}},
  booktitle = {Proceedings of the IEEE/CVF Conference on Computer Vision and Pattern Recognition Workshops},
  pages = {3876--3886},
  year = {2024},
}

@inproceedings{Chu2024CrackGauGAN,
  author = {H. Chu and W. Chen and L. Deng},
  title = {{CrackGauGAN: Semantic Layout-Based Crack Image Synthesis for Automated Crack Segmentation}},
  booktitle = {Proceedings of the 41st International Symposium on Automation and Robotics in Construction (ISARC)},
  pages = {1--8},
  year = {2024},
}

@inproceedings{Liu2021CrackFormer,
  author = {H. Liu and X. Miao and C. Mertz and C. Xu and H. Kong},
  title = {{CrackFormer: Transformer Network for Fine-Grained Crack Detection}},
  booktitle = {Proceedings of the IEEE/CVF International Conference on Computer Vision},
  pages = {3763--3772},
  year = {2021},
}

@inproceedings{Liu2025SCSegamba,
  author = {H. Liu and C. Jia and F. Shi and X. Cheng and S. Chen},
  title = {{SCSegamba: Lightweight Structure-Aware Vision Mamba for Crack Segmentation in Structures}},
  booktitle = {Proceedings of the IEEE/CVF Conference on Computer Vision and Pattern Recognition},
  pages = {29406--29416},
  year = {2025}
}

@article{Yang2020Feature,
  author = {F. Yang and L. Zhang and S. Yu and D. Prokhorov and X. Mei and H. Ling},
  title = {{Feature Pyramid and Hierarchical Boosting Network for Pavement Crack Detection}},
  journal = {IEEE Transactions on Intelligent Transportation Systems},
  volume = {21},
  number = {4},
  pages = {1525--1535},
  year = {2020},
}

@inproceedings{Jaziri2024Designing,
  author = {A. Jaziri and M. Mundt and A. Fernandez and V. Ramesh},
  title = {{Designing a hybrid neural system to learn real-world crack segmentation from fractal-based simulation}},
  booktitle = {Proceedings of the IEEE/CVF Winter Conference on Applications of Computer Vision},
  pages = {8636--8646},
  year = {2024},
}

@inproceedings{Liu2025LIDAR,
  author = {H. Liu and C. Jia and F. Shi and X. Cheng and M. Shi and X. Xie and S. Chen},
  title = {{LIDAR: Lightweight Adaptive Cue-Aware Fusion Vision Mamba for Multimodal Segmentation of Structural Cracks}},
  booktitle = {Proceedings of the 33rd ACM International Conference on Multimedia},
  pages = {1832--1841},
  year = {2025},
}

@article{Zhang2023CMX,
  author = {J. Zhang and H. Liu and K. Yang and X. Hu and R. Liu and R. Stiefelhagen},
  title = {{CMX: Cross-Modal Fusion for RGB-X Semantic Segmentation with Transformers}},
  journal = {IEEE Transactions on Intelligent Transportation Systems},
  volume = {24},
  number = {12},
  pages = {14679--14694},
  year = {2023}
}

@inproceedings{Zhang2023Delivering,
  author = {J. Zhang and R. Liu and H. Shi and K. Yang and S. Reiß and K. Peng and H. Fu and K. Wang and R. Stiefelhagen},
  title = {{Delivering Arbitrary-Modal Semantic Segmentation}},
  booktitle = {Proceedings of the IEEE/CVF Conference on Computer Vision and Pattern Recognition},
  pages = {1136--1147},
  year = {2023}
}

@inproceedings{Liang2022Multimodal,
  author = {Y. Liang and R. Wakaki and S. Nobuhara and K. Nishino},
  title = {{Multimodal Material Segmentation}},
  booktitle = {Proceedings of the IEEE/CVF Conference on Computer Vision and Pattern Recognition},
  pages = {19800--19808},
  year = {2022}
}

@inproceedings{Wan2025Sigma,
  author = {Z. Wan and P. Zhang and Y. Wang and S. Yong and S. Stepputtis and K. Sycara and Y. Xie},
  title = {{Sigma: Siamese Mamba Network for Multi-Modal Semantic Segmentation}},
  booktitle = {2025 IEEE/CVF Winter Conference on Applications of Computer Vision},
  pages = {1734--1744},
  year = {2025}
}

@inproceedings{Lee2023Multimodal,
  author = {Y.-L. Lee and Y.-H. Tsai and W.-C. Chiu and C.-Y. Lee},
  title = {{Multimodal Prompting with Missing Modalities for Visual Recognition}},
  booktitle = {Proceedings of the IEEE/CVF Conference on Computer Vision and Pattern Recognition},
  pages = {14943--14952},
  year = {2023}
}

@inproceedings{Hong2022Cross,
  author = {Y. Hong and H. Dai and Y. Ding},
  title = {{Cross-Modality Knowledge Distillation Network for Monocular 3D Object Detection}},
  booktitle = {European Conference on Computer Vision, Springer},
  pages = {87--104},
  year = {2022}
}

@inproceedings{Maheshwari2024Missing,
  author = {H. Maheshwari and Y.-C. Liu and Z. Kira},
  title = {{Missing Modality Robustness in Semi-Supervised Multi-Modal Semantic Segmentation}},
  booktitle = {Proceedings of the IEEE/CVF Winter Conference on Applications of Computer Vision},
  pages = {1020--1030},
  year = {2024}
}

@inproceedings{Dai2025Unbiased,
  author = {R. Dai and C. Li and Y. Yan and L. Mo and K. Qin and T. He},
  title = {{Unbiased Missing-Modality Multimodal Learning}},
  booktitle = {Proceedings of the IEEE/CVF International Conference on Computer Vision},
  pages = {24507--24517},
  year = {2025}
}

@article{Xie2025CCSD,
  author = {D. Xie and Y. Wu and Z. Ai and J. Min and Z. Jiang and S. Geng and L. Wang},
  title = {{CCSD: Cross-Modal Compositional Self-Distillation for Robust Brain Tumor Segmentation with Missing Modalities}},
  journal = {arXiv preprint arXiv:2511.14599},
  year = {2025}
}

@inproceedings{Ke2025Knowledge,
  author = {G. Ke and S. He and X. Wang and B. Wang and G. Chao and Y. Zhang and Y. Xie and H. Su},
  title = {{Knowledge Bridger: Towards Training-Free Missing Modality Completion}},
  booktitle = {Proceedings of the IEEE/CVF Conference on Computer Vision and Pattern Recognition},
  pages = {25864--25873},
  year = {2025}
}

@inproceedings{Xu2026MCMoE,
  author = {Huangbiao Xu and Huanqi Wu and Xiao Ke and Junyi Wu and Rui Xu and Jinglin Xu},
  title = {{MCMoE: Completing Missing Modalities with Mixture of Experts for Incomplete Multimodal Action Quality Assessment}},
  booktitle = {Proceedings of the AAAI Conference on Artificial Intelligence},
  year = {2026}
}

@inproceedings{Lyu2026TouchFormer,
  author = {Kailin Lyu and Long Xiao and Jianing Zeng and Junhao Dong and Xuexin Liu and Zhuojun Zou and Haoyue Yang and Lin Shu and Jie Hao},
  title = {{TouchFormer: A Robust Transformer-Based Framework for Multimodal Material Perception}},
  booktitle = {Proceedings of the AAAI Conference on Artificial Intelligence},
  year = {2026}
}

@article{Neverova2015ModDrop,
  author = {N. Neverova and C. Wolf and G. Taylor and F. Nebout},
  title = {{ModDrop: Adaptive Multi-Modal Gesture Recognition}},
  journal = {IEEE Transactions on Pattern Analysis and Machine Intelligence},
  volume = {38},
  number = {8},
  pages = {1692--1706},
  year = {2015}
}

@inproceedings{Wang2023Multi,
  author = {H. Wang and Y. Chen and C. Ma and J. Avery and L. Hull and G. Carneiro},
  title = {{Multi-Modal Learning with Missing Modality via Shared-Specific Feature Modelling}},
  booktitle = {Proceedings of the IEEE/CVF Conference on Computer Vision and Pattern Recognition},
  pages = {15878--15887},
  year = {2023}
}

@inproceedings{Nezakati2025MMP,
  author = {Niki Nezakati and Md Kaykobad Reza and Ameya Patil and Mashhour Solh and M. Salman Asif},
  title = {{MMP: Towards Robust Multi-Modal Learning with Masked Modality Projection}},
  booktitle = {2025 IEEE International Conference on Big Data (BigData)},
  pages = {1480--1485},
  year = {2025},
}

@inproceedings{Nam2024Modality,
  author = {J.-H. Nam and N. S. Syazwany and S. J. Kim and S.-C. Lee},
  title = {{Modality-Agnostic Domain Generalizable Medical Image Segmentation by Multi-Frequency in Multi-Scale Attention}},
  booktitle = {Proceedings of the IEEE/CVF Conference on Computer Vision and Pattern Recognition},
  pages = {11480--11491},
  year = {2024}
}

@inproceedings{Seichter2022Efficient,
  author = {D. Seichter and S. B. Fischedick and M. Kohler and H.-M. Groß},
  title = {{Efficient Multi-Task RGB-D Scene Analysis for Indoor Environments}},
  booktitle = {2022 International Joint Conference on Neural Networks, IEEE},
  pages = {1--10},
  year = {2022}
}

@inproceedings{Liu2021Swin,
  author = {Z. Liu and Y. Lin and Y. Cao and H. Hu and Y. Wei and Z. Zhang and S. Lin and B. Guo},
  title = {{Swin Transformer: Hierarchical Vision Transformer Using Shifted Windows}},
  booktitle = {Proceedings of the IEEE/CVF International Conference on Computer Vision},
  pages = {10012--10022},
  year = {2021}
}

@article{Chen2025HSPFormer,
  author = {S. Chen and T. Han and C. Zhang and J. Su and R. Wang and Y. Chen and Z. Wang and G. Cai},
  title = {{HSPFormer: Hierarchical Spatial Perception Transformer for Semantic Segmentation}},
  journal = {IEEE Transactions on Intelligent Transportation Systems},
  year = {2025},
}

@inproceedings{Wang2024PSSD,
  author = {H. Wang and X. Liang and T. Zhang and Y. Gu and W. Geng},
  title = {{PSSD-Transformer: Powerful Sparse Spike-Driven Transformer for Image Semantic Segmentation}},
  booktitle = {Proceedings of the 32nd ACM International Conference on Multimedia},
  pages = {758--767},
  year = {2024}
}

@inproceedings{Gu2024Mamba,
  author = {A. Gu and T. Dao},
  title = {{Mamba: Linear-Time Sequence Modeling with Selective State Spaces}},
  booktitle = {First Conference on Language Modeling},
  year = {2024}
}

@inproceedings{Gu2022Efficiently,
  author = {A. Gu and K. Goel and C. Ré},
  title = {{Efficiently Modeling Long Sequences with Structured State Spaces}},
  booktitle = {The International Conference on Learning Representations},
  year = {2022}
}

@inproceedings{Hatamizadeh2025MambaVision,
  author = {A. Hatamizadeh and J. Kautz},
  title = {{MambaVision: A Hybrid Mamba-Transformer Vision Backbone}},
  booktitle = {Proceedings of the IEEE/CVF Conference on Computer Vision and Pattern Recognition},
  pages = {25261--25270},
  year = {2025},
}

@inproceedings{Guo2025MambaIRv2,
  author = {H. Guo and Y. Guo and Y. Zha and Y. Zhang and W. Li and T. Dai and S.-T. Xia and Y. Li},
  title = {{MambaIRv2: Attentive State Space Restoration}},
  booktitle = {Proceedings of the IEEE/CVF Conference on Computer Vision and Pattern Recognition},
  pages = {28124--28133},
  year = {2025}
}

@inproceedings{Yu2025MambaOut,
  author = {W. Yu and X. Wang},
  title = {{MambaOut: Do We Really Need Mamba for Vision?}},
  booktitle = {Proceedings of the IEEE/CVF Conference on Computer Vision and Pattern Recognition},
  pages = {4484--4496},
  year = {2025}
}

@inproceedings{Peng2023RWKV,
  author = {Bo Peng and Eric Alcaide and Quentin Gregory Anthony and Alon Albalak and Samuel Arcadinho and Stella Biderman and Huanqi Cao and Xin Cheng and Michael Nguyen Chung and Leon Derczynski and Xingjian Du and Matteo Grella and Kranthi Kiran GV and Xuzheng He and Haowen Hou and Przemyslaw Kazienko and Jan Kocon and Jiaming Kong and Bartlomiej Koptyra and Hayden Lau and Krishna Sri Ipsit Mantri and Ferdinand Mom and Atsushi Saito and Guangyu Song and Xiangru Tang and Bolun Wang and Johan S. Wind and Stanislaw Wozniak and Ruichong Zhang and Zhenyuan Zhang and Qihang Zhao and Peng Zhou and Qinghua Zhou and Jian Zhu and Rui-Jie Zhu},
  title = {{RWKV: Reinventing RNNs for the Transformer Era}},
  booktitle = {Findings of the Association for Computational Linguistics: EMNLP 2023},
  pages = {14048--14077},
  year = {2023},
  publisher = {Association for Computational Linguistics},
}

@inproceedings{Duan2025Vision,
  author = {Y. Duan and W. Wang and Z. Chen and X. Zhu and L. Lu and T. Lu and Y. Qiao and H. Li and J. Dai and W. Wang},
  title = {{Vision-RWKV: Efficient and Scalable Visual Perception with RWKV-Like Architectures}},
  booktitle = {The Thirteenth International Conference on Learning Representations},
  year = {2025}
}

@inproceedings{He2025PointRWKV,
  author = {Q. He and J. Zhang and J. Peng and H. He and X. Li and Y. Wang and C. Wang},
  title = {{PointRWKV: Efficient RWKV-Like Model for Hierarchical Point Cloud Learning}},
  booktitle = {Proceedings of the AAAI Conference on Artificial Intelligence, vol. 39, no. 3},
  pages = {3410--3418},
  year = {2025},
}

@inproceedings{Xu2025URWKV,
  author = {R. Xu and Y. Niu and Y. Li and H. Xu and W. Liu and Y. Chen},
  title = {{URWKV: Unified RWKV Model with Multi-State Perspective for Low-Light Image Restoration}},
  booktitle = {Proceedings of the IEEE/CVF Conference on Computer Vision and Pattern Recognition},
  pages = {21267--21276},
  year = {2025}
}

@article{Yang2026Restore,
  author = {Z. Yang and J. Li and H. Zhang and D. Zhao and B. Wei and Y. Xu},
  title = {{Restore-RWKV: Efficient and Effective Medical Image Restoration with RWKV}},
  journal = {IEEE Journal of Biomedical and Health Informatics},
  volume = {30},
  number = {1},
  pages = {513--526},
  year = {2026},
}

@article{Chen2025Zig,
  author = {T. Chen and X. Zhou and Z. Tan and Y. Wu and Z. Wang and Z. Ye and T. Gong and Q. Chu and N. Yu and L. Lu},
  title = {{Zig-RiR: Zigzag RWKV-in-RWKV for Efficient Medical Image Segmentation}},
  journal = {IEEE Transactions on Medical Imaging},
  year = {2025}
}

@article{Li2025Pan,
  author = {X. Li and T. Hu and K. Cao and J. Zhang and C. Xie and M. Zhou and D. Hong},
  title = {{Pan-Sharpening via Causal-Aware Feature Distribution Calibration}},
  journal = {IEEE Transactions on Geoscience and Remote Sensing},
  year = {2025},
}

@article{Wang2023Channel,
  author = {Y. Wang and W. Huang and F. Sun and F. He and D. Tao},
  title = {{Channel Exchanging Networks for Multimodal and Multitask Dense Image Prediction}},
  journal = {IEEE Transactions on Pattern Analysis and Machine Intelligence},
  volume = {45},
  number = {5},
  pages = {5481--5496},
  year = {2023}
}

@inproceedings{Yin2025DFormerv2,
  author = {Bo-Wen Yin and Jiao-Long Cao and Ming-Ming Cheng and Qibin Hou},
  title = {{DFormerv2: Geometry Self-Attention for RGBD Semantic Segmentation}},
  booktitle = {Proceedings of the IEEE/CVF Conference on Computer Vision and Pattern Recognition},
  pages = {19345--19355},
  year = {2025}
}

@article{Chen2024Frequency,
  author = {Linwei Chen and Ying Fu and Lin Gu and Chenggang Yan and Tatsuya Harada and Gao Huang},
  title = {{Frequency-Aware Feature Fusion for Dense Image Prediction}},
  journal = {IEEE Transactions on Pattern Analysis and Machine Intelligence},
  volume = {46},
  number = {12},
  pages = {10763--10780},
  year = {2024},
}

@inproceedings{Li2025StitchFusion,
  author = {B. Li and D. Zhang and Z. Zhao and J. Gao and X. Li},
  title = {{StitchFusion: Weaving Any Visual Modalities to Enhance Multimodal Semantic Segmentation}},
  booktitle = {Proceedings of the 33rd ACM International Conference on Multimedia},
  year = {2025},
}

@article{Huang2025Deep,
  author = {L. Huang and S. Ruan and P. Decazes and T. Den{\oe}ux},
  title = {{Deep Evidential Fusion with Uncertainty Quantification and Reliability Learning for Multimodal Medical Image Segmentation}},
  journal = {Information Fusion},
  volume = {113},
  pages = {102648},
  year = {2025},
}

@article{Wang2024Dual,
  author = {J. Wang and Z. Zeng and P. K. Sharma and O. Alfarraj and A. Tolba and J. Zhang and L. Wang},
  title = {{Dual-path network combining CNN and transformer for pavement crack segmentation}},
  journal = {Automation in Construction},
  volume = {158},
  pages = {105217},
  year = {2024},
}

@inproceedings{Wang2022Multimodal,
  author = {Y. Wang and X. Chen and L. Cao and W. Huang and F. Sun and Y. Wang},
  title = {{Multimodal Token Fusion for Vision Transformers}},
  booktitle = {Proceedings of the IEEE/CVF Conference on Computer Vision and Pattern Recognition},
  pages = {12186--12195},
  year = {2022}
}

@inproceedings{Jia2024GeminiFusion,
  author = {D. Jia and J. Guo and K. Han and H. Wu and C. Zhang and C. Xu and X. Chen},
  title = {{GeminiFusion: Efficient Pixel-Wise Multimodal Fusion for Vision Transformer}},
  booktitle = {Proceedings of the 41st International Conference on Machine Learning},
  year = {2024}
}

@inproceedings{Dai2024Study,
  author = {Y. Dai and H. Chen and J. Du and R. Wang and S. Chen and J. Ma and H. Wang and C.-H. Lee},
  title = {{A Study of Dropout-Induced Modality Bias on Robustness to Missing Video Frames for Audio-Visual Speech Recognition}},
  booktitle = {Proceedings of the IEEE/CVF Conference on Computer Vision and Pattern Recognition},
  year = {2024}
}

@inproceedings{Sikdar2024SKD,
  author = {A. Sikdar and J. Teotia and S. Sundaram},
  title = {{SKD-Net: Spectral-Based Knowledge Distillation in Low-Light Thermal Imagery for Robotic Perception}},
  booktitle = {2024 IEEE International Conference on Robotics and Automation, IEEE},
  pages = {9041--9047},
  year = {2024}
}

@inproceedings{Zhang2025L3TC,
  author = {J. Zhang and Z. Cheng and Y. Zhao and S. Wang and D. Zhou and G. Lu and L. Song},
  title = {{L3TC: Leveraging RWKV for Learned Lossless Low-Complexity Text Compression}},
  booktitle = {Proceedings of the AAAI Conference on Artificial Intelligence, vol. 39, no. 12},
  pages = {13251--13259},
  year = {2025}
}

@inproceedings{Sun2025RWKV3D,
  author = {C. Sun and S. Pang and Y. Wang and L. Qi},
  title = {{RWKV3D: An RWKV-Based Model with Multiple Training Strategies for Point Cloud Analysis}},
  booktitle = {Proceedings of the 33rd ACM International Conference on Multimedia},
  pages = {650--659},
  year = {2025}
}

@inproceedings{He2025RWKV,
  author = {W. He and X. Chen and W. Chen and H. Wang and Y. Liu and R. Li},
  title = {{RWKV-PCSsc: Exploring RWKV Model for Point Cloud Semantic Scene Completion}},
  booktitle = {Proceedings of the 33rd ACM International Conference on Multimedia},
  pages = {161--170},
  year = {2025}
}

@inproceedings{Xie2025Learning,
  author = {D. Xie and X. Hu and Z. Wei and Z. Yang and Y. Jiang and Y. Zhou},
  title = {{Learning Structural Priors via Laplacian RWKV Diffusion with Light-Effect Dataset for Nighttime Visibility Enhancement}},
  booktitle = {Proceedings of the 33rd ACM International Conference on Multimedia},
  pages = {4590--4599},
  year = {2025}
}

@inproceedings{Zhou2025WKV,
  author = {M. Zhou and X. He and D. Hong and B. Huang},
  title = {{WKV-Sharing Embraced Random Shuffle RWKV High-Order Modeling for Pan-Sharpening}},
  booktitle = {Advances in Neural Information Processing Systems},
  year = {2025}
}

@inproceedings{Hao2024PrimKD,
  author = {Z. Hao and Z. Xiao and Y. Luo and J. Guo and J. Wang and L. Shen and H. Hu},
  title = {{PrimKD: Primary Modality Guided Multimodal Fusion for RGB-D Semantic Segmentation}},
  booktitle = {Proceedings of the 32nd ACM International Conference on Multimedia},
  pages = {1943--1951},
  year = {2024},
}

@article{Liu2025Part,
  author = {Y. Liu and C. Li and S. Xu and J. Han},
  title = {{Part-Whole Relational Fusion Towards Multi-Modal Scene Understanding}},
  journal = {International Journal of Computer Vision},
  volume = {133},
  number = {7},
  pages = {4483--4503},
  year = {2025}
}

@article{Reza2024MMSFormer,
  author = {Md Kaykobad Reza and Ashley Prater-Bennette and M. Salman Asif},
  title = {{MMSFormer: Multimodal Transformer for Material and Semantic Segmentation}},
  journal = {IEEE Open Journal of Signal Processing},
  volume = {5},
  pages = {599--610},
  year = {2024},
}

@inproceedings{Pei2025EfficientVMamba,
  author = {X. Pei and T. Huang and C. Xu},
  title = {{EfficientVMamba: Atrous Selective Scan for Light Weight Visual Mamba}},
  booktitle = {Proceedings of the AAAI Conference on Artificial Intelligence},
  pages = {6443--6451},
  year = {2025}
}

@inproceedings{Huang2024LocalMamba,
  author = {T. Huang and X. Pei and S. You and F. Wang and C. Qian and C. Xu},
  title = {{LocalMamba: Visual State Space Model with Windowed Selective Scan}},
  booktitle = {Advances in Neural Information Processing Systems},
  year = {2024}
}

@inproceedings{Liu2024VMamba,
  author = {Y. Liu and Y. Tian and Y. Zhao and H. Yu and L. Xie and Y. Wang and Q. Ye and Y. Liu},
  title = {{VMamba: Visual State Space Model}},
  booktitle = {Advances in Neural Information Processing Systems},
  year = {2024}
}

@inproceedings{Guo2024MambaIR,
  author = {Hang Guo and Jinmin Li and Tao Dai and Zhihao Ouyang and Xudong Ren and Shu-Tao Xia},
  title = {{MambaIR: A Simple Baseline for Image Restoration with State-Space Model}},
  booktitle = {European Conference on Computer Vision},
  year = {2024},
  publisher = {Springer},
}

@inproceedings{Yang2024PlainMamba,
  author = {Chenhongyi Yang and Zehui Chen and Miguel Espinosa and Linus Ericsson and Zhenyu Wang and Jiaming Liu and Elliot J. Crowley},
  title = {{PlainMamba: Improving Non-Hierarchical Mamba in Visual Recognition}},
  booktitle = {Proceedings of the British Machine Vision Conference},
  year = {2024}
}

@article{liu2022asphalt,
  title={Asphalt pavement crack detection based on convolutional neural network and infrared thermography},
  author={Liu, Fangyu and Liu, Jian and Wang, Linbing},
  journal={IEEE Transactions on Intelligent Transportation Systems},
  volume={23},
  number={11},
  pages={22145--22155},
  year={2022},
  publisher={IEEE}
}

@article{liu2024cmunet,
  title={CM-UNet: Hybrid CNN-Mamba UNet for Remote Sensing Image Semantic Segmentation},
  author={Liu Mushui and Dan, Jun and Lu, Ziqian and Yu, Yunlong and Li, Yingming and Li, Xi},
  journal={arXiv preprint arXiv:2405.10530},
  year={2024}
}

\end{document}